\documentclass[10pt,twocolumn,letterpaper]{article}

\usepackage[pagenumbers]{cvpr} 

\newcommand{\beginsupplement}{
    \setcounter{section}{0}
    \renewcommand{\thesection}{S}
    \renewcommand{\theHsection}{supp.\arabic{section}}
    \setcounter{subsection}{0}
    \renewcommand{\thesubsection}{S\arabic{subsection}}
    \renewcommand{\theHsubsection}{supp.\arabic{subsection}}
    \setcounter{subsubsection}{0}
    \renewcommand{\thesubsubsection}{S\arabic{subsection}.\arabic{subsubsection}}
    \renewcommand{\theHsubsubsection}{supp.\arabic{subsection}.\arabic{subsubsection}}
    \setcounter{table}{0}
    \renewcommand{\thetable}{S\arabic{table}}
    \renewcommand{\theHtable}{supp.\arabic{table}}
    \setcounter{figure}{0}
    \renewcommand{\thefigure}{S\arabic{figure}}
    \renewcommand{\theHfigure}{supp.\arabic{figure}}
    \setcounter{equation}{0}
    \renewcommand{\theequation}{S\arabic{equation}}
    \renewcommand{\theHequation}{supp.\arabic{equation}}
}

\usepackage{capt-of}
\makeatletter
\g@addto@macro\@maketitle{%
  \par\vspace{6pt}%
  \centering
  \includegraphics[width=\textwidth]{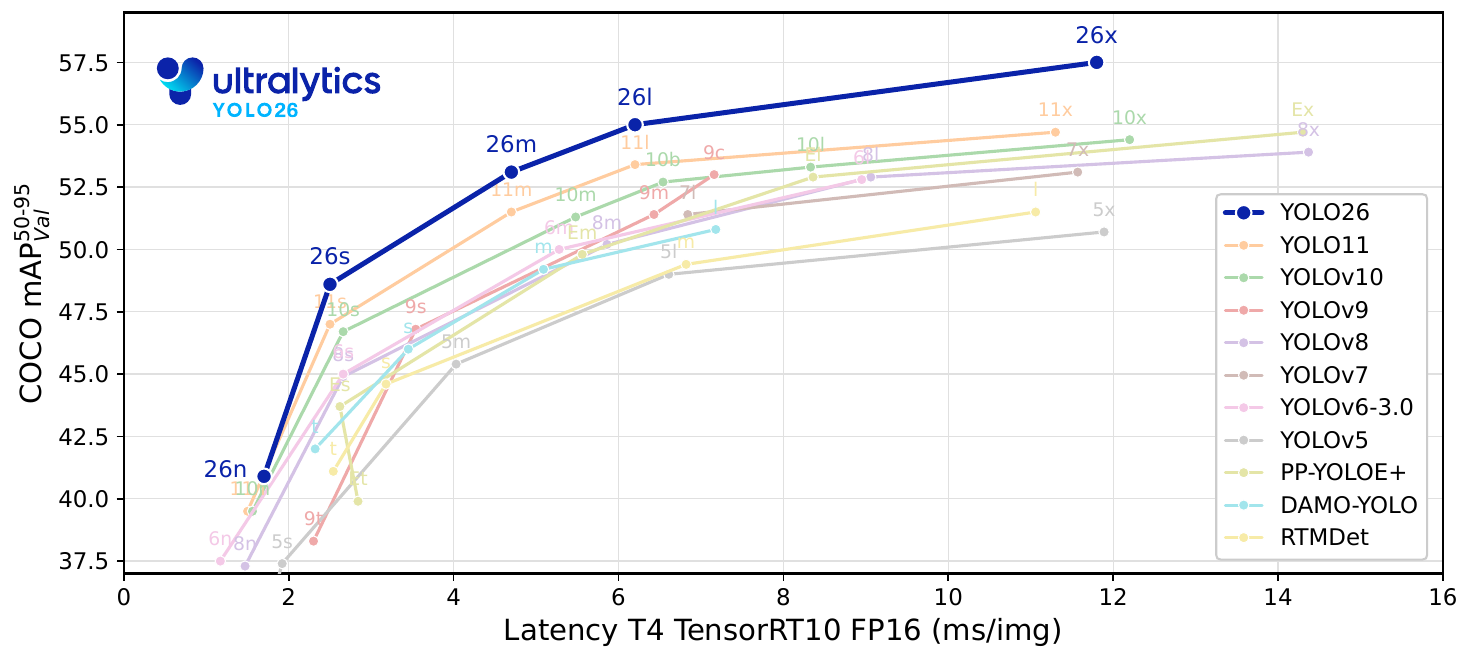}\par
  \vspace{2pt}%
  {%
    \@ifundefined{captionsetup}{}{\captionsetup{hypcap=false}}%
    \captionof{figure}{Accuracy--latency trade-off on COCO \texttt{val2017}. YOLO26 variants sit on or advance the Pareto front over prior YOLO versions and other real-time detectors, with the strongest AP--latency trade-off at the medium, large, and extra-large scales. Latency is measured on an NVIDIA T4 GPU with TensorRT FP16.}%
    \label{fig:teaser}%
  }%
  \vspace{6pt}%
}
\makeatother

\definecolor{cvprblue}{rgb}{0.21,0.49,0.74}
\usepackage[pagebackref,breaklinks,colorlinks,allcolors=cvprblue]{hyperref}

\def\paperID{*****} 
\def\confName{CVPR}
\def\confYear{2026}

\title{Ultralytics YOLO26: Unified Real-Time End-to-End Vision Models}

\author{
Glenn Jocher \qquad Jing Qiu \qquad Mengyu Liu \qquad Shuai Lyu \\ 
Fatih Cagatay Akyon \quad Muhammet Esat Kalfaoglu \\[6pt]
Ultralytics \\[-2pt]
{\small\texttt{\{glenn, jing, mengyu, louis, fatih, esat\}@ultralytics.com}}
}

\begin{document}
\maketitle

\begin{abstract}
Real-time vision demands models that are accurate, efficient, and simple to deploy across diverse hardware.
The YOLO family has become widely deployed for this reason, yet most YOLO detectors still rely on non-maximum suppression at inference, carry heavy detection heads due to Distribution Focal Loss, require long training schedules, and can leave the smallest objects without positive label assignments.
We present Ultralytics YOLO26, a unified real-time vision model family that addresses these limitations through coordinated architecture and training advances.
Architecturally, YOLO26 uses a dual-head design for native NMS-free end-to-end inference and removes DFL entirely, yielding a lighter head with unconstrained regression range.
For training, three techniques jointly improve accuracy while reducing cost: MuSGD, a hybrid Muon--SGD optimizer adapted from large language model training; Progressive Loss, which shifts supervision toward the inference-time head; and STAL, a label assignment strategy that guarantees positive coverage for small objects.
Beyond detection, YOLO26 introduces task-specific head and loss designs for instance segmentation, pose estimation, and oriented detection, producing consistent gains across tasks and scales.
The family spans five scales (n/s/m/l/x) and supports detection, instance segmentation, pose estimation, classification, and oriented detection in a single pipeline, with an open-vocabulary extension, YOLOE-26, for text-, visual-, and prompt-free inference.
Across all scales, YOLO26 achieves 40.9--57.5~mAP on COCO at 1.7--11.8,ms T4 TensorRT latency, advancing the accuracy--latency Pareto front over prior real-time detectors, while YOLOE-26x reaches 40.6~AP on LVIS minival under text prompting.
Code and models are available at \url{https://github.com/ultralytics/ultralytics}.
\end{abstract}

\section{Introduction}
\label{sec:intro}

Real-time object detection is a cornerstone of practical computer vision, powering applications from autonomous driving and robotics to surveillance and augmented reality, often on edge devices with tight latency and power budgets.
The field has advanced through several architectural shifts, but the central challenge remains unchanged: improving accuracy without sacrificing deployment simplicity and runtime efficiency.

Two-stage pipelines such as Faster R-CNN~\cite{ren2015faster} set strong accuracy baselines but at considerable inference cost.
One-stage detectors (SSD~\cite{liu2016ssd}, RetinaNet~\cite{lin2017focal}, YOLOv3~\cite{redmon2018yolov3}, Ultralytics YOLOv5~\cite{jocher2020yolov5}) traded proposal generation for dense prediction, drastically reducing latency.
Anchor-free designs such as FCOS~\cite{tian2019fcos} and Ultralytics YOLOv8~\cite{ultralytics2023yolov8} further simplified the detection head, while YOLOv10~\cite{wang2024yolov10} introduced consistent dual assignments to enable NMS-free inference.
In parallel, DETR~\cite{detr} cast detection as end-to-end set prediction, and its real-time descendants (RT-DETR~\cite{rtdetr}, D-FINE~\cite{dfine}, DEIM~\cite{deim}, RF-DETR~\cite{rfdetr}) have narrowed the accuracy gap with CNN-based detectors on standard benchmarks.
However, these transformer-based models often depend
on large pretrained vision backbones, deformable
attention or other custom operators, or fixed input
resolutions, which can complicate deployment across
heterogeneous hardware targets and reduce portability
across edge-oriented inference backends.
Throughout these shifts, the YOLO family has remained the most widely deployed real-time detector in industry.

Two structural properties underpin this durability.
First, \emph{deployment universality}: YOLO models rely on standard convolutional operators, enabling native export to TensorRT~\cite{tensorrt}, ONNX~\cite{onnx}, CoreML~\cite{coreml}, TFLite~\cite{tflite}, OpenVINO~\cite{openvino}, NCNN~\cite{ncnn_docs}, and ExecuTorch~\cite{executorch_docs} across cloud, mobile, and embedded platforms; Sec.~\ref{sec:model_variants_and_deployment} summarizes the broader export surface~\cite{ultralytics2026_export_docs}.
Second, \emph{multi-task unification}: a common backbone and neck support object detection, instance segmentation, image classification, pose estimation, and oriented bounding box (OBB) detection within a single training and deployment stack~\cite{ultralytics2026_tasks_docs}.
These properties make the YOLO paradigm a strong foundation for continued progress, provided its remaining limitations are addressed directly.

Despite this strong position, several concrete limitations persist across current YOLO-family detectors.
\textbf{(a)~NMS dependency and suboptimal dual-head training.}
Most CNN-based detectors still rely on non-maximum suppression at inference.
YOLOv10~\cite{wang2024yolov10} introduced a dual-head design to remove NMS, but applies identical loss weights to both heads throughout training: the one-to-one branch, the only head used at inference, receives the same optimization pressure as the dense one-to-many branch, leaving it under-trained relative to what targeted supervision could achieve.
\textbf{(b)~DFL parameter bloat and range limitation.}
The Distribution Focal Loss (DFL) module adopted since YOLOv8~\cite{ultralytics2023yolov8} expands bounding-box regression from 4~scalars to $4K$~logits per spatial location (typically $K{=}16$), inflating head parameters, an especially unfavorable trade-off for nano-scale models where the head can dominate total parameter count. For instance, YOLO11n with DFL has 2.6M parameters and 6.5~GFLOPs, whereas removing DFL reduces these figures to only 2.3M parameters and 5.2~GFLOPs, a 12\% parameter and 20\% FLOP reduction from this single module alone.
DFL also imposes a finite regression range of $(K{-}1) \times \text{stride}$ pixels per side, which becomes restrictive for large objects at high resolution.
D-FINE~\cite{dfine} addresses this by increasing~$K$, but the additional bins further compound head cost.
\textbf{(c)~Long training schedules.}
Standard SGD recipes still require roughly 600~epochs to reach competitive COCO accuracy, making rapid iteration expensive.
Recent work on the Muon optimizer~\cite{muon2025} has demonstrated roughly $2\times$ computational efficiency over AdamW in large language model training, yet no prior work has adapted Muon to object detection.
\textbf{(d)~TAL zero-assignment for small objects.}
The Task-Aligned Learning (TAL)~\cite{feng2021tood} label-assignment strategy selects candidate anchors based on geometric containment within ground-truth boxes.
After downsampling, objects whose spatial extent falls below the minimum stride have \emph{no} anchor inside their box, receiving zero positive assignments and zero gradient signal. In this regime, the smallest objects can receive no positive candidates and contribute no localization or classification signal during training.

We present \textbf{Ultralytics YOLO26}, a unified real-time vision model family built on YOLO11~\cite{ultralytics2024yolo11_docs}.
At the detector level, YOLO26 adopts a dual-head design for native NMS-free inference and removes DFL entirely, yielding a lighter regression head with unconstrained range.
To recover localization quality and improve optimization, YOLO26 combines three complementary training components: \emph{MuSGD}, a hybrid Muon--SGD optimizer~\cite{muon2025}; \emph{Progressive Loss}, which gradually shifts supervision toward the inference-time one-to-one head; and \emph{STAL} (Small-Target-Aware Label Assignment), which guarantees positive candidate coverage for tiny objects under TAL.

The resulting family spans five size variants (n/s/m/l/x) and supports detection, instance segmentation, classification, pose estimation, and OBB detection.
Beyond the shared detector, YOLO26 introduces task-specific extensions for instance segmentation, pose estimation, and oriented detection through a multi-scale proto pathway with auxiliary semantic supervision, an RLE-based uncertainty-aware keypoint objective~\cite{rle2021}, and a revised long-edge angle formulation with dedicated angle supervision.
Across scales, the YOLO26 family combines these extensions with the shared detector improvements, improving over YOLO11 by up to +3.7 mask AP on COCO instance segmentation, +7.2 pose AP on COCO keypoints, and +3.4 mAP on DOTA-v1.0 OBB detection.

Beyond closed-set detection, we extend the YOLO26 family to open-vocabulary scenarios by building \textbf{YOLOE-26}, which instantiates the open-vocabulary formulation of YOLOE~\cite{yoloe} on the YOLO26 detector.
YOLOE-26 retains the three inference modes of YOLOE (text-prompted, visual-prompted, and prompt-free) while introducing a stronger detector backbone, an upgraded text encoder (MobileCLIP2~\cite{mobileclip2}), a pseudo-label data engine, and decoupled segmentation training.
On LVIS minival, YOLOE-26x reaches 40.6~AP under text prompting, surpassing DetCLIP-T~\cite{yao2022detclip} by +6.2~AP (see Sec.~\ref{sec:suppl_yoloe_benchmarks}).

Figure~\ref{fig:teaser} summarizes the accuracy--latency trade-off on COCO \texttt{val2017}.
Across all model scales, YOLO26 sits on or advances the Pareto front, with the strongest AP--latency trade-off at the medium, large, and extra-large scales. At matched scales, YOLO26 improves COCO AP over YOLO11 by 1.6--2.8 points (full per-scale comparison in Sec.~\ref{sec:suppl_literature_comparison}) while also outperforming other recent real-time detectors across the accuracy--latency frontier.

In summary, the main contributions of this work are:
\begin{enumerate}
    \item A \textbf{DFL-free dual-head architecture} that provides native NMS-free end-to-end inference with a lighter regression head and unconstrained bounding-box range, while retaining an optional dense branch for accuracy-critical deployment.

    \item A \textbf{coordinated training pipeline} of three complementary components, applied jointly (MuSGD, Progressive Loss, and STAL), that accelerate convergence, align optimization with the end-to-end inference path, and guarantee supervision for the smallest objects.

    \item \textbf{Task-specific head and loss designs} for instance segmentation (multi-scale prototype fusion with auxiliary semantic supervision), pose estimation (uncertainty-aware RLE keypoint regression), and oriented detection (long-edge angle formulation with aspect-ratio-aware supervision), each yielding consistent gains over YOLO11 within a single unified family.

    \item \textbf{YOLOE-26}, an open-vocabulary extension that applies the YOLOE~\cite{yoloe} formulation to the YOLO26 detector with an upgraded text encoder, a pseudo-label data engine, and decoupled segmentation training, reaching 40.6~AP on LVIS minival under text prompting.
\end{enumerate}

\section{Related Work}
\subsection{CNN-based Object Detection: From Two-Stage to NMS-Free}

CNN-based object detection has progressed through design shifts that balance accuracy and efficiency. Early approaches were dominated by \emph{two-stage} pipelines that first generate candidate regions and then classify and refine them. R-CNN introduced this paradigm by combining region proposals with CNN features, Fast R-CNN improved efficiency by sharing computation across proposals via RoI pooling, and Faster R-CNN integrated proposal generation into the network with an RPN to enable end-to-end training and a strong speed--accuracy trade-off~\cite{girshick2014rich,girshick2015fast,ren2015faster}.

To reduce latency, \emph{one-stage} detectors remove the proposal stage and predict classes and boxes densely over feature maps. SSD demonstrated efficient multi-scale dense prediction with default boxes~\cite{liu2016ssd}, while RetinaNet addressed foreground--background imbalance via Focal Loss to improve dense detection accuracy~\cite{lin2017focal}. The YOLO line targets real-time operation, with YOLOv3 strengthening multi-scale prediction and backbone design~\cite{redmon2018yolov3}, and Ultralytics YOLOv5 providing a widely adopted, deployment-oriented implementation with refined training recipes~\cite{jocher2020yolov5}.

Anchor-free detectors further simplify dense prediction by removing hand-designed anchors and improving portability across datasets. FCOS formulates detection as per-pixel classification and box regression on feature maps~\cite{tian2019fcos}, while CenterNet represents objects via keypoints and their geometric relations~\cite{duan2019centernet}. Recent real-time systems have also adopted anchor-free heads in practice; Ultralytics YOLOv8 follows this direction with an anchor-free split-head design and deployment-oriented refinements for accuracy and efficiency~\cite{ultralytics2023yolov8}.

Beyond anchors, recent work targets \emph{end-to-end} detection by reducing reliance on post-processing, especially non-maximum suppression (NMS). YOLOv10 enables NMS-free inference via consistent dual assignments with two training branches: a one-to-many branch for dense supervision and a one-to-one branch that learns a single matched prediction per ground-truth instance for direct decoding at inference~\cite{wang2024yolov10}.

\subsection{Transformer-based Object Detection}

Transformer-based detectors cast detection as end-to-end \emph{set prediction} with one-to-one Hungarian matching, removing anchors and NMS. DETR introduced this formulation with a transformer encoder--decoder and a fixed set of object queries~\cite{detr}. Deformable DETR improved efficiency and convergence via sparse deformable attention, and introduced iterative box refinement and a two-stage top-$K$ proposal initialization variant~\cite{deformable_detr}. Follow-up work refined query design and training: DAB-DETR parameterizes queries as learnable anchor boxes refined across decoder layers~\cite{dab_detr}, DN-DETR accelerates convergence with denoising queries built from noisy ground-truth targets~\cite{dn_detr}, and DINO further improves initialization and optimization using mixed query selection, contrastive denoising, and a look-forward-twice refinement scheme~\cite{dino}. Beyond pure one-to-one training, hybrid matching methods increase positive supervision by combining one-to-one with auxiliary one-to-many assignments during training (e.g., H-DETR) while preserving one-to-one inference, and Group DETR similarly provides richer supervision via multiple query groups with per-group matching~\cite{hdetr,group_detr}.

Transformer detectors have also been adapted for real-time use by improving multi-scale feature handling and reducing computational overhead. RT-DETR targets practical speed--accuracy trade-offs via an efficient encoder design and deployment-oriented choices~\cite{rtdetr}, with later refinements such as RT-DETRv4 exploring stronger distillation strategies for compact models~\cite{rtdetrv4}. Concurrently, methods such as D-FINE and DEIM improve localization and training efficacy through refined regression/optimization and matching-aware objectives, with follow-up versions (e.g., DEIMv2) continuing to optimize convergence and accuracy in real-time DETR pipelines~\cite{dfine,deim,deimv2}. Lightweight designs such as LW-DETR further streamline the architecture for low-latency deployment~\cite{lwdetr}, and RF-DETR represents complementary efforts to build strong real-time detection transformer families with practical accuracy--latency trade-offs~\cite{rfdetr}.

\subsection{Instance Segmentation}

Instance segmentation extends object detection by predicting a pixel-accurate mask for each instance, with a persistent trade-off between accuracy and real-time efficiency. Mask R-CNN augments a two-stage detector with an RoI-aligned mask branch to produce high-quality masks~\cite{he2017mask}. To avoid per-RoI computation, proposal-free methods predict masks fully convolutionally: CondInst generates instance-conditioned mask heads via dynamically predicted convolution parameters applied to shared mask features~\cite{tian2020condinst}, while SOLOv2 assigns instances to locations and predicts dynamic mask kernels per location to produce masks from shared mask features~\cite{wang2020solov2}.

For real-time settings, \emph{prototype-based} approaches construct instance masks from a small set of shared bases. YOLACT predicts global prototype masks and per-instance coefficients, forming each mask through a linear combination followed by cropping, which enables fast mask generation with competitive accuracy~\cite{bolya2019yolact}. Ultralytics YOLO segmentation heads adopt a closely related prototype--coefficient formulation (e.g., YOLO11), predicting shared prototypes together with per-instance mask coefficients from detection features to build masks with minimal per-instance overhead~\cite{ultralytics2024yolo11_docs}.

More recently, transformer-based instance segmentation predicts masks via set-based decoding with stronger global context. Mask2Former employs masked-attention decoding to output a set of instance masks and corresponding classes~\cite{cheng2022mask2former}. MaskDINO further improves convergence and mask quality by incorporating denoising-based training and enhanced mask decoding within a unified transformer framework~\cite{li2023maskdino}.

\subsection{Pose Estimation}

Human pose estimation progressed from direct coordinate regression (DeepPose~\cite{deeppose2014}) to heatmap prediction via encoder--decoder networks (Stacked Hourglass~\cite{hourglass2016}) and deconvolutional baselines (SimpleBaselines~\cite{simplebaselines2018}). HRNet maintains high-resolution representations through parallel subnetwork fusion~\cite{hrnet2019}, while OpenPose takes a bottom-up approach by jointly predicting keypoints and Part Affinity Fields~\cite{openpose2021}. ViTPose demonstrated scalable transformer backbones for this task~\cite{vitpose2022}, and RTMPose achieves real-time performance via CSPNeXt with a coordinate classification head~\cite{rtmpose2023}.

Within YOLO-based pose estimation, YOLO-Pose introduced a heatmap-free single-stage formulation that jointly detects persons and regresses keypoints using a scale-aware OKS-based training objective~\cite{maji2022yolo}. Later YOLOv7 integrated a keypoint head into the broader YOLO family, and YOLOv8 and YOLO11 continued this direction with anchor-free pose heads~\cite{wang2023yolov7,ultralytics2023yolov8,ultralytics2024yolo11_docs}. YOLO26 builds on this direct-regression lineage and further augments it with RLE~\cite{rle2021}, replacing conventional regression with a normalizing-flow-based probabilistic objective for more accurate and calibrated keypoint localization.

\subsection{Oriented Bounding Box Detection}

Oriented object detection extends axis-aligned detection to boxes at arbitrary rotation angles, motivated by aerial imagery and scene text where upright boxes produce excessive overlap. Early methods adapted region-proposal networks with rotated anchors~\cite{rrpn2018}, while two-stage detectors such as RoI Transformer~\cite{roitransformer2019} and Oriented R-CNN~\cite{orientedrcnn2021} applied rotation-aware heads to achieve strong accuracy. A central difficulty is \emph{angle representation}: na\"{i}ve regression suffers from boundary discontinuities at the periodicity boundary ($\pm\pi/2$ for symmetric boxes), where a small geometric change causes a large loss spike. Circular Smooth Label (CSL)~\cite{csl2020} addresses this by recasting angle prediction as circular classification. Gaussian-based methods bypass the issue by modelling rotated boxes as 2D Gaussians and measuring similarity via Wasserstein distance (GWD~\cite{gwd2021}) or a Kalman-filter-inspired IoU surrogate (KFIoU~\cite{kfiou2023}). However, the Gaussian modeling method would cause angle ambiguity of square rotated objects, and an additional angle loss~\cite{yu2023mkiou} was proposed to penalize angle deviation. PSC and PSCD~\cite{yu2023phase} introduce phase based orientation angle coders, where PSC solves the boundary discontinuity issue and PSCD further reduces the ambiguity of square-like objects by introducing a dual-frequency phase representation. Anchor-free designs such as S2ANet~\cite{s2anet2021} align features to oriented proposals for finer localization, and modern backbones such as LSKNet~\cite{lsknet2023} exploit large selective kernels suited to the elongated structures common in aerial scenes.

Within the YOLO family, YOLOv8~\cite{ultralytics2023yolov8} and YOLO11~\cite{ultralytics2024yolo11_docs} introduced OBB heads with a dedicated angle branch trained via ProbIoU~\cite{probiou2021}. YOLO26 extends this lineage with a dedicated aspect-ratio-aware angle loss and an optimized decoder, addressing angle ambiguity and boundary discontinuities in rotated box representations.

\subsection{Open-Vocabulary Detection and Segmentation}
\label{sec:related_ov}

Open-vocabulary detection methods can be broadly categorized
by how they specify target categories at inference time:
through text prompts, visual prompts, or no prompts at all.
\textbf{Text-prompted detection} grounds natural-language
queries to visual regions via vision--language pretraining.
GLIP~\cite{li2022glip} reformulates detection as phrase
grounding, aligning region features with token-level text
embeddings through a contrastive objective.
Grounding DINO~\cite{liu2024groundingdino} extends this idea
to a transformer detector with tight cross-modal fusion,
achieving strong zero-shot localization.
YOLO-World~\cite{cheng2024yoloworld} brings text-conditioned
detection into the real-time regime by injecting CLIP text
embeddings into a YOLO neck via re-parameterizable
cross-attention, and YOLO-UniOW~\cite{wang2024yolouniow}
further unifies open-world detection under a single
YOLO-family model.
Despite their generality, text prompts are ill-suited for
objects that resist concise linguistic description, such as
novel industrial defects or fine-grained biological specimens.
\textbf{Visual-prompted detection} addresses this gap by
supplying reference images or regions in place of text.
OWL-ViT~\cite{minderer2022owlvit} and
OV-DETR~\cite{zang2022ovdetr} process image exemplars through
a shared CLIP encoder alongside text queries, supporting both
modalities under a unified architecture.
DINOv~\cite{li2024dinov} explores reference regions as
in-context examples for generic and referring vision tasks,
while T-Rex2~\cite{jiang2024trex2} achieves tighter
multi-modal alignment via region-level contrastive training.
For segmentation, SEEM~\cite{zou2024seem} and
Semantic-SAM~\cite{li2024semanticsam} handle diverse prompt
types---including points, boxes, and reference masks---across
panoptic and part-level granularities.
\textbf{Prompt-free detection} removes the dependence on
explicit queries altogether by coupling detectors with
generative language models.
GRiT~\cite{wu2022grit} attaches an autoregressive text
decoder to a region-proposal backbone for joint dense
captioning and detection.
DetCLIPv3~\cite{yao2024detclipv3} trains an object captioner
on web-scale data to generate rich label information for
arbitrary regions, and
GenerateU~\cite{lin2024generateu} uses a language model to
produce object names in free form, decoupling recognition from
any predefined vocabulary.
YOLOE~\cite{yoloe} unifies all three paradigms---text-prompted,
visual-prompted, and prompt-free inference---within a single
real-time model through its RepRTA, SAVPE, and LRPCHead components.
Our YOLOE-26 inherits this unified formulation and advances it
with a stronger backbone, an upgraded text encoder, a
pseudo-label data engine, and decoupled segmentation training.

\section{Methodology}

YOLO26 builds on the YOLO11 family as a unified
real-time vision framework, while revising both the
shared detector design and the task-specific heads.
At the detection level, the methodology combines a
dual-head end-to-end formulation, DFL-free box
regression, MuSGD, Progressive Loss, and
Small-Target-Aware Label Assignment (STAL). This
section first presents the shared architecture and
training pipeline, and then describes the
task-specific extensions for segmentation, pose
estimation, oriented bounding boxes, and the
open-vocabulary YOLOE-26 variant.

\subsection{Overview}

Relative to YOLO11~\cite{ultralytics2024yolo11_docs},
YOLO26 is organized around three design goals:
\emph{end-to-end simplicity}, \emph{deployment
efficiency}, and \emph{stronger optimization}. These
goals are realized through a native NMS-free
one-to-one inference path, a lightweight
DFL-free regression head, and a training pipeline
that couples MuSGD, Progressive Loss, and STAL.

\begin{figure*}[t]
    \centering
    \makebox[\textwidth][c]{\includegraphics[width=1.1\textwidth]{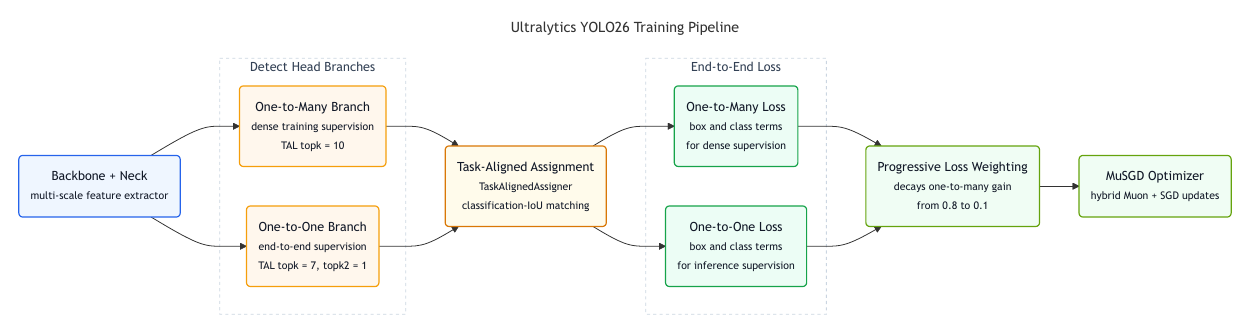}}
    \caption{Training pipeline of Ultralytics YOLO26. The shared backbone and neck feed the dual detection heads, STAL improves assignment robustness for tiny targets, Progressive Loss reweights the one-to-many and one-to-one objectives over training, and MuSGD performs the final parameter updates.}
    \label{fig:yolo26_training_pipeline}
\end{figure*}

Figure~\ref{fig:yolo26_training_pipeline} summarizes how these components interact during training: the shared backbone and neck feed the one-to-many and one-to-one heads, STAL preserves tiny-object assignments, Progressive Loss reweights the branch objectives over time, and MuSGD updates the model parameters.

\subsection{Architecture Design}

The main paper focuses on the architectural ideas most directly tied to the detection objective, while the full model schematic and the constituent block compositions are provided in Sec.~\ref{sec:suppl_architecture} of the supplementary materials; see also Supplementary Figs.~\ref{fig:suppl_yolo26_architecture} and~\ref{fig:suppl_yolo26_blocks}. Table~\ref{tab:method_component_summary} maps the main detection components to their implementation names in the Ultralytics repository.

\begin{table}[t]
\centering
\small
\setlength{\tabcolsep}{4pt}
\resizebox{\columnwidth}{!}{%
\begin{tabular}{lll}
\toprule
\textbf{Component} & \textbf{Implementation} & \textbf{Role} \\
\midrule
Dual head & \texttt{Detect} & E2E and NMS paths \\
Direct regression & \texttt{reg\_max=1} & DFL-free boxes \\
STAL & \texttt{TaskAlignedAssigner} & Tiny-object coverage \\
Progressive Loss & \texttt{E2ELoss} & Branch reweighting \\
Optimizer & \texttt{MuSGD} & Faster convergence \\
\bottomrule
\end{tabular}
}
\vspace{2pt}
\caption{Main YOLO26 detection components and their corresponding implementation names in the Ultralytics repository.}
\label{tab:method_component_summary}
\end{table}

\subsubsection{End-to-End NMS-Free Detection}

Most CNN-based detectors rely on non-maximum suppression (NMS) to remove duplicate predictions at inference time. YOLO26 instead exposes a \emph{dual-head} design that supports both end-to-end NMS-free decoding and conventional dense prediction, enabling deployment-specific trade-offs.

\paragraph{One-to-One Head (default).}
The one-to-one head produces a fixed-size set of predictions without requiring NMS, outputting at most 300 detections per image (shape $(N,300,6)$ for batch size $N$). Following YOLOv10~\cite{wang2024yolov10}, we train the dual heads with \emph{consistent dual-path label assignment} based on Task-Aligned Learning (TAL)~\cite{feng2021tood}: both heads use the same TAL formulation, but with different matching cardinalities. In the current implementation, the one-to-one path first forms a TAL candidate set with $\text{topk}=7$ and then applies a secondary $\text{topk2}=1$ filter, yielding a unique end-to-end assignment per ground-truth instance and enabling direct decoding at inference.

\paragraph{One-to-Many Head.}
The one-to-many head retains the standard dense YOLO-style prediction and uses TAL with $\text{topk}=10$ to provide richer positive supervision during training, producing outputs of shape $(N, n_c+4, 8400)$ where $n_c$ is the number of classes. This head uses NMS at inference and typically yields slightly higher accuracy at the cost of additional post-processing overhead.

Overall, the dual-head design offers a practical accuracy--latency knob: the one-to-one head prioritizes simplicity and speed (NMS-free, fixed output), while the one-to-many head targets maximum accuracy when NMS overhead is acceptable.

\subsubsection{Distribution Focal Loss Removal}
\label{sec:distribution_focal_loss_removal}

YOLO26 removes the Distribution Focal Loss (DFL) module from the detection head. DFL was introduced in Generalized Focal Loss (GFL)~\cite{li2020gfl} as a distribution-based box-regression formulation and is now widely used in recent YOLO-style detectors, including Ultralytics YOLOv8~\cite{ultralytics2023yolov8} and later variants such as YOLOv9~\cite{wang2024yolov9}, YOLOv10~\cite{wang2024yolov10}, and YOLO11~\cite{ultralytics2024yolo11_docs}. In DFL, each box side is predicted as a discrete distribution over $K$ bins (typically $K{=}16$) and decoded by expectation:
\begin{equation}
d=\sum_{i=0}^{K-1} i \cdot \mathrm{softmax}(z)_i,\quad z\in\mathbb{R}^{K},
\end{equation}
which expands regression from 4 scalars to $4K$ logits per location and increases head parameters/compute accordingly---a particularly unfavorable trade-off for nano-scale models.

DFL also imposes a finite discrete support range. Since $d\in[0,K{-}1]$ before multiplying by stride $s$, the maximum per-side distance is $(K{-}1)s$ pixels (thus width/height are bounded by $\approx 2(K{-}1)s$ at a given feature level). With $K{=}16$, this is $\approx 30s$ (e.g., $\sim960$ pixels at $s{=}32$), which can become restrictive for large objects at high resolution (e.g., 1280). Increasing $K$ (e.g., D-FINE uses larger bin counts) alleviates this but further increases head cost~\cite{dfine}. YOLO26 instead adopts a simpler regression head and recovers localization quality via complementary training objectives, namely Progressive Loss (Sec.~\ref{sec:progressive_loss}) and STAL (Sec.~\ref{sec:small_object_aware_label_assignment}).
The quantitative and qualitative evidence for this design choice is provided in the DFL-removal ablation in Sec.~\ref{sec:dfl_removal_ablation}. In particular, Fig.~\ref{fig:dfl_large_qualitative} uses 1280-resolution training setups to expose a failure mode that is less apparent at 640, showing that the DFL-free head more reliably preserves the full extent of large objects.

Finally, removing DFL simplifies export and improves compatibility with constrained runtimes and accelerators that favor standard operators and minimal decoding overhead. Overall, this choice reflects an explicit accuracy--efficiency trade-off for resource-constrained deployment while preserving strong detection performance through compensating training enhancements.

\subsection{Training Methodology}

Starting from the shared backbone and neck features, YOLO26 optimizes the one-to-many and one-to-one heads jointly, applies STAL during label assignment to preserve supervision for tiny targets, and combines the two branch losses through Progressive Loss before updating model parameters with MuSGD. This view highlights how the proposed training components interact as a single pipeline rather than as isolated modifications.

\subsubsection{MuSGD Optimizer}

YOLO26 adopts \emph{MuSGD}, a hybrid optimizer that combines Muon~\cite{muon2025} with standard SGD-momentum~\cite{sgd_pytorch}, motivated by recent large-scale training practice~\cite{kimi_k2}. Muon applies momentum updates followed by a lightweight orthogonalization of the momentum-derived update (via a few Newton--Schulz iterations), which improves update conditioning and can stabilize optimization~\cite{muon2025}. We leverage this property for detector training while retaining SGD as a robust baseline component.

Concretely, MuSGD applies a weighted mixture of the Muon update and the SGD update to multi-dimensional parameters (e.g., convolution kernels and linear weights), and uses pure SGD for 1D parameters such as biases and normalization scales. This parameter-type split keeps scale/shift parameters stable while benefiting from orthogonalized updates on high-dimensional weight tensors, improving training stability and accelerating convergence in practice (see Sec.~\ref{sec:musgd_ablation}).


\subsubsection{Progressive Loss}
\label{sec:progressive_loss}

Dual-head end-to-end detectors exhibit an inherent optimization asymmetry during training. The dense one-to-many branch receives broader positive supervision and is therefore easier to optimize in the early stage, whereas the one-to-one branch is more constrained but ultimately determines the model's end-to-end inference behavior. Applying fixed, identical loss weights to both branches throughout training under-utilizes this asymmetry and can leave the inference branch insufficiently optimized.

To address this issue, YOLO26 introduces \emph{Progressive Loss}, a curriculum-style reweighting strategy that gradually transfers optimization emphasis from the dense branch to the one-to-one branch over the course of training. Early in optimization, the dense branch is emphasized to stabilize feature learning and provide reliable supervision. As training progresses, the one-to-one branch receives increasingly greater emphasis so that the optimization objective becomes better aligned with the NMS-free inference path used at deployment.

Formally, the total detection loss is written as
\begin{equation}
\mathcal{L}_{\text{total}}
=
\alpha(t)\,\mathcal{L}_{\text{one2many}}
+
\bigl(1-\alpha(t)\bigr)\,\mathcal{L}_{\text{one2one}},
\end{equation}
where $t$ denotes the current epoch and $\alpha(t)$ is a linearly decreasing schedule defined by
\begin{equation}
\alpha(t) = \max\!\left(1 - \frac{t}{\max(E-1,\,1)},\, 0\right)\,(\alpha_{\text{init}} - \alpha_{\text{final}}) + \alpha_{\text{final}},
\end{equation}
with $E$ the total number of training epochs, $\alpha_{\text{init}}$ the initial one-to-many weight, and $\alpha_{\text{final}}$ the final one-to-many weight.
Thus, the objective transitions smoothly from dense supervision to inference-oriented supervision over time.
The exact values of $\alpha_{\text{init}}$, $\alpha_{\text{final}}$, and branch-specific assignment settings are deferred to the implementation details in Sec.~\ref{sec:implementation_details}.

Progressive Loss complements the asymmetric assignment used by the two branches: the one-to-many branch benefits from richer candidate supervision, while the one-to-one branch uses a stricter assignment tailored to end-to-end prediction. By matching the loss emphasis to these distinct roles, Progressive Loss improves early-stage optimization stability while better aligning the final model with deployment-time behavior.

\subsubsection{Small-Target-Aware Label Assignment (STAL)}
\label{sec:small_object_aware_label_assignment}

Task-aligned assignment first restricts supervision to anchor centers that fall inside each ground-truth box. While this geometric filtering is effective for normal-scale objects, it becomes brittle for very small instances: after feature-map discretization, a tiny box may contain no valid anchor centers at all. In that case, the object receives zero positive candidates and contributes no localization or classification signal during training.

YOLO26 addresses this failure mode with \emph{Small-Target-Aware Label Assignment} (STAL), which decouples the geometry used for candidate selection from the geometry used for regression. Let a ground-truth box be parameterized as $g_i=(x_i, y_i, w_i, h_i)$, and let $s_{\min}$ denote the smallest feature-pyramid stride. During candidate filtering only, STAL constructs an assignment surrogate
\begin{equation}
\tilde{g}_i=(x_i, y_i, \tilde{w}_i, \tilde{h}_i),
\end{equation}
where each spatial dimension is adjusted independently as
\begin{equation}
\tilde{d}_i =
\begin{cases}
s_{\mathrm{ref}}, & d_i < s_{\min}, \\
d_i, & \text{otherwise},
\end{cases}
\qquad d_i \in \{w_i, h_i\},
\end{equation}
where $s_{\mathrm{ref}}$ is a fixed reference size derived from the feature pyramid. In the current implementation, it is set to the second pyramid stride when available; the exact default values are given in Sec.~\ref{sec:implementation_details}. For each ground-truth object $i$ and anchor center $a_j$, we then define a binary candidate mask
\begin{equation}
M_{ij} =
\begin{cases}
1, & \text{if anchor center } a_j \text{ lies inside } \tilde{g}_i, \\
0, & \text{otherwise},
\end{cases}
\end{equation}
where $a_j$ is the $j$-th anchor center.

Importantly, STAL modifies only the candidate-selection mask. The original box $g_i$ is preserved for task-aligned scoring, final target assignment, and box regression, so the detector is still optimized against the true object extent. This makes STAL deliberately conservative: it does not alter localization targets or inflate supervision for ordinary objects, but it prevents pathological zero-positive cases for tiny instances that would otherwise be dropped by the assignment pipeline.

\subsection{Task-Specific Extensions}

Beyond the shared detection architecture and training
methodology, YOLO26 introduces task-specific
extensions for instance segmentation, pose
estimation, and oriented bounding box detection.
These extensions preserve the common end-to-end
backbone and neck while adapting the prediction
heads and supervision schemes to the structural
requirements of each task.

\paragraph{Image Classification.} YOLO26 classification variants reuse the standard Ultralytics \texttt{Classify} head on the shared backbone. Because the classification branch introduces no new task-specific decoding rule, it is treated as a supported model-family task rather than a separate methodological contribution; the optimizer transfer check is reported in Sec.~\ref{sec:suppl_musgd_cls} of the supplementary materials.

\subsubsection{Instance Segmentation}\label{sec:ins_seg}

YOLO-style instance segmentation adopts a prototype-based representation, where a shared prototype tensor is predicted once per image and each positive instance predicts a coefficient vector to reconstruct its mask~\cite{bolya2019yolact, ultralytics2024yolo11_docs}. Given a prototype tensor $\mathbf{P} \in \mathbb{R}^{K \times H \times W}$, where $\mathbf{P}_k \in \mathbb{R}^{H \times W}$ denotes the $k$-th prototype map, and instance-specific coefficients $\mathbf{c}_i \in \mathbb{R}^{K}$, the mask logit for instance $i$, denoted by $\hat{\mathbf{M}}_i \in \mathbb{R}^{H \times W}$, is formed as
\begin{equation}
\hat{\mathbf{M}}_i = \sum_{k=1}^{K} c_{ik} \mathbf{P}_k.
\end{equation}
YOLO26 retains this lightweight reconstruction rule, while introducing two segmentation-specific integrations: a multi-scale proto pathway for prototype generation and a training-only auxiliary semantic supervision branch.

\paragraph{Multi-Scale Proto Module.} Let $\{\mathbf{X}_\ell\}_{\ell=1}^{L}$ denote the segmentation features across pyramid levels, with $\mathbf{X}_1$ being the highest-resolution feature. Standard YOLO segmentation heads generate prototypes from $\mathbf{X}_1$ alone. YOLO26 instead constructs a fused proto feature
\begin{equation}
\mathbf{F}_{\text{proto}} = \mathbf{X}_1 + \sum_{\ell=2}^{L} \mathcal{U}\!\left(\phi_\ell(\mathbf{X}_\ell)\right),
\end{equation}
where $\phi_\ell(\cdot)$ is a learnable $1{\times}1$ projection and $\mathcal{U}(\cdot)$ upsamples features to the spatial resolution of $\mathbf{X}_1$. The shared prototype tensor is then produced as
\begin{equation}
\mathbf{P} = \tilde{\mathcal{G}}(\mathbf{F}_{\text{proto}}),
\end{equation}
where $\tilde{\mathcal{G}}(\cdot)$ denotes the prototype-generation stack applied to the fused feature. This modification preserves the prototype-coefficient formulation while enriching the prototypes with higher-level semantic context and broader scale coverage.

\paragraph{Auxiliary Semantic Segmentation Loss.} YOLO26 further attaches a training-only semantic segmentation branch to the shared fused feature $\mathbf{F}_{\text{proto}}$, predicting dense per-class logits before prototype generation. The supervision target is a semantic mask derived from the instance annotations by merging mask pixels according to their class labels. We supervise this branch with a balanced BCE+Dice objective, which provides dense class-aware gradients in addition to the instance-mask loss. Importantly, this branch is auxiliary: it is inactive at evaluation time and is explicitly removed during model fusion, so it introduces no additional inference-time cost.

\subsubsection{Pose Estimation}\label{sec:pose_estimation}

In prior YOLO pose models, a pose head is attached to directly regress the keypoint coordinates $(x, y)$ and visibility scores, and training uses an Object Keypoint Similarity (OKS)-based loss~\cite{maji2022yolo}, which normalizes keypoint localization error by person scale and per-keypoint OKS constants. YOLO26 extends this scheme with \emph{Residual Log-Likelihood Estimation} (RLE)~\cite{rle2021} to achieve principled per-joint uncertainty modeling. In addition to coordinate outputs, a parallel \emph{sigma branch} predicts per-axis uncertainty $\boldsymbol{\sigma}{=}(\sigma_x,\sigma_y)\!\in\!(0,1)^2$ for each joint. The coordinate residual is normalized accordingly:
\begin{equation}
  \boldsymbol{\varepsilon} = \frac{\hat{\mathbf{x}} - \mathbf{x}^{*}}{\boldsymbol{\sigma}},
\end{equation}
where $\hat{\mathbf{x}}$ is the predicted joint location and $\mathbf{x}^{*}$ is the ground truth. A shared \emph{RealNVP} normalizing flow~\cite{dinh2017realnvp} estimates $\log\varphi(\boldsymbol{\varepsilon})$, the log-density of the normalized residual under a learned distribution. The training objective combines an explicit base (e.g., Laplace) negative log-likelihood term $\mathcal{L}_{q}$ with the learned residual term:
\begin{equation}
\begin{aligned}
  \mathcal{L}_{\mathrm{RLE}}
    &= \log\boldsymbol{\sigma} - \log\varphi(\boldsymbol{\varepsilon}) + \mathcal{L}_{q}(\sigma, \varepsilon) \\
    &= \log\boldsymbol{\sigma} - \log\varphi(\boldsymbol{\varepsilon})
    + \underbrace{\log(2\boldsymbol{\sigma}) + |\boldsymbol{\varepsilon}|}_{-\log\,\mathrm{Laplace}(\hat{\mathbf{x}};\,\mathbf{x}^{*},\boldsymbol{\sigma})},
\end{aligned}
\end{equation}
where the residual term anchors the uncertainty scale and stabilizes early training. Joints with higher predicted $\boldsymbol{\sigma}$---e.g., occluded or inherently ambiguous keypoints---are effectively down-weighted, yielding improved localization without discarding any predictions. The decoding path is further streamlined for inference speed.

\subsubsection{Oriented Bounding Box Detection}

\paragraph{YOLO26 OBB Parameterization.} 
In YOLO OBB models, a separate branch is adopted to predict the orientation angle. In previous versions, the oriented bounding box follows the OpenCV~\cite{bradski2000opencv} convention, where the angle is defined as the acute angle between the box width and the positive x-axis, with a range of $(0, 90^\circ]$. Under this definition, width and height are not strictly fixed, which introduces ambiguity for objects whose orientations are close to 0 or $90^\circ$. Small orientation changes near these boundaries can cause edge swapping between width and height, making angle regression discontinuous and unstable. In YOLO26, the angle definition is changed to the long-edge definition following MMRotate~\cite{zhou2022mmrotate}, where the angle range is $[-45^\circ, 135^\circ)$ and the width is constrained to be larger than the height. This formulation alleviates the boundary ambiguity near 0 or $90^\circ$ and reduces the instability caused by edge swapping. In addition, previous OBB models predict an oriented angle logit $z$ and then map it through a sigmoid transform,
\begin{equation}
  \hat{\theta} = \left(\sigma(z) - 0.25\right)\pi,
\end{equation}
which compresses predictions into a fixed interval and introduces an additional squashing nonlinearity near the interval boundaries. YOLO26 instead predicts the angle directly,
\begin{equation}
  \hat{\theta} = z,
\end{equation}
removing the extra squashing nonlinearity.

\paragraph{Angle-Loss for Square Objects.} For square or near-square objects, the ProbIoU \cite{probiou2021} loss used in YOLO11 models becomes insensitive to angle variations because the Gaussian representation is nearly invariant to rotation when width $\approx$ height, making angle prediction ambiguous and unstable. To address this issue, an angle loss is specifically designed for square objects in YOLO26. We first recap the angle supervision used in the rotated box formulation. Since an oriented rectangle is unchanged under a $180^\circ$ rotation, $(x, y, w, h, \theta)$ and $(x, y, w, h, \theta + \pi)$ represent the same geometry. The angular residual should therefore be measured modulo $\pi$ rather than on the real line. Let
\begin{equation}
  \Delta \theta_i = \hat{\theta}_i - \theta_i^*, \qquad
  \tilde{\Delta \theta}_i
  =
  \Delta \theta_i
  -
  \mathrm{round}\!\left(\frac{\Delta \theta_i}{\pi}\right)\pi,
\end{equation}
where $\hat{\theta}_i$ and $\theta_i^*$ denote the predicted and target angles, respectively. The angle loss is then defined as
\begin{equation}
  \begin{aligned}
  \mathcal{L}_{\mathrm{angle}}
  =
  \frac{1}{S}
  \sum_{i\in\mathcal{F}}
  q_i\,\omega_i \,
  \sin^2\!\left(2\tilde{\Delta \theta}_i\right),
  \\
  S = \max\!\left(\sum_{i\in\mathcal{F}} q_i, 1\right),
  \qquad
  \omega_i
  =
  \exp\!\left(
    -\frac{\log^2(w_i^*/h_i^*)}{\lambda^2}
  \right).
  \end{aligned}
\end{equation}
where $\mathcal{F}$ denotes the foreground assignments, $q_i$ is the assignment weight produced by TAL, $S$ is the corresponding normalization factor, $\omega_i$ denotes an aspect-ratio-aware factor computed from the target box dimensions $(w_i^*, h_i^*)$, and $\lambda$ is the hyper-parameter. The double-angle penalty is used as auxiliary supervision for square and near-square boxes, for which rotations separated by $90^\circ$ become geometrically ambiguous. Elongated boxes receive smaller $\omega_i$ and remain primarily constrained by the rotated IoU loss.

\subsection{Model Variants and Deployment}
\label{sec:model_variants_and_deployment}

YOLO26 provides a unified model family spanning five size variants (n, s, m, l, x) across multiple computer vision tasks: object detection, instance segmentation, image classification, pose estimation, and oriented object detection~\cite{ultralytics2026_tasks_docs}. Each variant supports training, validation, inference, and native PyTorch checkpoints~\cite{ultralytics2026_train_docs,ultralytics2026_val_docs,ultralytics2026_predict_docs}. For deployment, Ultralytics supports 19 export targets beyond PyTorch: TorchScript, ONNX, OpenVINO, TensorRT, CoreML, TensorFlow SavedModel, TensorFlow GraphDef, TensorFlow Lite, TensorFlow Edge TPU, TensorFlow.js, PaddlePaddle, MNN, NCNN, IMX, RKNN, ExecuTorch, Axelera AI, DEEPX, and Qualcomm QNN~\cite{ultralytics2026_export_docs}.

\begin{figure*}[t]
    \centering
    \includegraphics[width=0.8\textwidth]{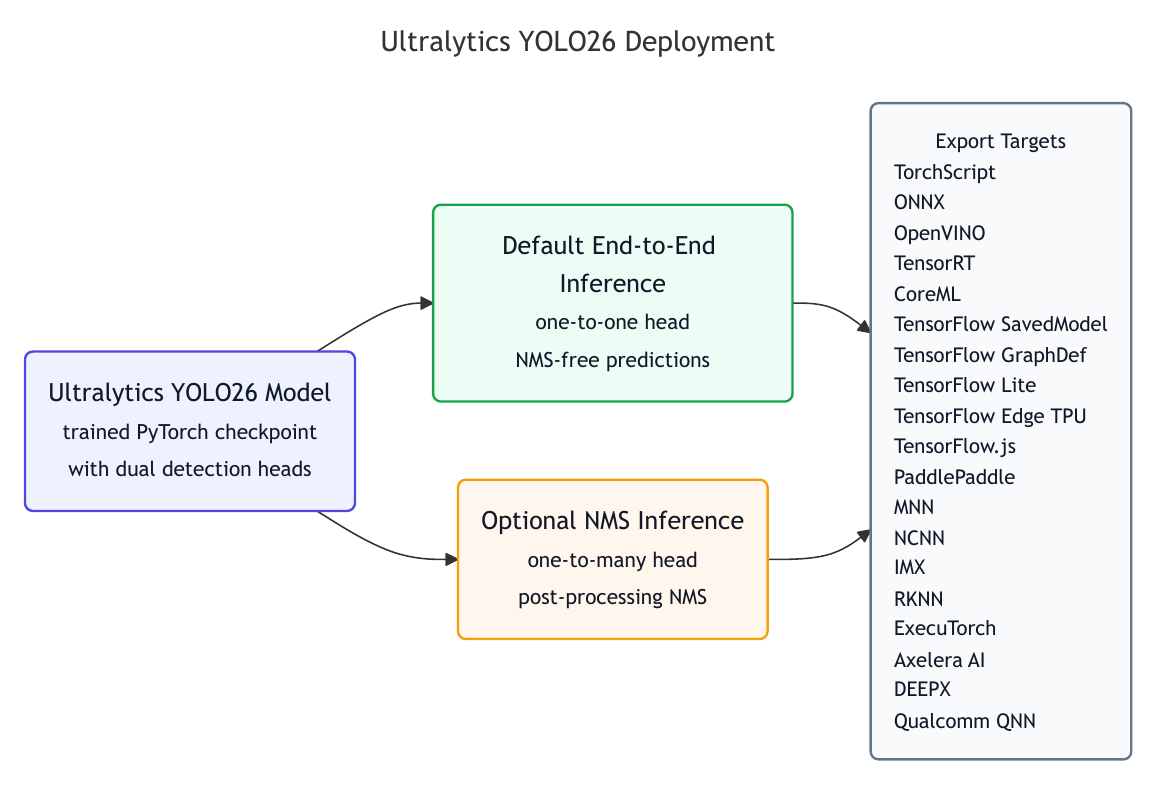}
    \caption{Deployment pipeline of Ultralytics YOLO26. A trained model supports two inference paths: the default one-to-one NMS-free path and the optional one-to-many path with NMS, while preserving compatibility with a broad set of export targets.}
    \label{fig:yolo26_deployment_pipeline}
\end{figure*}

Figure~\ref{fig:yolo26_deployment_pipeline} illustrates the deployment side of the framework. After training, the same YOLO26 model can be executed through the default one-to-one NMS-free path or the optional one-to-many path with NMS, while preserving compatibility with standard export targets. Some runtimes that do not support the top-$K$ operations required by end-to-end decoding automatically fall back to the non-end-to-end branch during export. This separation between training and deployment emphasizes that YOLO26 is designed not only for strong optimization behavior but also for practical inference integration across heterogeneous runtimes.

The architecture achieves up to 43\% faster CPU inference compared to previous YOLO versions (YOLO26n vs. YOLO11n in ONNX format, benchmarked on an Intel Xeon CPU @ 2.00\,GHz), with reduced model size and memory footprint, making it particularly suitable for edge deployment scenarios where GPU acceleration is unavailable or impractical.

\subsection{YOLOE-26: Open-Vocabulary Detection and Segmentation}
\label{sec:yoloe26}
YOLOE~\cite{yoloe} extends the YOLO framework with
embedding-based classification to support
text-prompted, visual-prompted, and prompt-free
inference within a single model. For text prompts,
the Re-parameterizable Region-Text Alignment
(RepRTA) strategy~\cite{yoloe} uses a
MobileCLIP~\cite{mobileclip} text encoder and a
lightweight auxiliary network to align region
features with text embeddings. For visual prompts,
the Spatial-Aware Visual Prompt Embedding module
(\texttt{SAVPE})~\cite{yoloe} produces visual
embeddings through decoupled semantic and activation
branches. For prompt-free inference, the Lightweight
Region Proposal and Classification head
(\texttt{LRPCHead})~\cite{yoloe} leverages a built-in
vocabulary and dedicated feature embeddings to detect
generic objects without requiring a language model at
inference time.\\
In the current implementation, YOLOE and YOLOE-26
instantiate the \texttt{BNContrastiveHead}.
Let $\mathbf{F}_l$ denote the detection feature map
at pyramid level $l$. YOLOE-26 predicts
box-branch outputs and prompt-conditioned
classification scores through separate heads:
\begin{equation}
  \mathbf{B}_l = \mathcal{B}_l(\mathbf{F}_l), \qquad
  \mathbf{Z}_l = \mathcal{E}_l(\mathbf{F}_l),
\end{equation}
where $\mathbf{B}_l$ denotes the box-branch outputs
and $\mathbf{Z}_l$ denotes the classification
embeddings. Given normalized prompt embeddings
$\mathbf{W} \in \mathbb{R}^{B \times K \times C}$
for $K$ categories, the score map is
\begin{equation}
  \mathbf{S}_l = \exp(\tau)\cdot\bigl(\mathrm{BN}(\mathbf{Z}_l)
  \otimes \mathbf{W}\bigr) + b,
  \label{eq:bn-head}
\end{equation}
where
$\mathbf{S}_l \in \mathbb{R}^{B \times K \times H \times W}$, which serves as the open-vocabulary equivalent
of the classification logits in standard YOLO detectors,
and is subsequently passed through a sigmoid function
to produce the final per-class confidence scores. $\mathrm{BN}(\cdot)$ applies batch normalization to
 classification embeddings, $\otimes$ denotes
the inner product of the channel category across spatial
locations, and $\tau$, $b$ are learnable scalars
initialized to $-1$ and $-10$, respectively. The
box and classification branches remain aligned by
shared spatial indices, so each location on the level
$l$ produces a paired box prediction and class-score
vector. YOLOE is trained in three stages: a
text-prompt stage (TP), a visual-prompt stage (VP) 
and a prompt-free stage (PF). The TP stage serves
as the shared initialization stage, and the VP and
PF models are each fine-tuned independently from
the TP checkpoint. \\
YOLOE-26 instantiates the open-vocabulary formulation of YOLOE~\cite{yoloe} on top of the YOLO26 detector family. It retains the same three inference modes while introducing four modifications.
\textbf{(1) Backbone upgrade.} The original YOLO11-based detector is replaced by the YOLO26 backbone, neck, and end-to-end detection stack. In practice, the TP model is initialized from released pretrained YOLO26 detector weights, transferring the closed-set design improvements described above to the open-vocabulary setting without altering the prompting interface.
\textbf{(2) Text encoder upgrade.} The MobileCLIP~\cite{mobileclip} text encoder is upgraded to MobileCLIP2~\cite{mobileclip2}, yielding stronger text--visual alignment.
\textbf{(3) Data engine.} Upon visualizing the training set, we observed that a substantial number of objects are left unannotated in the original annotations, yet can be reliably detected by a trained YOLOE model. Motivated by this observation, we employ a pretrained YOLOE teacher, prompted with a built-in vocabulary of 4585 classes~\cite{huang2025open}, to refine the training set. Predicted boxes that are absent from the original annotations, have $\mathrm{IoU} < 0.5$ with any existing ground-truth box, and exceed a confidence threshold of 0.5 are appended as pseudo-labels for a second round of training.
\textbf{(4) Decoupled segmentation training.} The segmentation head is disabled during the initial text-prompt training stage and is trained in a separate subsequent stage from the TP checkpoint, allowing the backbone to focus on open-vocabulary detection. This contrasts with the original YOLOE, which trains the segmentation head jointly with the text-prompt branch.

\section{Experiments}

YOLO26 is evaluated on standard object detection,
instance segmentation, pose estimation, oriented
bounding box detection, and open-vocabulary detection.
Implementation details and ablation studies are
presented first, followed by the main COCO detection
results, the task-specific results, and the YOLOE-26
results.

\subsection{Implementation Details}
\label{sec:implementation_details}

Unless otherwise stated, YOLO26 detection models are trained in end-to-end mode with a direct regression head (\texttt{reg\_max}{=}1). For the reported COCO detection benchmarks, we use a two-stage training recipe: all model sizes are first pretrained on Objects365-v1~\cite{shao2019objects365} for 150 epochs, and are then fine-tuned on COCO. The COCO fine-tuning schedule is model-size dependent, using 245/70/80/60/40 epochs for the n/s/m/l/x variants, respectively. The global batch size is 128 across model scales in both stages. Exact per-size optimizer, schedule, loss, augmentation, and checkpoint-recorded internal settings for the Objects365-v1 pretraining stage and COCO fine-tuning stage are summarized in Sec.~\ref{sec:suppl_training_recipes} of the supplementary materials and Tables~\ref{tab:suppl_pretrain_recipe_optimizer}--\ref{tab:suppl_training_recipe_internal}.

Across both Objects365-v1 pretraining and COCO fine-tuning, we adopt a model-size-aware augmentation policy. In both stages, larger models are regularized with stronger scale, mixup, and copy-paste augmentation overall, while YOLO26n uses the mildest recipe. Mosaic is used heavily during most of training in both stages and is disabled near the end of training through the \texttt{close\_mosaic} schedule. The exact Objects365-v1 pretraining and COCO fine-tuning augmentation settings are provided in Sec.~\ref{sec:suppl_training_recipes} of the supplementary materials and Tables~\ref{tab:suppl_pretrain_recipe_aug} and~\ref{tab:suppl_training_recipe_aug}, respectively.

The Progressive Loss weights are initialized to emphasize the one-to-many branch in the early stage and are then shifted toward the one-to-one branch over training. In the current implementation, the one-to-many and one-to-one weights are initialized as $(0.8, 0.2)$ and linearly updated over training to $(0.1, 0.9)$, respectively. The update is applied once per training epoch. We keep these values fixed across experiments unless otherwise noted, and treat them as implementation choices rather than independently tuned hyperparameters.

For STAL, the current implementation uses the default three-level detection pyramid with strides $[8,16,32]$. Accordingly, the smallest stride is $s_{\min}=8$, and the reference size is set to the next stride level, $s_{\mathrm{ref}}=16$. In practice, this means that during candidate filtering only, any ground-truth width or height below 8 pixels is clamped to 16 pixels, while the original box remains unchanged for subsequent matching and regression.

\subsection{Component-Wise Ablation}
\label{sec:componentwise_ablation}

\begin{table*}[t]
\centering
\small
\setlength{\tabcolsep}{6pt}
\begin{tabular}{@{}lcccccc@{}}
\toprule
\textbf{Model} & \textbf{AP (E2E)} & \textbf{AP (Non-E2E)} & \textbf{Params (M)} & \textbf{FLOPs (G)} & \textbf{Latency (ms)} \\
\midrule
YOLO11s (Baseline) & -- & 47.0 & 9.4 & 21.5 & 2.5 \\
$-$ DFL & -- & 46.4 & 9.1 & 20.1 & 2.3 \\
+ L1 Loss & -- & 46.6 & 9.1 & 20.1 & 2.3 \\
+ STAL & -- & 46.8 & 9.1 & 20.1 & 2.3 \\
+ Backbone/neck refinement & -- & 47.0 & 9.5 & 20.7 & 2.5 \\
+ E2E & 46.4 & 47.0 & 9.5 & 20.7 & 2.5 \\
+ Progressive Loss & 46.7 & 47.2 & 9.5 & 20.7 & 2.5 \\
+ MuSGD & 47.1 & 47.6 & 9.5 & 20.7 & 2.5 \\
+ Objects365 Pretrained & 47.4 & 48.0 & 9.5 & 20.7 & 2.5 \\
+ Hyperparameter Search & 47.8 & 48.6 & 9.5 & 20.7 & 2.5 \\
\bottomrule
\end{tabular}
\vspace{2pt}
\caption{Step-by-step ablation from the YOLO11s baseline to the final YOLO26s configuration. Replacing DFL with direct box regression and L1 loss reduces complexity with limited accuracy change, STAL and the backbone/neck refinement recover and improve the NMS-based AP, and end-to-end training with Progressive Loss improves the one-to-one branch. MuSGD, Objects365 pretraining, and hyperparameter search provide the remaining gains to the final result.}
\label{tab:componentwise_ablation}
\end{table*}

Table~\ref{tab:componentwise_ablation} summarizes the incremental evolution from the YOLO11 baseline to the final YOLO26s configuration. Removing DFL and replacing it with direct box regression plus L1 loss largely preserves accuracy while reducing model complexity, and STAL then recovers additional performance with improved small-object assignment. We further apply a light backbone/neck refinement by inserting one additional attention layer in the detection neck, which improves accuracy while keeping latency essentially unchanged. Enabling end-to-end training with Progressive Loss then improves the one-to-one branch, while MuSGD, Objects365 pretraining, and hyperparameter search provide further gains to reach the final result.

\subsection{Core Design Ablations}
\label{sec:core_design_ablations}

In this subsection, we isolate the main design choices that distinguish YOLO26 from the baseline formulation. The following ablations evaluate the effects of DFL removal, MuSGD, Progressive Loss, and STAL, showing how each component contributes to optimization behavior, accuracy, and deployment-oriented design.

\subsubsection{DFL Removal}
\label{sec:dfl_removal_ablation}

\begin{table}[t]
\centering
\small
\setlength{\tabcolsep}{5pt}
\begin{tabular}{cccccc}
\toprule
\textbf{Res.} & \textbf{DFL} & \textbf{AP} $\uparrow$ & \textbf{AP$_S$} & \textbf{AP$_M$} & \textbf{AP$_L$} \\
\midrule
640 & w/ & 46.0 & 27.3 & 50.4 & 62.8 \\
640 & w/o & \textbf{46.3} & \textbf{27.9} & \textbf{50.6} & \textbf{63.8} \\
\midrule
1280 & w/ & 49.8 & \textbf{36.0} & \textbf{55.7} & 61.8 \\
1280 & w/o & \textbf{51.1} & 35.9 & 55.2 & \textbf{64.0} \\
\bottomrule
\end{tabular}
\vspace{2pt}
\caption{Controlled DFL removal ablation on COCO using YOLO26s at 640 and 1280 resolution. Removing DFL improves AP at both resolutions, with the gain increasing from +0.3 AP / +1.0 AP$_L$ at 640 to +1.3 AP / +2.2 AP$_L$ at 1280.}
\label{tab:dfl_removal_ablation}
\end{table}

Table~\ref{tab:componentwise_ablation} shows that removing DFL alone costs 0.6~AP on the YOLO11s baseline (47.0 to 46.4), but this gap is fully recovered by L1 supervision (+0.2), STAL (+0.2), and the backbone/neck refinement (+0.2), yielding a strictly lighter and faster head at the same accuracy ($-$0.3M parameters, $-$1.4~GFLOPs, $-$0.2~ms latency). To verify the effect more directly, we conduct controlled 640 and 1280 ablations using YOLO26s under a matched COCO training protocol, training the models with and without DFL independently from scratch under identical settings except for image size.

\begin{figure}[t]
\centering
\begin{subfigure}[b]{0.95\columnwidth}
  \includegraphics[width=\textwidth]{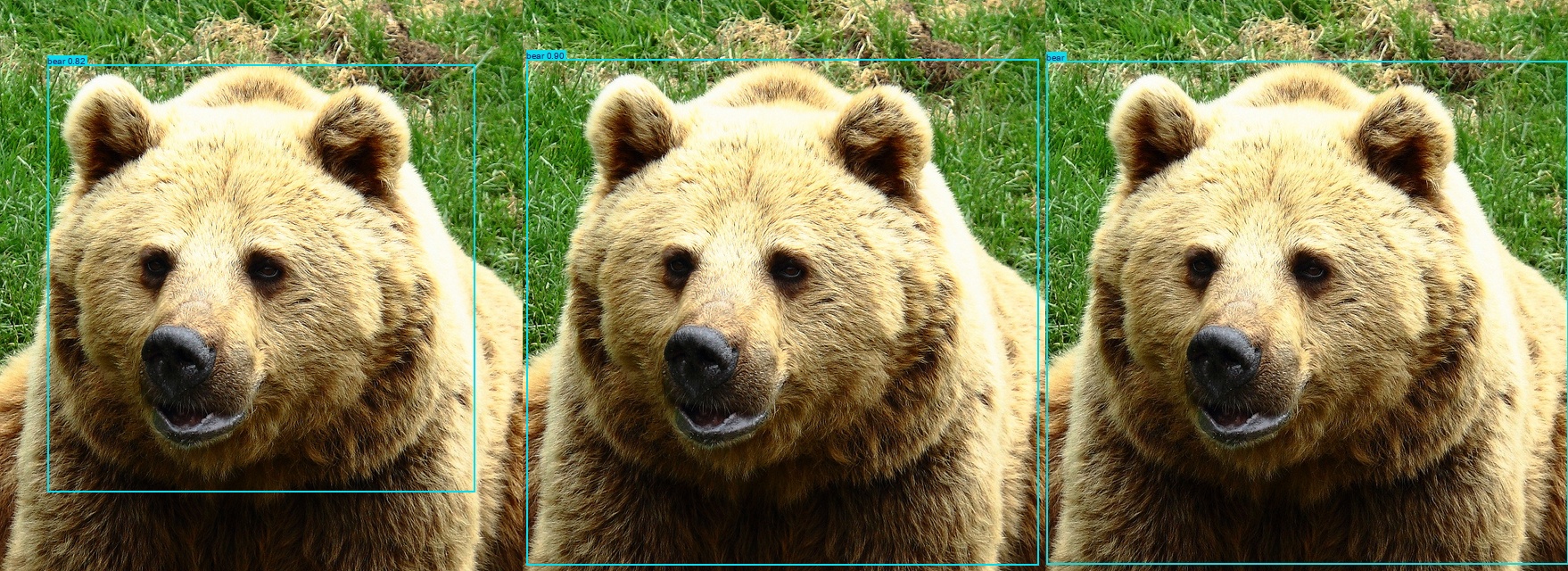}
\end{subfigure}
\\[4pt]
\begin{subfigure}[b]{0.95\columnwidth}
  \includegraphics[width=\textwidth]{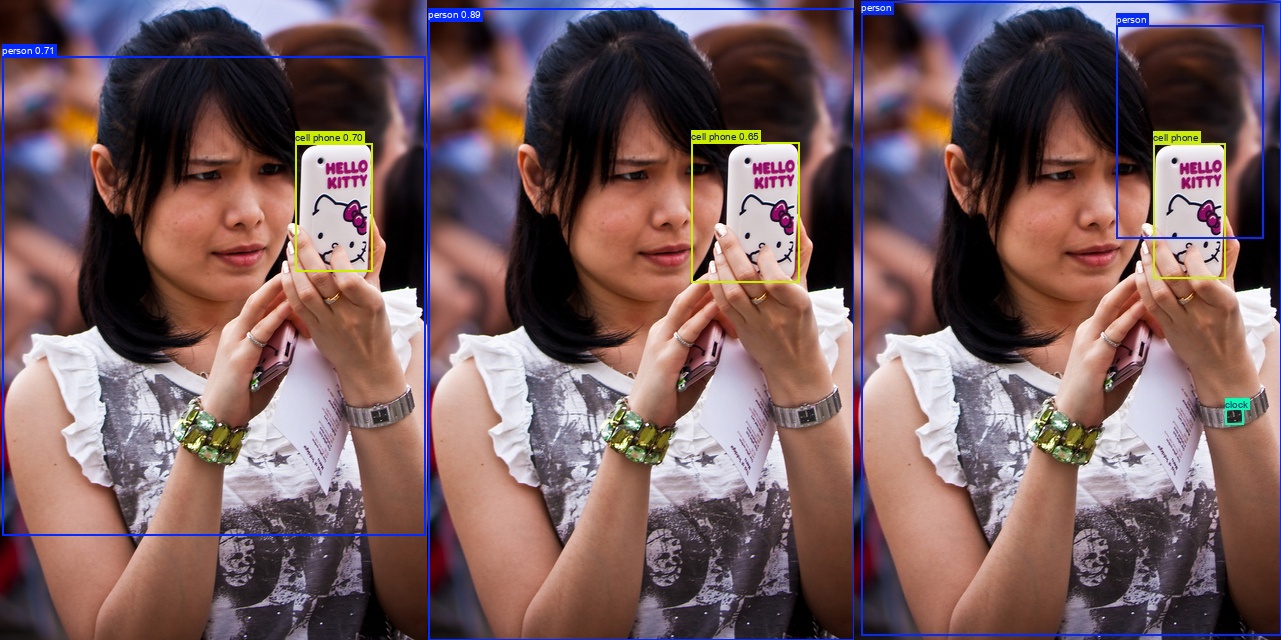}
\end{subfigure}
\\[4pt]
\begin{subfigure}[b]{0.95\columnwidth}
  \includegraphics[width=\textwidth]{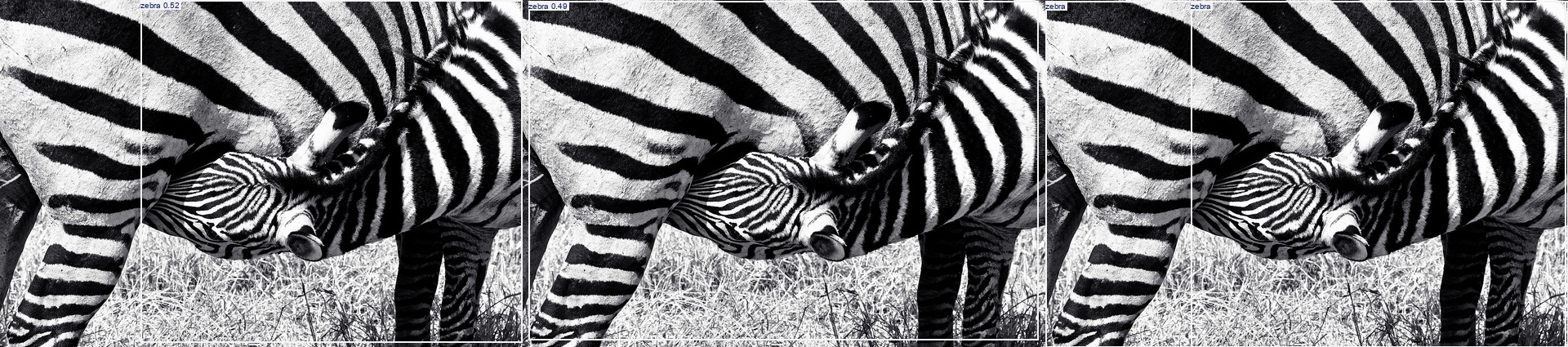}
\end{subfigure}
\\[4pt]
\begin{subfigure}[b]{0.95\columnwidth}
  \includegraphics[width=\textwidth]{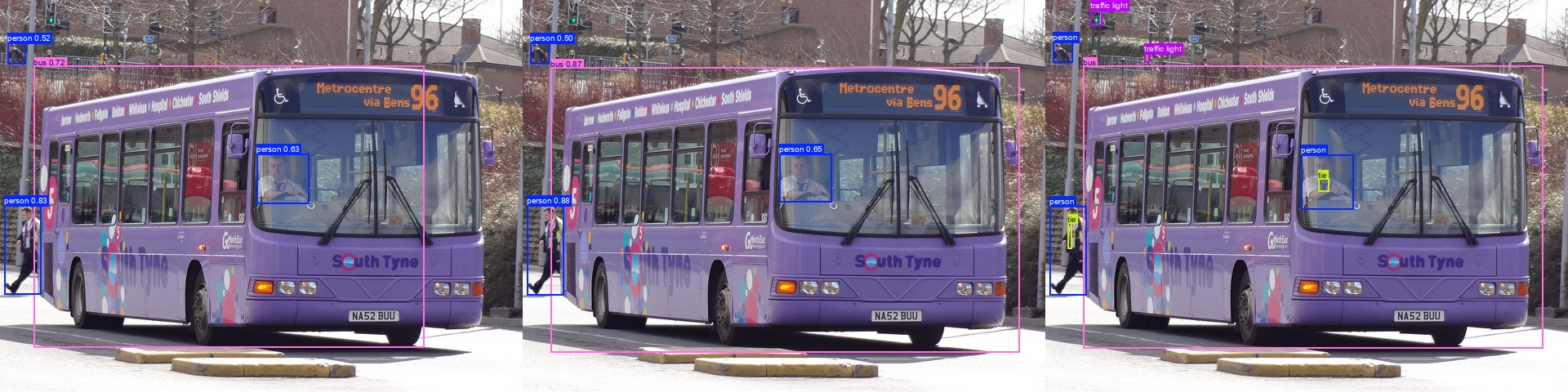}
\end{subfigure}
\\[4pt]
\begin{subfigure}[b]{0.95\columnwidth}
  \includegraphics[width=\textwidth]{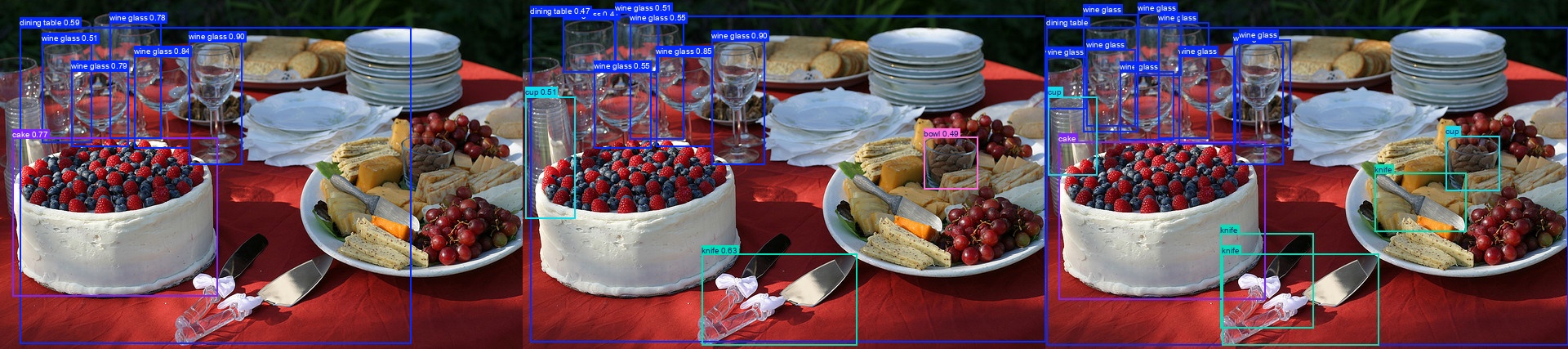}
\end{subfigure}

\caption{Qualitative comparison for large-object localization at 1280 resolution. From left to right, each row shows predictions from the model with DFL, predictions from the counterpart without DFL, and the ground-truth annotations. The DFL-free head better preserves full-object extent on large targets, supporting the quantitative gains reported in Table~\ref{tab:dfl_removal_ablation}.}
\label{fig:dfl_large_qualitative}
\end{figure}

Table~\ref{tab:dfl_removal_ablation} confirms that, in the YOLO26s configuration, the DFL-free head improves AP at both 640 and 1280 resolution, and that the benefit becomes more pronounced at the higher resolution. Even at 640, removing DFL improves AP and AP$_L$, indicating that the finite support of DFL with \texttt{reg\_max}{=}16 already restricts regression quality for the largest targets. The effect becomes stronger at 1280, where larger-object regression spans longer distances and therefore exposes this range limitation more clearly. In particular, the AP$_L$ gain grows from +1.0 at 640 to +2.2 at 1280, supporting the claim that direct regression becomes more favorable as resolution increases. Together, the two tables indicate that DFL's localization benefit is replaceable by simpler components, while its finite-range cost becomes more limiting at higher resolution. Figure~\ref{fig:dfl_large_qualitative} further shows that the DFL-free head better preserves the full extent of large objects at 1280 resolution.

\subsubsection{MuSGD}
\label{sec:musgd_ablation}

\begin{table}[t]
\centering
\small
\setlength{\tabcolsep}{8pt}
\begin{tabular}{lcc}
\toprule
\textbf{Optimizer} & \textbf{Epochs} $\downarrow$ & \textbf{COCO mAP} $\uparrow$ \\
\midrule
SGD  & 600 & 47.0 \\
MuSGD & 500 & \textbf{47.4} \\
\bottomrule
\end{tabular}
\vspace{2pt}
\caption{MuSGD improves convergence speed and final accuracy when training YOLO26 from scratch on COCO. MuSGD reaches 47.4 mAP in 500 epochs versus 47.0 mAP for SGD in 600 epochs (16.7\% fewer epochs).}
\label{tab:musgd_vs_sgd}
\end{table}

We compare MuSGD against standard SGD when training YOLO26 from scratch on COCO. Table~\ref{tab:musgd_vs_sgd} shows that MuSGD reaches a higher final accuracy with a shorter schedule, improving mAP by +0.4 while reducing training from 600 to 500 epochs. This result supports MuSGD as a practical optimization improvement for the main detection setting.

To further validate MuSGD beyond detection, we also perform a controlled ImageNet classification comparison in which the backbone architecture and training recipe are held fixed and only the optimizer differs. The detailed results are provided in Sec.~\ref{sec:suppl_musgd_cls} of the supplementary materials and Table~\ref{tab:suppl_musgd_cls}; the same trend holds there, indicating that the MuSGD advantage transfers beyond the detection setting.

\subsubsection{Progressive Loss}
\label{sec:progloss_ablation}

\begin{table}[t]
\centering
\small
\setlength{\tabcolsep}{4pt}
\begin{tabular}{cccc}
\toprule
\textbf{Start (o2m, o2o)} & \textbf{End (o2m, o2o)} & \textbf{mAP (E2E)} $\uparrow$ \\
\midrule
(0.5, 0.5) & (0.5, 0.5) & 46.4 \\
\midrule
(1.0, 0.0) & (0.1, 0.9) & 46.4 \\
(1.0, 0.0) & (0.2, 0.8) & 46.4 \\
(1.0, 0.0) & (0.3, 0.7) & 46.3 \\
(0.8, 0.2) & (0.1, 0.9) & \textbf{46.7} \\
(0.9, 0.1) & (0.1, 0.9) & 46.3 \\
\bottomrule
\end{tabular}
\vspace{2pt}
\caption{Ablation of Progressive Loss schedules on COCO using YOLO11s as the baseline. The default schedule $(0.8, 0.2)\rightarrow(0.1, 0.9)$ gives the best end-to-end mAP.}
\label{tab:progloss_ablation}
\end{table}

Table~\ref{tab:progloss_ablation} studies how the weighting between the one-to-many and one-to-one branches should evolve during training. The fixed equal-weight baseline reaches 46.4 E2E AP, while the best scheduled variant, $(0.8, 0.2)\rightarrow(0.1, 0.9)$, improves this to 46.7. Starting from $(1.0, 0.0)$ underperforms consistently, indicating that fully suppressing the one-to-one branch early in training is suboptimal. A near one-to-many-dominated start, such as $(0.9, 0.1)$, also hurts performance. Overall, the best schedule is the one that gives the one-to-one branch nonzero supervision from the beginning and then gradually emphasizes it later.

\subsubsection{STAL}
\label{sec:stal_ablation}

\begin{table}[t]
\centering
\small
\setlength{\tabcolsep}{6pt}
\begin{tabular}{lccccc}
\toprule
\textbf{Method} & \textbf{$s_{\mathrm{ref}}$} & \textbf{AP} $\uparrow$ & \textbf{AP$_S$} & \textbf{AP$_M$} & \textbf{AP$_L$} \\
\midrule
TAL (baseline) & -- & 46.6 & 29.0 & 51.4 & \textbf{63.9} \\
\midrule
STAL & 8  & 46.6 & 27.7 & 51.6 & 63.8 \\
STAL & 16 & \textbf{46.8} & \textbf{29.6} & \textbf{51.6} & 63.8 \\
STAL & 32 & 46.5 & 28.3 & 51.3 & 63.7 \\
\bottomrule
\end{tabular}
\vspace{2pt}
\caption{Ablation of the STAL reference size $s_{\mathrm{ref}}$ on COCO using YOLO11s as the baseline. Vanilla TAL achieves 46.6 AP. STAL with $s_{\mathrm{ref}}=16$ yields the best overall result at 46.8 AP (+0.2), with notable gains in AP$_S$ (29.6 vs.\ 29.0). Both $s_{\mathrm{ref}}=8$ and $s_{\mathrm{ref}}=32$ match or slightly underperform the baseline, confirming that the next-stride reference size is the most effective choice.}
\label{tab:stal_ablation}
\end{table}

\begin{figure}[t]
\centering
\begin{subfigure}[b]{0.95\columnwidth}
  \includegraphics[width=\textwidth]{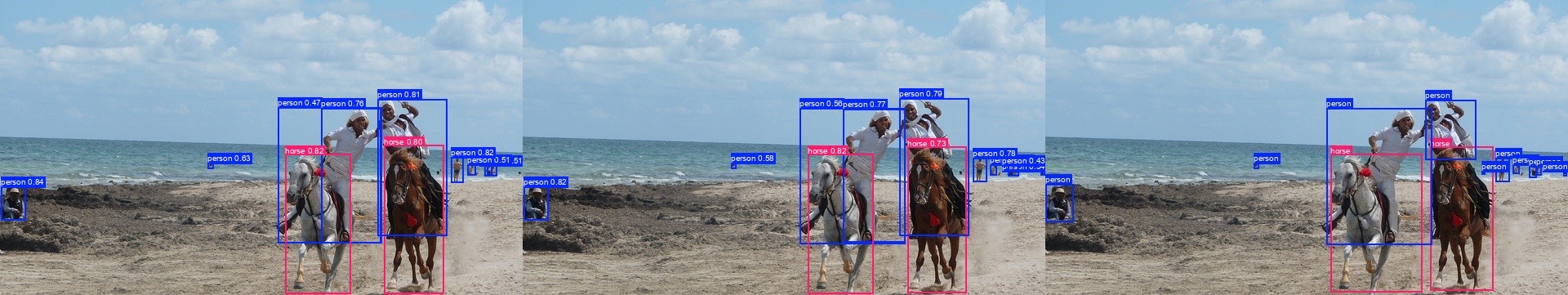}
\end{subfigure}
\\[4pt]
\begin{subfigure}[b]{0.95\columnwidth}
  \includegraphics[width=\textwidth]{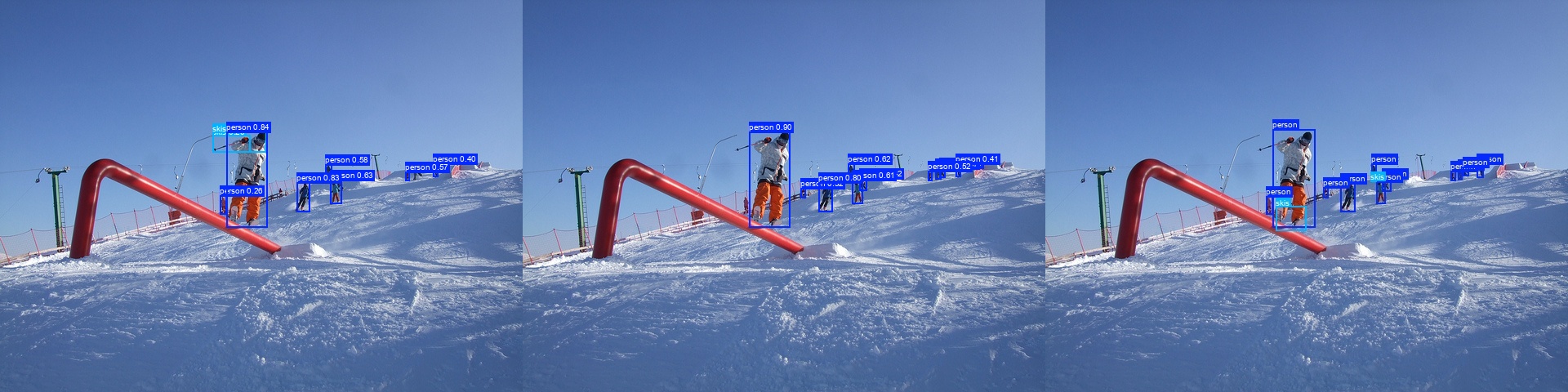}
\end{subfigure}
\\[4pt]
\begin{subfigure}[b]{0.95\columnwidth}
  \includegraphics[width=\textwidth]{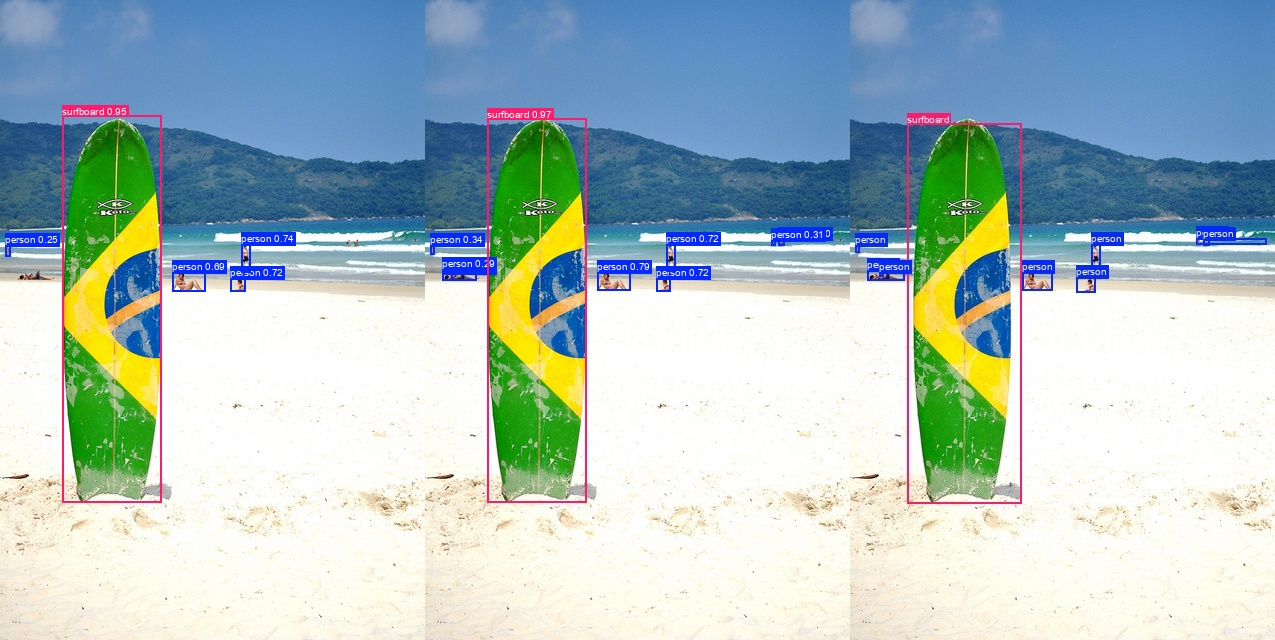}
\end{subfigure}
\\[4pt]
\begin{subfigure}[b]{0.95\columnwidth}
  \includegraphics[width=\textwidth]{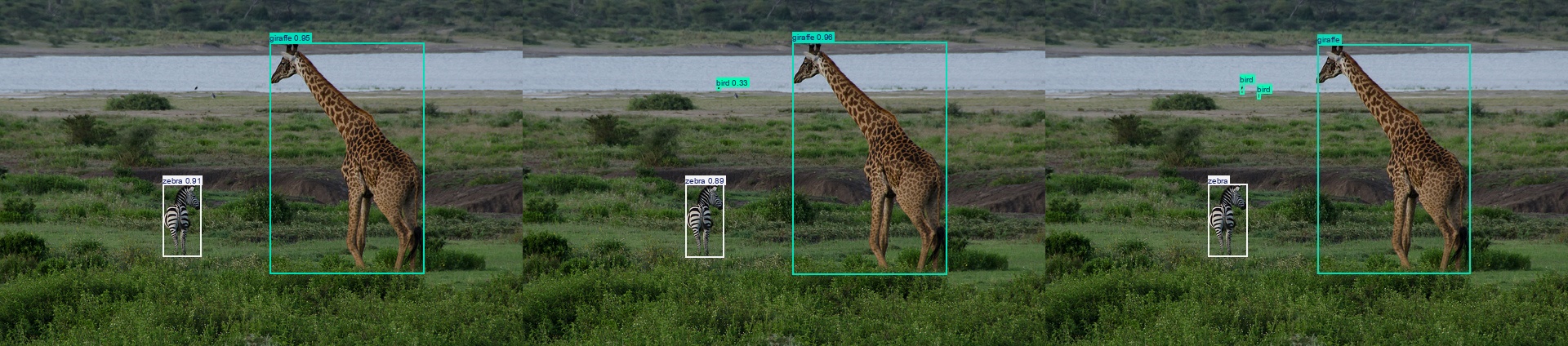}
\end{subfigure}
\\[4pt]
\begin{subfigure}[b]{0.95\columnwidth}
  \includegraphics[width=\textwidth]{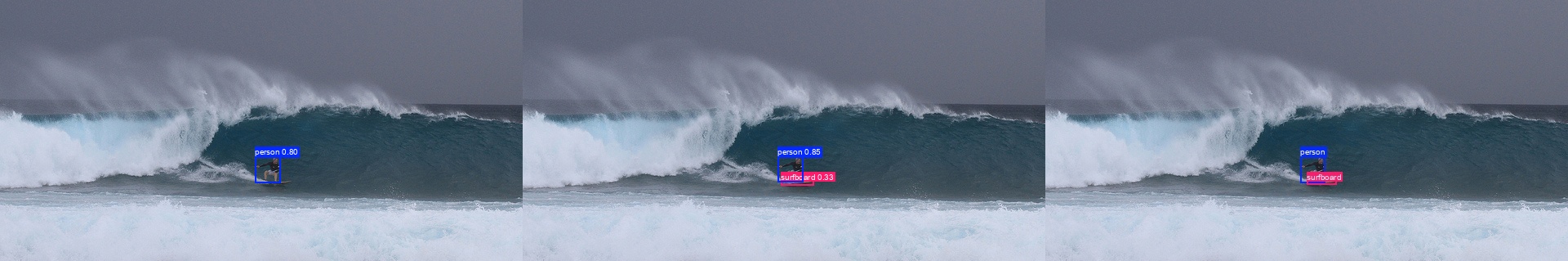}
\end{subfigure}

\caption{Qualitative comparison between TAL baseline (left), STAL with $s_{\mathrm{ref}}=16$ (middle), and ground-truth annotations (right) on COCO validation images. All predictions are generated at a confidence threshold of 0.25. STAL improves small-object detection by ensuring sufficient anchor coverage for tiny ground-truth boxes, leading to fewer missed detections and tighter localization that more closely matches the ground truth.}
\label{fig:stal_qualitative}
\end{figure}

Table~\ref{tab:stal_ablation} evaluates the reference size $s_{\mathrm{ref}}$ in STAL against vanilla TAL on COCO with YOLO11s as the baseline. Choosing the next stride level, $s_{\mathrm{ref}}=16$, gives the best overall result at 46.8 AP, improving the baseline by +0.2 AP and increasing AP$_S$ from 29.0 to 29.6. Setting $s_{\mathrm{ref}}=8$ does not improve overall AP and noticeably reduces AP$_S$, which suggests that the adjustment is too weak to stabilize assignment for very small objects. Increasing the reference size further to 32 also degrades performance, indicating that excessive enlargement begins to distort the intended scale prior. Overall, these results support the default choice of $s_{\mathrm{ref}}=16$ in Section~\ref{sec:implementation_details}. Figure~\ref{fig:stal_qualitative} further shows that STAL recovers small-object detections that the TAL baseline misses or localizes poorly.

\subsection{Main Detection Results on COCO}
\label{sec:main_coco_detection}

\begin{table*}[t]
\centering
\small
\setlength{\tabcolsep}{6pt}
\begin{tabular}{lccccccc}
\toprule
\textbf{Model} & \textbf{Size} & \textbf{mAP} & \textbf{mAP (E2E)} & \textbf{CPU ONNX} & \textbf{T4 TRT10} & \textbf{Params} & \textbf{FLOPs} \\
 & \textbf{(px)} & \textbf{val 50--95} & \textbf{val 50--95} & \textbf{(ms)} & \textbf{(ms)} & \textbf{(M)} & \textbf{(B)} \\
\midrule
YOLO26n & 640 & 40.9 & 40.1 & 38.9 & 1.7 & 2.4 & 5.4 \\
YOLO26s & 640 & 48.6 & 47.8 & 87.2 & 2.5 & 9.5 & 20.7 \\
YOLO26m & 640 & 53.1 & 52.5 & 220.0 & 4.7 & 20.4 & 68.2 \\
YOLO26l & 640 & 55.0 & 54.4 & 286.2 & 6.2 & 24.8 & 86.4 \\
YOLO26x & 640 & 57.5 & 56.9 & 525.8 & 11.8 & 55.7 & 193.9 \\
\bottomrule
\end{tabular}
\vspace{2pt}
\caption{Released YOLO26 detection benchmarks on COCO from the Ultralytics repository~\cite{ultralytics2026yolo26_docs}. `mAP` denotes standard validation of the one-to-many branch with NMS (`end2end=False`), whereas `mAP (E2E)` denotes true end-to-end validation of the default one-to-one branch. Speed values are the published CPU ONNX (Intel Xeon CPU @ 2.00\,GHz) and T4 TensorRT10 benchmarks for the released models.}
\label{tab:yolo26_coco_detection}
\end{table*}

Table~\ref{tab:yolo26_coco_detection} summarizes the released YOLO26 detection results on COCO~\cite{ultralytics2026yolo26_docs}. We report both the standard validation mAP and the end-to-end mAP because YOLO26 supports two inference paths. The non-E2E numbers correspond to the one-to-many branch evaluated with NMS, while the E2E numbers correspond to the one-to-one branch used for true end-to-end inference without NMS.

\paragraph{End-to-End vs. NMS Flexibility.} This distinction is practically important. The one-to-one head provides a simpler end-to-end deployment path and remains close to the NMS-based variant, trailing by only 0.6--0.8 AP across model scales. At the same time, YOLO26 does not force a single deployment mode: if the target platform or inference stack can execute NMS efficiently, the one-to-many head can still be preferred when the highest possible AP is the priority. Conversely, when deployment simplicity, tighter integration, or NMS-free inference is more valuable, the default one-to-one path provides a cleaner alternative.

\paragraph{Comparison with Recent Real-Time Detectors.} A full grouped s/m/l/x comparison with recent real-time detectors is provided in Sec.~\ref{sec:suppl_literature_comparison} of the supplementary materials and Table~\ref{tab:suppl_selected_literature_comparison}. At the standard NMS operating point, YOLO26 provides the strongest overall AP--latency trade-off in this comparison, achieving the best AP in the medium, large, and extra-large groups while remaining competitive in latency.

\subsection{Results on Various Vision Tasks}
\label{sec:task_results}
To show the robustness of our proposed methods, we present YOLO26 models on various vision tasks. Apart from task-specific optimization strategies, all other training and testing settings are kept consistent with those used for the detection models. 

\subsubsection{Instance Segmentation}
\begin{table}[t]
\centering
\small
\setlength{\tabcolsep}{4pt}
\begin{tabular}{lcc}
\toprule
\textbf{Method} & \textbf{mAP} & $\mathbf{AP}^{50}$ \\
\midrule
YOLO11s (baseline) & 32.0 & 51.1 \\
+ multi-scale proto module & 32.4 & 51.7 \\
+ auxiliary loss & 32.7 & 52.0  \\
\bottomrule
\end{tabular}
\vspace{2pt}
\caption{Effectiveness of module modification and auxiliary loss for instance segmentation on COCO.}
\label{tab:ins_seg_ablation}
\end{table}

As mentioned in Sec.~\ref{sec:ins_seg}, we adopt the multi-scale Proto Module and auxiliary semantic segmentation loss for instance segmentation.
We evaluate the YOLO11s segmentation model on the COCO instance segmentation dataset to investigate the effect of each proposed method. As shown in Table~\ref{tab:ins_seg_ablation}, the multi-scale Proto Module improves the mAP from 32.0\% to 32.4\%, indicating that embedding higher-level semantic concepts into prototype maps can improve mask quality. We use equal BCE and Dice weights for the auxiliary loss, improving accuracy from 32.4\% to 32.7\% without compromising inference speed.

We integrate the proposed methods into YOLO26 to build
the new segmentation models. Following the training
policy of YOLO26 detection models, we adopt the
Objects365-v1 pretrained weights and fine-tune on the
COCO instance segmentation dataset. The full YOLO11 and
YOLO26 family comparison is provided in
Sec.~\ref{sec:suppl_task_benchmarks} of the
supplementary materials and
Table~\ref{tab:yolo26_coco_ins_seg}. At the standard
NMS operating point, YOLO26 improves box AP by
\textbf{+1.6 to +2.5} and mask AP by \textbf{+2.4 to
+3.7} over YOLO11 across scales, while the end-to-end
path remains close to the NMS-based variant.

\subsubsection{Pose Estimation}

\begin{table}[ht]
\centering
\small
\setlength{\tabcolsep}{18pt}
\begin{tabular}{lccc}
\toprule
\textbf{$w_{OKS}$} & \textbf{$w_{RLE}$} & \textbf{mAP} \\
\midrule
48  & 0 & 61.5 \\
48  & 1 & 60.8 \\
\textbf{24}  & \textbf{1} & \textbf{63.0} \\
24  & 2 & 62.5 \\
12  & 1 & 62.6 \\
0   & 1 & 61.9 \\
0   & 2 & 62.4 \\
\bottomrule
\end{tabular}
\vspace{2pt}
\caption{Ablation study on different weight configurations of the OKS and RLE loss, evaluated with YOLO26s pose model on the COCO keypoints validation set.}
\label{tab:pose_loss_ablation}
\end{table}

Pose estimation requires precise localization of
keypoints under varying scales and appearance changes,
and top-down pipelines remain sensitive to the quality
of the underlying detector. We therefore use a
combination of RLE loss and OKS loss while keeping the
other training settings aligned with the detection
models; see Sec.~\ref{sec:pose_estimation} for the
loss formulation. Table~\ref{tab:pose_loss_ablation}
evaluates different loss weight settings on YOLO26s
and shows that $w_{OKS}=24$ and $w_{RLE}=1$ gives the
best mAP.
The full YOLO11 and YOLO26 family comparison is
provided in Sec.~\ref{sec:suppl_task_benchmarks} of the
supplementary materials and
Table~\ref{tab:yolo26_coco_pose}. In the E2E setting,
YOLO26 improves pose AP by \textbf{+2.1 to +7.2} over
YOLO11 across scales, while the end-to-end and
NMS-based variants remain nearly equivalent, differing
by at most 0.2 AP.

\subsubsection{Oriented Bounding Box Detection}

\begin{table}[t]
\centering
\small
\setlength{\tabcolsep}{4pt}
\begin{tabular}{lcc}
\toprule
\textbf{Angle definition} & \textbf{mAP} & $\mathbf{AP}^{50}$ \\
\midrule
$(0, 90^\circ]$      & 47.7 & 75.0 \\
$[-45^\circ, 135^\circ)$  & \textbf{49.0} & \textbf{75.4} \\
\bottomrule
\end{tabular}
\vspace{2pt}
\caption{Comparison of different angle definitions for OBB detection on DOTA-v1.0 validation set, using YOLO26s model without the angle loss as the backbone.}
\label{tab:obb_angle_ablation}
\end{table}

\begin{table}[ht]
\centering
\small
\setlength{\tabcolsep}{18pt}
\begin{tabular}{lcc}
\toprule
\textbf{$\lambda$} & \textbf{mAP} & $\mathbf{AP}^{50}$ \\
\midrule
---  & 49.0 & 75.4 \\
1  & 49.4 & 75.5 \\
2  & 49.5 & 75.4 \\
\textbf{3}  & \textbf{50.2} & \textbf{76.0} \\
4  & 49.8 & 75.7 \\
5  & 47.1 & 75.5 \\
\bottomrule
\end{tabular}
\vspace{2pt}
\caption{Ablation study of different $\lambda$ values used in the angle loss on the DOTA-v1.0 validation set, using YOLO26s as the backbone. "---" indicates no angle loss is used.}
\label{tab:obb_lambda_ablation}
\end{table}

We evaluate our models on the DOTA-v1.0~\cite{xia2018dota} dataset, one of the largest and most commonly used datasets for oriented object detection, containing 2,806 images and 188,282 instances across 15 categories. We split the images into overlapping $1024 \times 1024$ crops and train our models on the training set.

We conduct ablation studies on the DOTA-v1.0 validation
set to evaluate the effectiveness of the proposed
methods, using YOLO26s without the angle loss as the
baseline. As shown in
Table~\ref{tab:obb_angle_ablation}, using the
$[-45^\circ, 135^\circ)$ angle definition improves the
mAP from 47.7 to 49.0, indicating that a more
continuous angle parameterization is beneficial for OBB
detection. Table~\ref{tab:obb_lambda_ablation}
summarizes the ablations over the $\lambda$
hyperparameter in the angle loss, where "---"
indicates that the angle loss is not used. All
configurations with the angle loss outperform the
version without it, and $\lambda=3$ gives the highest
mAP at 50.2. The full YOLO11 and YOLO26 comparison on
the DOTA-v1.0 test set is provided in
Sec.~\ref{sec:suppl_task_benchmarks} of the
supplementary materials and
Table~\ref{tab:yolo26_dota_obb}. YOLO26 improves OBB
mAP by \textbf{+2.5 to +3.4} over YOLO11 across scales,
with larger AP$_{75}$ gains of \textbf{+4.6 to +6.0},
indicating clearer improvements under stricter
localization metrics.

\begin{figure}[t]
\centering

\begin{minipage}[t]{0.48\columnwidth}
  \centering
  \includegraphics[width=\textwidth]{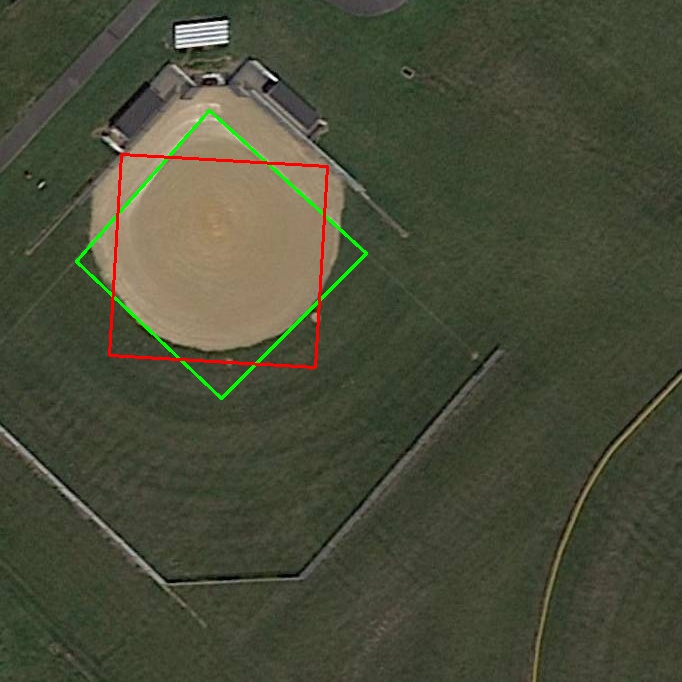}\vspace{2pt}
  \includegraphics[width=\textwidth]{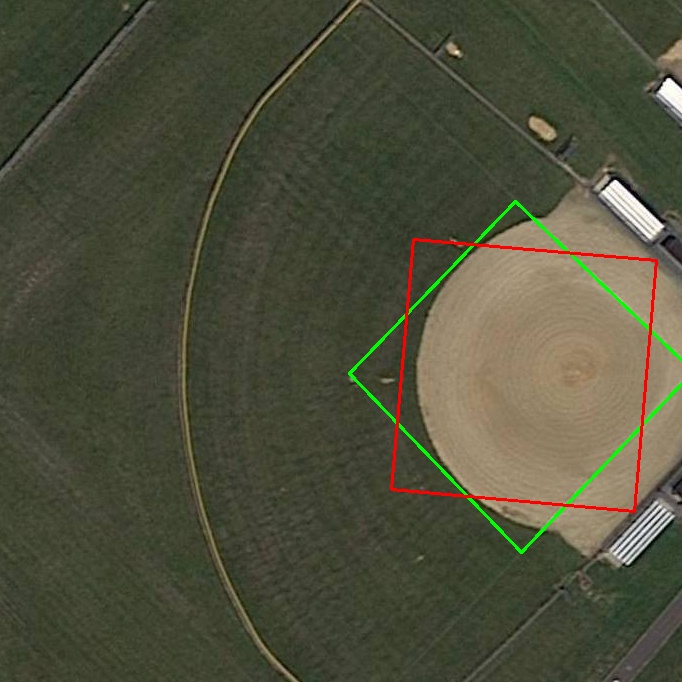}\vspace{2pt}
  \includegraphics[width=\textwidth]{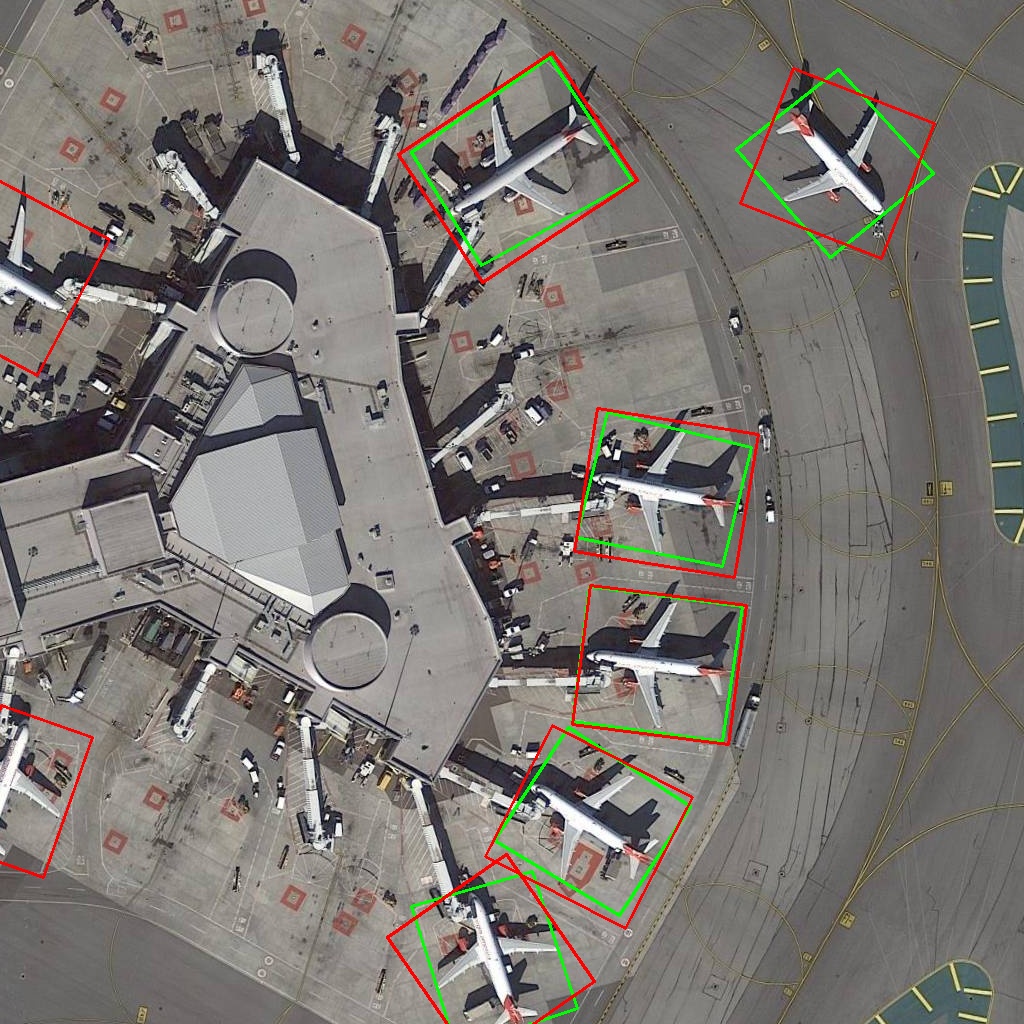}\vspace{2pt}
  \includegraphics[width=\textwidth]{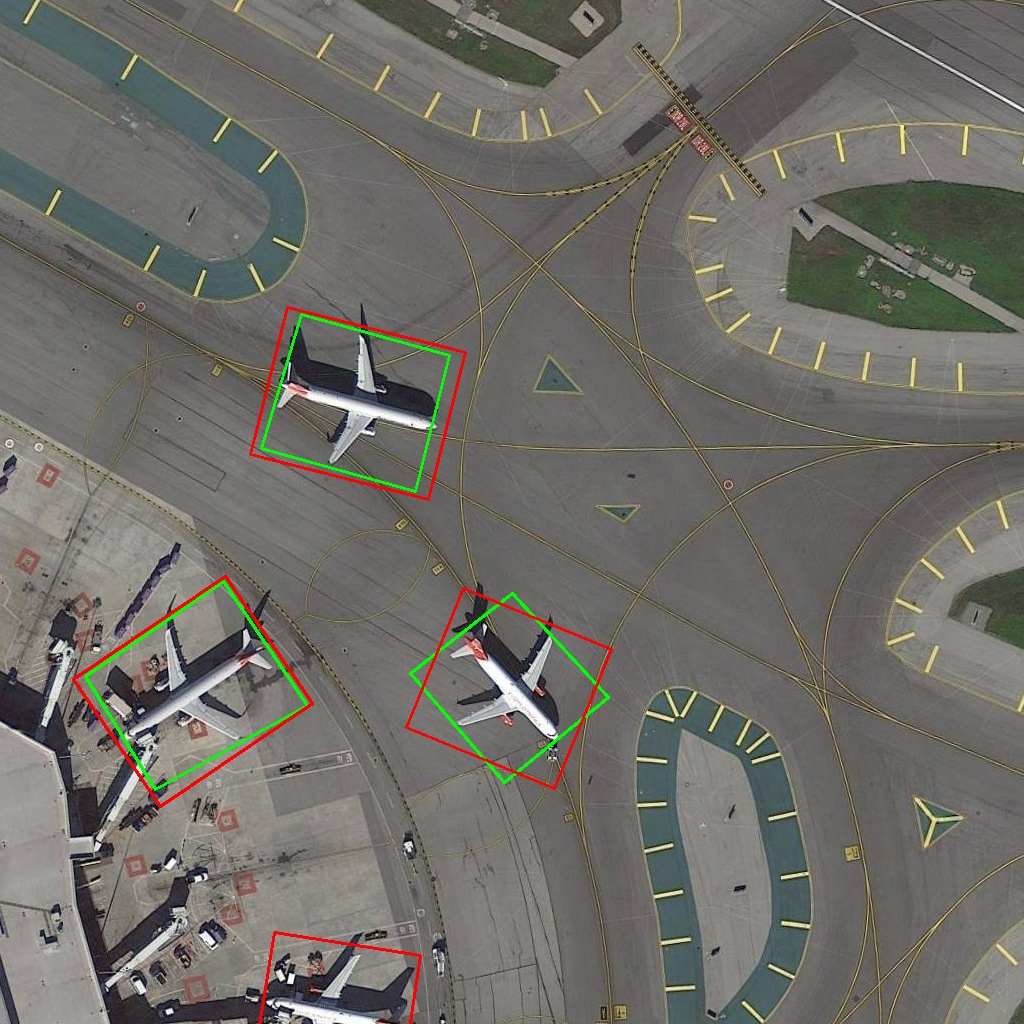}\vspace{2pt}
  {\small YOLO11x-obb}
\end{minipage}
\hspace{0pt}
\begin{minipage}[t]{0.48\columnwidth}
  \centering
  \includegraphics[width=\textwidth]{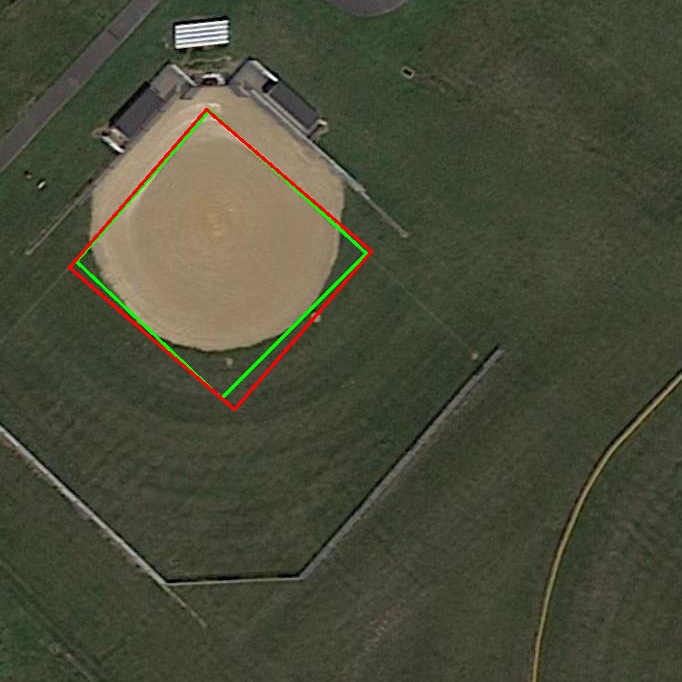}\vspace{2pt}
  \includegraphics[width=\textwidth]{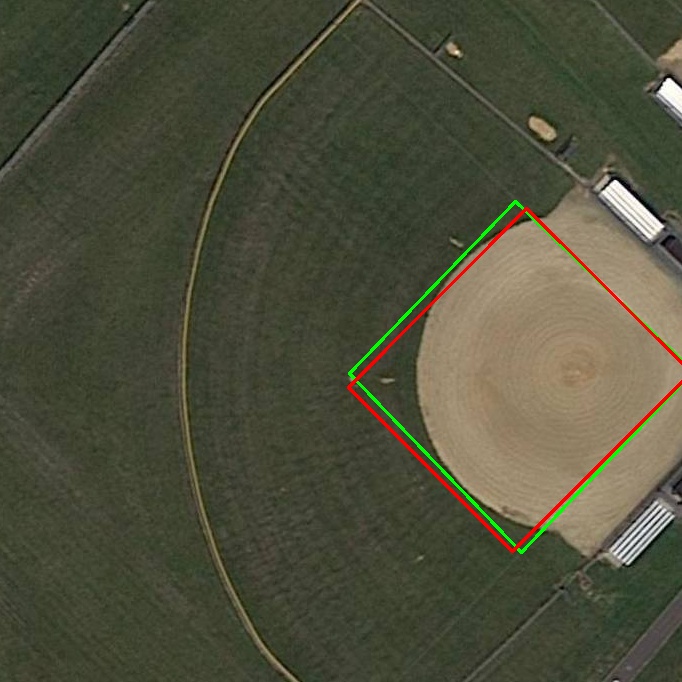}\vspace{2pt}
  \includegraphics[width=\textwidth]{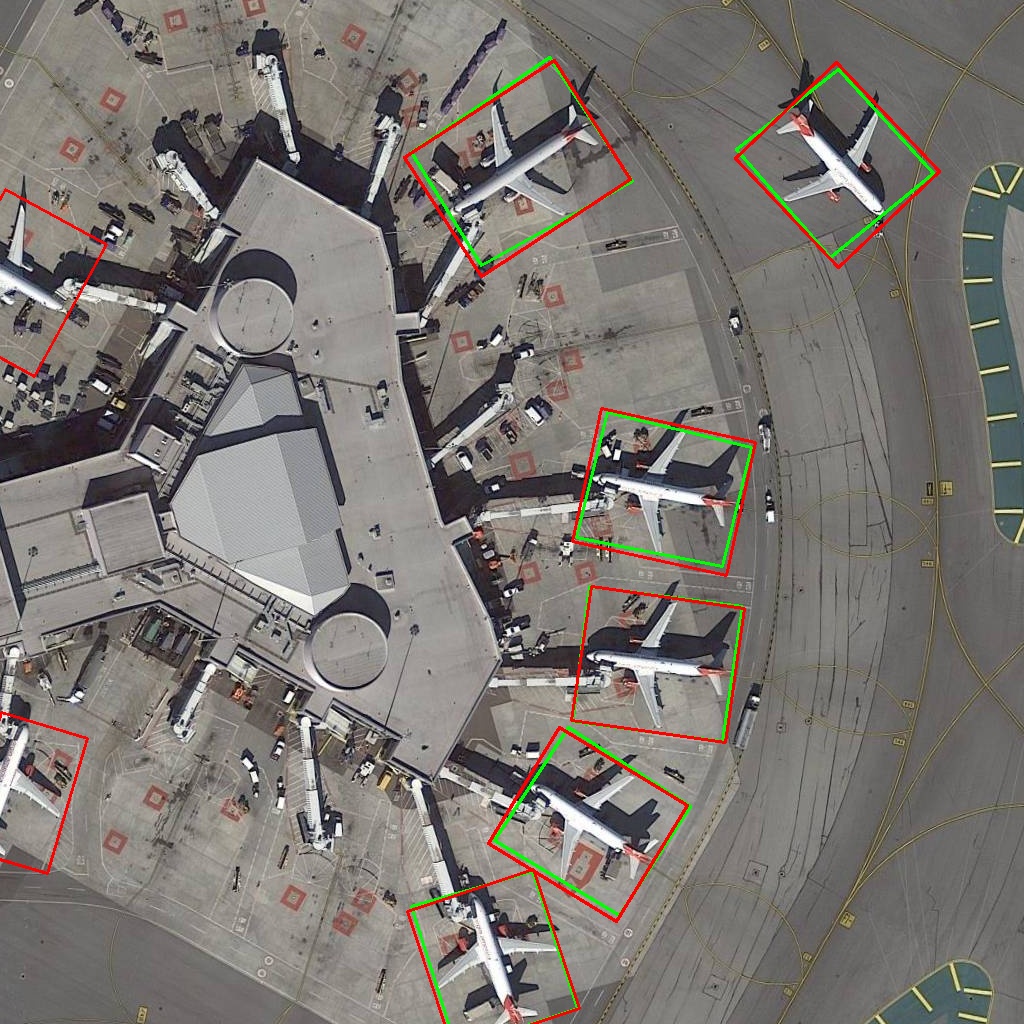}\vspace{2pt}
  \includegraphics[width=\textwidth]{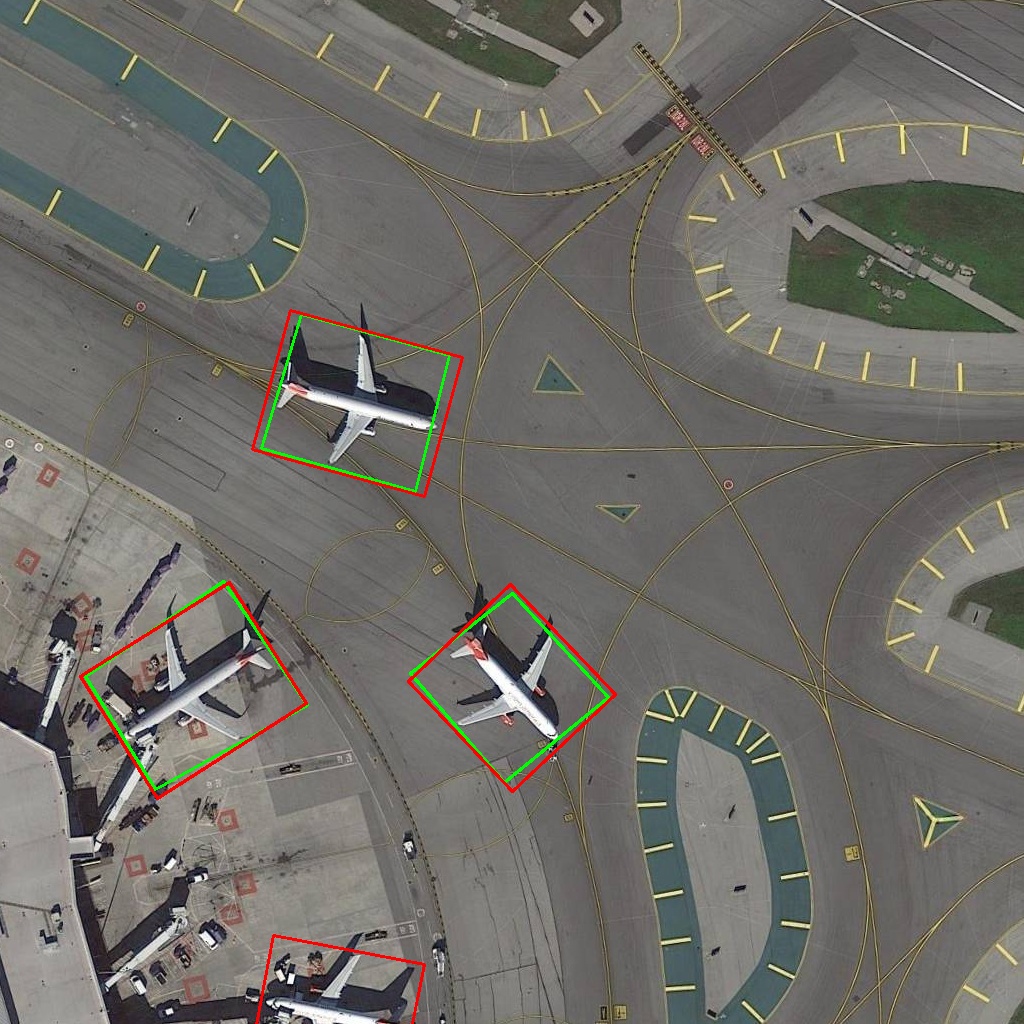}\vspace{2pt}
  {\small YOLO26x-obb}
\end{minipage}

\caption{Qualitative comparisons between YOLO26x-obb and
YOLO11x-obb models for square rotated objects
detection. Green boxes are ground truth, while red
boxes indicate predictions. YOLO26 demonstrates better
angle predictions over YOLO11 on square objects.}
\label{fig:obb_qualitative}
\end{figure}

Additional qualitative examples are provided in
Fig.~\ref{fig:obb_qualitative}. They show that
YOLO26 produces visibly better angle predictions than
YOLO11 on square rotated objects, consistent with the
larger gains observed in AP75 and overall mAP than in
AP50.

\subsection{YOLOE-26 Results}
\label{sec:yoloe26_results}

\subsubsection{Ablation Study}
\begin{table}[t]
\centering
\scriptsize
\setlength{\tabcolsep}{3pt}
\caption{Ablation study of YOLOE-26s-TP on the LVIS \texttt{minival} split. Starting from the YOLOE-11s-TP baseline, we progressively introduce the YOLOE-26 modifications described in Sec.~\ref{sec:yoloe26} and report $\text{AP}$ (mAP$_{50{:}95}$) under both end-to-end (E2E) and non-end-to-end (non-E2E) evaluation protocols.}
\label{tab:ablation}
\begin{tabular}{c|ccc|cc}
\toprule
Backbone & Decoupled Seg. & Text Enc. Upgrade & Data Engine & AP$_{\text{E2E}}$ & AP$_{\text{non-E2E}}$ \\
\midrule
v8s & --- & ---& --- & ---   & 27.9 \\
11s & --- & ---& --- & ---   & 27.5 \\
\midrule
26s & --- & ---& --- & 27.8 & 29.0 \\
26s & \checkmark & ---& --- & 28.5 & 29.5 \\
26s & \checkmark & \checkmark & --- & 28.8 & 29.7 \\
26s & \checkmark & \checkmark & \checkmark & \textbf{29.9} & \textbf{31.0} \\
\bottomrule
\end{tabular}
\end{table}
Table~\ref{tab:ablation} isolates the four YOLOE-26 modifications introduced in Sec.~\ref{sec:yoloe26}: Backbone upgrade, Decoupled segmentation training, Text encoder upgrade, and Data engine. Unless otherwise stated, all rows use the same TP training setup with Objects365-v1~\cite{shao2019objects365} and GoldG grounding data~\cite{zhang2022glipv2}, including GQA~\cite{hudson2019gqa} and Flickr30k~\cite{plummer2015flickr30k} with COCO images excluded, and use MobileCLIP as the default text encoder. We report AP under both end-to-end (E2E) and non-end-to-end (non-E2E) evaluation protocols.

For reference, the YOLOE-v8s-TP and YOLOE-11s-TP baselines achieve 27.9 and 27.5~AP$_{\text{non-E2E}}$ respectively. We begin from the latter. Applying the Backbone upgrade by replacing with pretrained YOLO26s weights yields 27.8/29.0~AP (E2E/non-E2E), a notable +1.5~AP gain under the non-E2E protocol. Enabling Decoupled segmentation training further improves E2E AP by +0.7 to 28.5 and non-E2E AP by +0.5 to 29.5, supporting the view that removing the auxiliary segmentation objective reduces task interference. The Text encoder upgrade from MobileCLIP to MobileCLIP2 brings an additional +0.3/+0.2~AP (E2E/non-E2E), reaching 28.8/29.7. Finally, the Data engine described in Sec.~\ref{sec:yoloe26} yields the largest single improvement of +1.1/+1.3~AP, reaching the final 29.9/31.0~AP result. This gain is likely due to the improved grounding data quality and diversity, which better supports the open-vocabulary detection capability of the model.

\subsubsection{Prompt-based evaluation}
The full text-prompted and visual-prompted detection
comparison is provided in
Sec.~\ref{sec:suppl_yoloe_benchmarks} of the
supplementary materials and
Table~\ref{tab:yoloe26_det_prompt_detail}. Overall,
YOLOE-26 improves over earlier YOLOE variants across
scales under both text and visual prompting. Under
text prompting, the Non-E2E branch improves AP by
\textbf{+1.9--2.9} over YOLOE-v8 and
\textbf{+2.4--3.3} over YOLOE-11 across the s/m/l
models; under visual prompting, the corresponding
gains are \textbf{+2.1--2.9} and \textbf{+2.3--2.6}.
The largest model achieves the strongest prompt-based
results, reaching \textbf{40.6~AP} with text prompts
and \textbf{38.5~AP} with visual prompts, while the
lightweight YOLOE-26n still reaches 24.7~AP with only
3.9M parameters. For text-prompted inference, the
end-to-end head remains close to the Non-E2E variant,
trailing by at most 1.1 AP across scales. Under visual
prompting, the gap ranges from 1.0 to 2.6 AP.

For zero-shot segmentation, the full comparison is
provided in Sec.~\ref{sec:suppl_yoloe_benchmarks} of the
supplementary materials and Table~\ref{tab:seg}.
YOLOE-26 likewise improves zero-shot mask prediction
across the s/m/l models under both text and visual
prompts, with $\text{AP}^{m}$ gains of
\textbf{+1.3--3.0} over YOLOE-v8 and
\textbf{+1.8--2.9} over YOLOE-11. YOLOE-26x reaches
the best overall result of
\textbf{27.4 / 26.7 $\text{AP}^{m}$} under text/visual
prompting, indicating that the YOLO26 backbone and
training refinements benefit zero-shot segmentation as
well as open-vocabulary detection.

\subsubsection{Prompt-free evaluation}
The full prompt-free comparison is provided in
Sec.~\ref{sec:suppl_yoloe_benchmarks} of the
supplementary materials and
Table~\ref{tab:yoloe26_pf}. YOLOE-26 remains
competitive across the model family in the prompt-free
setting. In the standard Non-E2E setting, YOLOE-26
improves AP by \textbf{+0.8--1.7} over YOLOE-v8 and
\textbf{+0.9--2.0} over YOLOE-11 across the s/m/l
models, reaching up to \textbf{31.1~AP} on LVIS
\texttt{minival}. The E2E head stays within
\textbf{0.7--1.2~AP} of the corresponding Non-E2E
models while removing post-processing.

\section{Conclusion}

We presented YOLO26, a unified real-time vision model family that combines a dual-head NMS-free architecture with MuSGD, Progressive Loss, and STAL to improve the accuracy--latency trade-off across five model scales.
By removing DFL and strengthening optimization and label assignment, YOLO26 achieves a lighter detection head while preserving localization quality.
Beyond detection, the YOLO26 family combines task-specific refinements (multi-scale prototype fusion for segmentation, uncertainty-aware keypoint regression for pose, and revised OBB parameterization with dedicated angle supervision) with the shared detector improvements, improving over YOLO11 by up to +2.5 box AP and +3.7 mask AP on COCO instance segmentation, +7.2 AP on COCO pose estimation, and +3.4 mAP on DOTA-v1.0 OBB detection, while maintaining a unified training and deployment pipeline with native export support across the 19 non-PyTorch targets exposed by Ultralytics.
On COCO, YOLO26 reaches 40.9--57.5~mAP at 1.7--11.8\,ms T4 TensorRT latency.
YOLOE-26 further extends the family to open-vocabulary detection, where YOLOE-26x reaches 40.6~AP on LVIS minival under text prompting and 38.5~AP under visual prompting, while YOLOE-26 also remains competitive in the prompt-free setting (up to 31.1~AP), showing that the stronger YOLO26 detector and the additional open-vocabulary refinements jointly improve performance in the open-vocabulary setting.
Future work includes broader evaluation beyond COCO-centric benchmarks and further exploration of learned or task-adaptive loss-schedule shapes beyond the linear $\alpha(t)$ used here, as well as pretraining beyond Objects365-v1, including web-scale or grounding-style corpora.

\clearpage
{
    \small
    \bibliographystyle{ieeenat_fullname}
    \bibliography{main}

@inproceedings{girshick2014rich,
  title={Rich feature hierarchies for accurate object detection and semantic segmentation},
  author={Girshick, Ross and Donahue, Jeff and Darrell, Trevor and Malik, Jitendra},
  booktitle={Proceedings of the IEEE conference on computer vision and pattern recognition},
  pages={580--587},
  year={2014}
}

@inproceedings{girshick2015fast,
  title={Fast r-cnn},
  author={Girshick, Ross},
  booktitle={Proceedings of the IEEE international conference on computer vision},
  pages={1440--1448},
  year={2015}
}

@article{ren2015faster,
  title={Faster r-cnn: Towards real-time object detection with region proposal networks},
  author={Ren, Shaoqing and He, Kaiming and Girshick, Ross and Sun, Jian},
  journal={Advances in neural information processing systems},
  volume={28},
  year={2015}
}

@inproceedings{he2017mask,
  title={Mask r-cnn},
  author={He, Kaiming and Gkioxari, Georgia and Doll{\'a}r, Piotr and Girshick, Ross},
  booktitle={Proceedings of the IEEE international conference on computer vision},
  pages={2961--2969},
  year={2017}
}

@inproceedings{liu2016ssd,
  title={Ssd: Single shot multibox detector},
  author={Liu, Wei and Anguelov, Dragomir and Erhan, Dumitru and Szegedy, Christian and Reed, Scott and Fu, Cheng-Yang and Berg, Alexander C},
  booktitle={European conference on computer vision},
  pages={21--37},
  year={2016},
  organization={Springer}
}

@inproceedings{lin2017focal,
  title={Focal loss for dense object detection},
  author={Lin, Tsung-Yi and Goyal, Priya and Girshick, Ross and He, Kaiming and Doll{\'a}r, Piotr},
  booktitle={Proceedings of the IEEE international conference on computer vision},
  pages={2980--2988},
  year={2017}
}

@inproceedings{li2020gfl,
  title     = {Generalized Focal Loss: Learning Qualified and Distributed Bounding Boxes for Dense Object Detection},
  author    = {Li, Xiang and Wang, Wenhai and Wu, Lijun and Chen, Shuo and Hu, Xiaolin and Li, Jun and Tang, Jinhui and Yang, Jian},
  booktitle = {Advances in Neural Information Processing Systems},
  volume    = {33},
  pages     = {21002--21012},
  publisher = {Curran Associates, Inc.},
  year      = {2020},
  url       = {https://proceedings.neurips.cc/paper_files/paper/2020/file/f0bda020d2470f2e74990a07a607ebd9-Paper.pdf}
}

@article{redmon2018yolov3,
  title={Yolov3: An incremental improvement},
  author={Redmon, Joseph and Farhadi, Ali},
  journal={arXiv preprint arXiv:1804.02767},
  year={2018}
}

@misc{jocher2020yolov5,
  title     = {ultralytics/yolov5: Initial Release},
  author    = {Jocher, Glenn and others},
  year      = {2020},
  publisher = {Zenodo},
  doi       = {10.5281/zenodo.3908560},
  url       = {https://zenodo.org/records/3908560}
}

@inproceedings{tian2019fcos,
  title={Fcos: Fully convolutional one-stage object detection},
  author={Tian, Zhi and Shen, Chunhua and Chen, Hao and He, Tong},
  booktitle={Proceedings of the IEEE/CVF international conference on computer vision},
  pages={9627--9636},
  year={2019}
}

@inproceedings{duan2019centernet,
  title={Centernet: Keypoint triplets for object detection},
  author={Duan, Kaiwen and Bai, Song and Xie, Lingxi and Qi, Honggang and Huang, Qingming and Tian, Qi},
  booktitle={Proceedings of the IEEE/CVF international conference on computer vision},
  pages={6569--6578},
  year={2019}
}

@misc{ultralytics2023yolov8,
  title        = {Explore Ultralytics YOLOv8},
  author       = {{Ultralytics}},
  year         = {2023},
  howpublished = {\url{https://docs.ultralytics.com/models/yolov8/}},
  note         = {Online documentation (no formal paper).}
}

@article{wang2024yolov10,
  title={Yolov10: Real-time end-to-end object detection},
  author={Wang, Ao and Chen, Hui and Liu, Lihao and Chen, Kai and Lin, Zijia and Han, Jungong and others},
  journal={Advances in Neural Information Processing Systems},
  volume={37},
  pages={107984--108011},
  year={2024}
}

@misc{ultralytics2024yolo11_docs,
  title        = {Ultralytics YOLO11},
  author       = {{Ultralytics}},
  year         = {2024},
  howpublished = {\url{https://docs.ultralytics.com/models/yolo11/}},
  note         = {Online documentation (no formal paper). Accessed: 2026-02-02}
}

@misc{ultralytics2026yolo26_docs,
  title        = {Ultralytics YOLO26},
  author       = {{Ultralytics}},
  year         = {2026},
  howpublished = {\url{https://docs.ultralytics.com/models/yolo26/}},
  note         = {Online documentation and released benchmark tables. Accessed: 2026-03-13}
}

@misc{ultralytics2026_export_docs,
  title        = {Ultralytics Export Mode},
  author       = {{Ultralytics}},
  year         = {2026},
  howpublished = {\url{https://docs.ultralytics.com/modes/export/}},
  note         = {Online documentation describing supported export formats and deployment targets. Accessed: 2026-04-10}
}

@misc{ultralytics2026_tasks_docs,
  title        = {Ultralytics Computer Vision Tasks},
  author       = {{Ultralytics}},
  year         = {2026},
  howpublished = {\url{https://docs.ultralytics.com/tasks/}},
  note         = {Online documentation describing supported computer vision tasks. Accessed: 2026-06-01}
}

@misc{ultralytics2026_train_docs,
  title        = {Ultralytics Train Mode},
  author       = {{Ultralytics}},
  year         = {2026},
  howpublished = {\url{https://docs.ultralytics.com/modes/train/}},
  note         = {Online documentation describing model training workflows. Accessed: 2026-06-01}
}

@misc{ultralytics2026_val_docs,
  title        = {Ultralytics Val Mode},
  author       = {{Ultralytics}},
  year         = {2026},
  howpublished = {\url{https://docs.ultralytics.com/modes/val/}},
  note         = {Online documentation describing model validation workflows. Accessed: 2026-06-01}
}

@misc{ultralytics2026_predict_docs,
  title        = {Ultralytics Predict Mode},
  author       = {{Ultralytics}},
  year         = {2026},
  howpublished = {\url{https://docs.ultralytics.com/modes/predict/}},
  note         = {Online documentation describing model inference workflows. Accessed: 2026-06-01}
}

@misc{ncnn_docs,
  title        = {ncnn Documentation},
  author       = {{ncnn Contributors}},
  year         = {2026},
  howpublished = {\url{https://ncnn.readthedocs.io/en/latest/}},
  note         = {Official documentation for the ncnn inference framework. Accessed: 2026-04-10}
}

@misc{executorch_docs,
  title        = {ExecuTorch Documentation},
  author       = {{PyTorch Contributors}},
  year         = {2026},
  howpublished = {\url{https://docs.pytorch.org/executorch/stable/}},
  note         = {Official documentation for ExecuTorch. Accessed: 2026-04-10}
}

@inproceedings{tian2025yolov12,
  title        = {YOLOv12: Attention-Centric Real-Time Object Detectors},
  author       = {Tian, Yunjie and Ye, Qixiang and Doermann, David},
  booktitle    = {Advances in Neural Information Processing Systems},
  year         = {2025}
}

@article{lv2024rt,
  title        = {RT-DETRv2: Improved Baseline with Bag-of-Freebies for Real-Time Detection Transformer},
  author       = {Lv, Wenyu and Zhao, Yian and Chang, Qinyao and Huang, Kui and Wang, Guanzhong and Liu, Yi},
  journal      = {arXiv preprint arXiv:2407.17140},
  year         = {2024}
}

@inproceedings{feng2021tood,
  title     = {TOOD: Task-Aligned One-Stage Object Detection},
  author    = {Feng, Chengjian and Zhong, Yujie and Gao, Yu and Scott, Matthew R. and Huang, Weilin},
  booktitle = {Proceedings of the IEEE/CVF International Conference on Computer Vision (ICCV)},
  pages     = {3490--3499},
  year      = {2021},
  doi       = {10.1109/ICCV48922.2021.00349}
}

@inproceedings{shao2019objects365,
  title={Objects365: A Large-scale, High-quality Dataset for Object Detection},
  author={Shao, Shuai and Li, Zeming and Zhang, Tianyuan and Peng, Chao and Yu, Gang and Li, Jing and Zhang, Xiangyu and Sun, Jian},
  booktitle={Proceedings of the IEEE/CVF International Conference on Computer Vision},
  pages={8425--8434},
  year={2019}
}

@inproceedings{hudson2019gqa,
  title={GQA: A New Dataset for Real-World Visual Reasoning and Compositional Question Answering},
  author={Hudson, Drew A. and Manning, Christopher D.},
  booktitle={Proceedings of the IEEE/CVF Conference on Computer Vision and Pattern Recognition (CVPR)},
  pages={6700--6709},
  year={2019},
  doi={10.1109/CVPR.2019.00686}
}

@inproceedings{plummer2015flickr30k,
  title={Flickr30k Entities: Collecting Region-to-Phrase Correspondences for Richer Image-to-Sentence Models},
  author={Plummer, Bryan A. and Wang, Liwei and Cervantes, Chris M. and Caicedo, Juan C. and Hockenmaier, Julia and Lazebnik, Svetlana},
  booktitle={Proceedings of the IEEE International Conference on Computer Vision (ICCV)},
  pages={2641--2649},
  year={2015}
}

@inproceedings{wang2024yolov9,
  title={Yolov9: Learning what you want to learn using programmable gradient information},
  author={Wang, Chien-Yao and Yeh, I-Hau and Mark Liao, Hong-Yuan},
  booktitle={European conference on computer vision},
  pages={1--21},
  year={2024},
  organization={Springer}
}

@inproceedings{detr,
  title     = {End-to-End Object Detection with Transformers},
  author    = {Carion, Nicolas and Massa, Francisco and Synnaeve, Gabriel and Usunier, Nicolas and Kirillov, Alexander and Zagoruyko, Sergey},
  booktitle = {Computer Vision -- ECCV 2020},
  series    = {Lecture Notes in Computer Science},
  volume    = {12346},
  pages     = {213--229},
  publisher = {Springer, Cham},
  year      = {2020},
  doi       = {10.1007/978-3-030-58452-8_13}
}

@inproceedings{deformable_detr,
  title     = {Deformable {DETR}: Deformable Transformers for End-to-End Object Detection},
  author    = {Zhu, Xizhou and Su, Weijie and Lu, Lewei and Li, Bin and Wang, Xiaogang and Dai, Jifeng},
  booktitle = {International Conference on Learning Representations (ICLR)},
  year      = {2021},
  url       = {https://openreview.net/forum?id=gZ9hCDWe6ke}
}

@inproceedings{dab_detr,
  title     = {{DAB-DETR}: Dynamic Anchor Boxes are Better Queries for {DETR}},
  author    = {Liu, Shilong and Li, Feng and Zhang, Hao and Yang, Xiao and Qi, Xianbiao and Su, Hang and Zhu, Jun and Zhang, Lei},
  booktitle = {International Conference on Learning Representations (ICLR)},
  year      = {2022},
  url       = {https://openreview.net/forum?id=oMI9PjOb9Jl}
}

@inproceedings{dn_detr,
  title     = {{DN-DETR}: Accelerate {DETR} Training by Introducing Query DeNoising},
  author    = {Li, Feng and Zhang, Hao and Liu, Shilong and Guo, Jian and Ni, Lionel M. and Zhang, Lei},
  booktitle = {Proceedings of the IEEE/CVF Conference on Computer Vision and Pattern Recognition (CVPR)},
  month     = {June},
  year      = {2022},
  pages     = {13619--13627}
}

@inproceedings{dino,
  title     = {{DINO}: {DETR} with Improved DeNoising Anchor Boxes for End-to-End Object Detection},
  author    = {Zhang, Hao and Li, Feng and Liu, Shilong and Zhang, Lei and Su, Hang and Zhu, Jun and Ni, Lionel and Shum, Heung-Yeung},
  booktitle = {International Conference on Learning Representations (ICLR)},
  year      = {2023},
  url       = {https://openreview.net/forum?id=3mRwyG5one}
}

@inproceedings{hdetr,
  title     = {{DETRs} With Hybrid Matching},
  author    = {Jia, Ding and Yuan, Yuhui and He, Haodi and Wu, Xiaopei and Yu, Haojun and Lin, Weihong and Sun, Lei and Zhang, Chao and Hu, Han},
  booktitle = {Proceedings of the IEEE/CVF Conference on Computer Vision and Pattern Recognition (CVPR)},
  month     = {June},
  year      = {2023},
  pages     = {19702--19712}
}

@inproceedings{group_detr,
  title     = {Group {DETR}: Fast {DETR} Training with Group-Wise One-to-Many Assignment},
  author    = {Chen, Qiang and Chen, Xiaokang and Wang, Jian and Zhang, Shan and Yao, Kun and Feng, Haocheng and Han, Junyu and Ding, Errui and Zeng, Gang and Wang, Jingdong},
  booktitle = {Proceedings of the IEEE/CVF International Conference on Computer Vision (ICCV)},
  month     = {October},
  year      = {2023},
  pages     = {6633--6642}
}

@inproceedings{rtdetr,
  title     = {{DETRs} Beat {YOLOs} on Real-time Object Detection},
  author    = {Zhao, Yian and Lv, Wenyu and Xu, Shangliang and Wei, Jinman and Wang, Guanzhong and Dang, Qingqing and Liu, Yi and Chen, Jie},
  booktitle = {Proceedings of the IEEE/CVF Conference on Computer Vision and Pattern Recognition (CVPR)},
  month     = {June},
  year      = {2024},
  pages     = {16965--16974}
}

@inproceedings{dfine,
  title     = {{D-FINE}: Redefine Regression Task of {DETRs} as Fine-grained Distribution Refinement},
  author    = {Peng, Yansong and Li, Hebei and Wu, Peixi and Zhang, Yueyi and Sun, Xiaoyan and Wu, Feng},
  booktitle = {International Conference on Learning Representations (ICLR)},
  year      = {2025},
  url       = {https://proceedings.iclr.cc/paper_files/paper/2025/hash/6cf58a87e3097e7d1f9be3e8693a93de-Abstract-Conference.html}
}

@inproceedings{deim,
  title     = {{DEIM}: {DETR} with Improved Matching for Fast Convergence},
  author    = {Huang, Shihua and Lu, Zhichao and Cun, Xiaodong and Yu, Yongjun and Zhou, Xiao and Shen, Xi},
  booktitle = {Proceedings of the IEEE/CVF Conference on Computer Vision and Pattern Recognition (CVPR)},
  month     = {June},
  year      = {2025},
  pages     = {15162--15171}
}

@article{deimv2,
  title   = {Real-Time Object Detection Meets {DINOv3}},
  author  = {Huang, Shihua and Hou, Yongjie and Liu, Longfei and Yu, Xuanlong and Shen, Xi},
  journal = {arXiv preprint arXiv:2509.20787},
  year    = {2025},
  url     = {https://arxiv.org/abs/2509.20787}
}

@article{rtdetrv4,
  title   = {{RT-DETRv4}: Painlessly Furthering Real-Time Object Detection with Vision Foundation Models},
  author  = {Liao, Zijun and Zhao, Yian and Shan, Xin and Yan, Yu and Liu, Chang and Lu, Lei and Ji, Xiangyang and Chen, Jie},
  journal = {arXiv preprint arXiv:2510.25257},
  year    = {2025},
  url     = {https://arxiv.org/abs/2510.25257}
}

@article{lwdetr,
  title   = {{LW-DETR}: A Transformer Replacement to {YOLO} for Real-Time Detection},
  author  = {Chen, Qiang and Su, Xiangbo and Zhang, Xinyu and Wang, Jian and Chen, Jiahui and Shen, Yunpeng and Han, Chuchu and Chen, Ziliang and Xu, Weixiang and Li, Fanrong and Zhang, Shan and Yao, Kun and Ding, Errui and Zhang, Gang and Wang, Jingdong},
  journal = {arXiv preprint arXiv:2406.03459},
  year    = {2024},
  url     = {https://arxiv.org/abs/2406.03459}
}

@article{rfdetr,
  title   = {{RF-DETR}: Neural Architecture Search for Real-Time Detection Transformers},
  author  = {Robinson, Isaac and Robicheaux, Peter and Popov, Matvei and Ramanan, Deva and Peri, Neehar},
  journal = {arXiv preprint arXiv:2511.09554},
  year    = {2025},
  url     = {https://arxiv.org/abs/2511.09554}
}

@article{muon2025,
  title   = {Muon is Scalable for LLM Training},
  author  = {Liu, Jingyuan and Su, Jianlin and Yao, Xingcheng and Jiang, Zhejun and Lai, Guokun and Du, Yulun and Qin, Yidao and Xu, Weixin and Lu, Enzhe and Yan, Junjie and Chen, Yanru and Zheng, Huabin and Liu, Yibo and Yin, Bohong and He, Weiran and Zhu, Han and Wang, Yuzhi and Wang, Jianzhou and Dong, Mengnan and Zhang, Zheng and Kang, Yongsheng and Zhang, Hao and Xu, Xinran and Zhang, Yutao and Wu, Yuxin and Zhou, Xinyu and Yang, Zhilin},
  journal = {arXiv preprint arXiv:2502.16982},
  year    = {2025},
  url     = {https://arxiv.org/abs/2502.16982}
}

@article{kimi_k2,
  title   = {Kimi K2: Open Agentic Intelligence},
  author  = {{Moonshot AI}},
  journal = {arXiv preprint arXiv:2507.20534},
  year    = {2025},
  url     = {https://arxiv.org/abs/2507.20534}
}

@misc{sgd_pytorch,
  title        = {torch.optim.SGD},
  author       = {{PyTorch Contributors}},
  year         = {2026},
  howpublished = {\url{https://docs.pytorch.org/docs/stable/generated/torch.optim.SGD.html}},
  note         = {Accessed: 2026-02-02}
}

@misc{onnx,
  title        = {{ONNX}: Open Neural Network Exchange},
  author       = {{ONNX Contributors}},
  year         = {2019},
  howpublished = {\url{https://github.com/onnx/onnx}},
  note         = {Accessed: 2026-02-05}
}

@misc{tensorrt,
  title        = {{NVIDIA TensorRT}: High-Performance Deep Learning Inference SDK},
  author       = {{NVIDIA}},
  year         = {2024},
  howpublished = {\url{https://github.com/NVIDIA/TensorRT}},
  note         = {Accessed: 2026-02-05}
}

@misc{coreml,
  title        = {{Core ML Tools}},
  author       = {{Apple}},
  year         = {2024},
  howpublished = {\url{https://github.com/apple/coremltools}},
  note         = {Accessed: 2026-02-05}
}

@misc{tflite,
  title        = {{TensorFlow Lite}},
  author       = {{Google}},
  year         = {2024},
  howpublished = {\url{https://ai.google.dev/edge/litert}},
  note         = {Accessed: 2026-02-05}
}

@misc{openvino,
  title        = {{Intel Distribution of OpenVINO Toolkit}},
  author       = {{Intel}},
  year         = {2024},
  howpublished = {\url{https://github.com/openvinotoolkit/openvino}},
  note         = {Accessed: 2026-02-05}
}

@inproceedings{rle2021,
  title     = {Human Pose Regression with Residual Log-likelihood Estimation},
  author    = {Li, Jiefeng and Bian, Siyuan and Zeng, Ailing and Wang, Can and Pang, Bo and Liu, Wentao and Lu, Cewu},
  booktitle = {Proceedings of the IEEE/CVF International Conference on Computer Vision (ICCV)},
  pages     = {11025--11034},
  year      = {2021}
}

@inproceedings{bolya2019yolact,
  title     = {YOLACT: Real-Time Instance Segmentation},
  author    = {Bolya, Daniel and Zhou, Chong and Xiao, Fanyi and Lee, Yong Jae},
  booktitle = {Proceedings of the IEEE/CVF International Conference on Computer Vision (ICCV)},
  month     = {October},
  year      = {2019},
  pages     = {9157--9166}
}

@inproceedings{tian2020condinst,
  title     = {Conditional Convolutions for Instance Segmentation},
  author    = {Tian, Zhi and Shen, Chunhua and Chen, Hao},
  booktitle = {Computer Vision -- ECCV 2020},
  series    = {Lecture Notes in Computer Science},
  publisher = {Springer, Cham},
  year      = {2020},
  doi       = {10.1007/978-3-030-58452-8_17}
}

@inproceedings{wang2020solov2,
  title     = {SOLOv2: Dynamic and Fast Instance Segmentation},
  author    = {Wang, Xinlong and Zhang, Rufeng and Kong, Tao and Li, Lei and Shen, Chunhua},
  booktitle = {Advances in Neural Information Processing Systems (NeurIPS)},
  year      = {2020}
}

@inproceedings{cheng2022mask2former,
  title     = {Masked-Attention Mask Transformer for Universal Image Segmentation},
  author    = {Cheng, Bowen and Misra, Ishan and Schwing, Alexander G. and Kirillov, Alexander and Girdhar, Rohit},
  booktitle = {Proceedings of the IEEE/CVF Conference on Computer Vision and Pattern Recognition (CVPR)},
  month     = {June},
  year      = {2022},
  pages     = {1290--1299}
}

@inproceedings{li2023maskdino,
  title     = {Mask DINO: Towards a Unified Transformer-Based Framework for Object Detection and Segmentation},
  author    = {Li, Feng and Zhang, Hao and Xu, Huaizhe and Liu, Shilong and Zhang, Lei and Ni, Lionel M. and Shum, Heung-Yeung},
  booktitle = {Proceedings of the IEEE/CVF Conference on Computer Vision and Pattern Recognition (CVPR)},
  month     = {June},
  year      = {2023},
  pages     = {3041--3050}
}

@inproceedings{deeppose2014,
  title     = {{DeepPose}: Human Pose Estimation via Deep Neural Networks},
  author    = {Toshev, Alexander and Szegedy, Christian},
  booktitle = {Proceedings of the IEEE Conference on Computer Vision and Pattern Recognition (CVPR)},
  pages     = {1653--1660},
  year      = {2014}
}

@inproceedings{hourglass2016,
  title     = {Stacked Hourglass Networks for Human Pose Estimation},
  author    = {Newell, Alejandro and Yang, Kaiyu and Deng, Jia},
  booktitle = {Computer Vision -- ECCV 2016},
  series    = {Lecture Notes in Computer Science},
  volume    = {9912},
  pages     = {483--499},
  publisher = {Springer, Cham},
  year      = {2016}
}

@inproceedings{simplebaselines2018,
  title     = {Simple Baselines for Human Pose Estimation and Tracking},
  author    = {Xiao, Bin and Wu, Haiping and Wei, Yichen},
  booktitle = {Computer Vision -- ECCV 2018},
  series    = {Lecture Notes in Computer Science},
  volume    = {11210},
  pages     = {472--487},
  publisher = {Springer, Cham},
  year      = {2018}
}

@inproceedings{hrnet2019,
  title     = {Deep High-Resolution Representation Learning for Human Pose Estimation},
  author    = {Sun, Ke and Xiao, Bin and Liu, Dong and Wang, Jingdong},
  booktitle = {Proceedings of the IEEE/CVF Conference on Computer Vision and Pattern Recognition (CVPR)},
  pages     = {5693--5703},
  year      = {2019}
}

@article{openpose2021,
  title   = {{OpenPose}: Realtime Multi-Person 2{D} Pose Estimation Using Part Affinity Fields},
  author  = {Cao, Zhe and Hidalgo, Gines and Simon, Tomas and Wei, Shih-En and Sheikh, Yaser},
  journal = {IEEE Transactions on Pattern Analysis and Machine Intelligence},
  volume  = {43},
  number  = {1},
  pages   = {172--186},
  year    = {2021}
}

@inproceedings{vitpose2022,
  title     = {{ViTPose}: Simple Vision Transformer Baselines for Human Pose Estimation},
  author    = {Xu, Yufei and Zhang, Jing and Zhang, Qiming and Tao, Dacheng},
  booktitle = {Advances in Neural Information Processing Systems},
  volume    = {35},
  pages     = {38571--38584},
  year      = {2022}
}

@article{rtmpose2023,
  title   = {{RTMPose}: Real-Time Multi-Person Pose Estimation Based on {MMPose}},
  author  = {Jiang, Tao and Lu, Peng and Zhang, Li and Ma, Ningsheng and Han, Rui and Lyu, Chengqi and Li, Yining and Chen, Kai},
  journal = {arXiv preprint arXiv:2303.07399},
  year    = {2023},
  url     = {https://arxiv.org/abs/2303.07399}
}

@article{probiou2021,
  title   = {Gaussian Bounding Boxes and Probabilistic Intersection-over-Union for Object Detection},
  author  = {Llerena, Jeffri M. and Zeni, Luis Felipe and Kristen, Lucas N. and Jung, Claudio},
  journal = {arXiv preprint arXiv:2106.06072},
  year    = {2021},
  url     = {https://arxiv.org/abs/2106.06072}
}

@article{rrpn2018,
  title     = {Arbitrary-Oriented Scene Text Detection via Rotation Proposals},
  author    = {Ma, Jianqi and Shao, Weiyuan and Ye, Hao and Wang, Li and Wang, Hong and Zheng, Yingbin and Xue, Xiangyang},
  journal   = {IEEE Transactions on Multimedia},
  volume    = {20},
  number    = {11},
  pages     = {3111--3122},
  year      = {2018}
}

@inproceedings{roitransformer2019,
  title     = {Learning {RoI} Transformer for Oriented Object Detection in Aerial Images},
  author    = {Ding, Jian and Xue, Nan and Long, Yang and Xia, Gui-Song and Lu, Qikai},
  booktitle = {Proceedings of the IEEE/CVF Conference on Computer Vision and Pattern Recognition (CVPR)},
  pages     = {2849--2858},
  year      = {2019}
}

@inproceedings{orientedrcnn2021,
  title     = {Oriented {R-CNN} for Object Detection},
  author    = {Xie, Xingxing and Cheng, Gong and Wang, Jiabao and Yao, Xiwen and Han, Junwei},
  booktitle = {Proceedings of the IEEE/CVF International Conference on Computer Vision (ICCV)},
  pages     = {3520--3529},
  year      = {2021}
}

@inproceedings{csl2020,
  title     = {Arbitrary-Oriented Object Detection with Circular Smooth Label},
  author    = {Yang, Xue and Yan, Junchi},
  booktitle = {Proceedings of the European Conference on Computer Vision (ECCV)},
  pages     = {677--694},
  year      = {2020}
}

@inproceedings{gwd2021,
  title     = {Rethinking Rotated Object Detection with {Gaussian} Wasserstein Distance Loss},
  author    = {Yang, Xue and Yan, Junchi and Qi, Ming and Wang, Wentao and Zhang, Xiaopeng and Tian, Qi},
  booktitle = {Proceedings of the International Conference on Machine Learning (ICML)},
  pages     = {11580--11591},
  year      = {2021}
}

@inproceedings{kfiou2023,
  title     = {The {KFIoU} Loss for Rotated Object Detection},
  author    = {Yang, Xue and Liu, Gefan and Yang, Jirui and Yi, Jie and Liao, Wentong and He, Tao and Zhang, Jian and Yan, Junchi},
  booktitle = {International Conference on Learning Representations (ICLR)},
  year      = {2023}
}

@article{s2anet2021,
  title     = {Align Deep Features for Oriented Object Detection},
  author    = {Han, Jiaming and Ding, Jian and Li, Jie and Xia, Gui-Song},
  journal   = {IEEE Transactions on Geoscience and Remote Sensing},
  volume    = {60},
  pages     = {1--11},
  year      = {2022}
}

@inproceedings{lsknet2023,
  title     = {{LSKNet}: A Foundation Lightweight Backbone for Remote Sensing},
  author    = {Li, Yuxuan and Hou, Qibin and Zheng, Zhaohui and Cheng, Ming-Ming and Yang, Jian and Li, Xiang},
  booktitle = {Proceedings of the IEEE/CVF International Conference on Computer Vision (ICCV)},
  pages     = {4829--4840},
  year      = {2023}
}

@inproceedings{dinh2017realnvp,
  title     = {Density Estimation Using Real-Valued Non-Volume Preserving ({Real NVP}) Transformations},
  author    = {Dinh, Laurent and Sohl-Dickstein, Jascha and Bengio, Samy},
  booktitle = {International Conference on Learning Representations (ICLR)},
  year      = {2017}
}

@inproceedings{wang2023yolov7,
  title     = {{YOLOv7}: Trainable Bag-of-Freebies Sets New State-of-the-Art for Real-Time Object Detectors},
  author    = {Wang, Chien-Yao and Bochkovskiy, Alexey and Liao, Hong-Yuan Mark},
  booktitle = {Proceedings of the IEEE/CVF Conference on Computer Vision and Pattern Recognition (CVPR)},
  pages     = {7464--7475},
  year      = {2023}
}

@inproceedings{maji2022yolo,
  title={{YOLO-Pose}: Enhancing {YOLO} for Multi Person Pose Estimation Using Object Keypoint Similarity Loss},
  author={Maji, Debapriya and Nagori, Soyeb and Mathew, Manu and Poddar, Deepak},
  booktitle={Proceedings of the IEEE/CVF Conference on Computer Vision and Pattern Recognition (CVPR) Workshops},
  pages={2637--2646},
  year={2022},
  url={https://openaccess.thecvf.com/content/CVPR2022W/ECV/html/Maji_YOLO-Pose_Enhancing_YOLO_for_Multi_Person_Pose_Estimation_Using_Object_CVPRW_2022_paper.html}
}

@inproceedings{zhou2022mmrotate,
  title   = {MMRotate: A Rotated Object Detection Benchmark using PyTorch},
  author  = {Zhou, Yue and Yang, Xue and Zhang, Gefan and Wang, Jiabao and Liu, Yanyi and
             Hou, Liping and Jiang, Xue and Liu, Xingzhao and Yan, Junchi and Lyu, Chengqi and
             Zhang, Wenwei and Chen, Kai},
  booktitle={Proceedings of the 30th ACM International Conference on Multimedia},
  year={2022}
}

@article{bradski2000opencv,
  title={The OpenCV Library},
  author={Bradski, Gary},
  journal={Dr. Dobb's Journal of Software Tools},
  year={2000}
}

@inproceedings{yoloe,
  title     = {{YOLOE}: Real-Time Seeing Anything},
  author    = {Wang, Ao and Liu, Lihao and Chen, Hui and Lin, Zijia and Han, Jungong and Ding, Guiguang},
  booktitle = {Proceedings of the IEEE/CVF International Conference on Computer Vision (ICCV)},
  year      = {2025},
  eprint    = {2503.07465},
  archivePrefix = {arXiv},
  primaryClass  = {cs.CV},
}

@inproceedings{mobileclip,
  title     = {{MobileCLIP}: Fast Image-Text Models through Multi-Modal Reinforced Training},
  author    = {Vasu, Pavan Kumar Anasosalu and Pouransari, Hadi and Faghri, Fartash and Vemulapalli, Raviteja and Tuzel, Oncel},
  booktitle = {Proceedings of the IEEE/CVF Conference on Computer Vision and Pattern Recognition (CVPR)},
  year      = {2024},
  eprint    = {2311.17049},
  archivePrefix = {arXiv},
  primaryClass  = {cs.CV},
}

@article{mobileclip2,
  title   = {{MobileCLIP2}: Improving Multi-Modal Reinforced Training},
  author  = {Faghri, Fartash and Vasu, Pavan Kumar Anasosalu and Koc, Cem and Shankar, Vaishaal and Toshev, Alexander T. and Tuzel, Oncel and Pouransari, Hadi},
  journal = {Transactions on Machine Learning Research (TMLR)},
  year    = {2025},
  eprint  = {2508.20691},
  archivePrefix = {arXiv},
  primaryClass  = {cs.CV},
}

@inproceedings{li2022glip,
  title     = {Grounded Language-Image Pre-training},
  author    = {Li, Liunian Harold and Zhang, Pengchuan and Zhang, Haotian and Yang, Jianwei and Li, Chunyuan and Zhong, Yiwu and Wang, Lijuan and Yuan, Lu and Zhang, Lei and Hwang, Jenq-Neng and Chang, Kai-Wei and Gao, Jianfeng},
  booktitle = {Proceedings of the IEEE/CVF Conference on Computer Vision and Pattern Recognition (CVPR)},
  year      = {2022},
  eprint    = {2112.03857},
  archivePrefix = {arXiv},
  primaryClass  = {cs.CV},
}

@inproceedings{zhang2022glipv2,
  title     = {{GLIPv2}: Unifying Localization and Vision-Language Understanding},
  author    = {Zhang, Haotian and Zhang, Pengchuan and Hu, Xiaowei and Chen, Yen-Chun and Li, Liunian Harold and Dai, Xiyang and Wang, Lijuan and Yuan, Lu and Hwang, Jenq-Neng and Gao, Jianfeng},
  booktitle = {Advances in Neural Information Processing Systems (NeurIPS)},
  year      = {2022},
  eprint    = {2206.05836},
  archivePrefix = {arXiv},
  primaryClass  = {cs.CV},
}

@inproceedings{liu2024groundingdino,
  title     = {Grounding {DINO}: Marrying {DINO} with Grounded Pre-Training for Open-Set Object Detection},
  author    = {Liu, Shilong and Zeng, Zhaoyang and Ren, Tianhe and Li, Feng and Zhang, Hao and Yang, Jie and Jiang, Qing and Li, Chunyuan and Yang, Jianwei and Su, Hang and Zhu, Jun and Zhang, Lei},
  booktitle = {Proceedings of the European Conference on Computer Vision (ECCV)},
  year      = {2024},
  eprint    = {2303.05499},
  archivePrefix = {arXiv},
  primaryClass  = {cs.CV},
}

@inproceedings{yao2022detclip,
  title     = {{DetCLIP}: Dictionary-Enriched Visual-Concept Paralleled Pre-training for Open-world Detection},
  author    = {Yao, Lewei and Han, Jianhua and Wen, Youpeng and Liang, Xiaodan and Xu, Dan and Zhang, Wei and Li, Zhenguo and Xu, Chunjing and Xu, Hang},
  booktitle = {Advances in Neural Information Processing Systems (NeurIPS)},
  year      = {2022},
  eprint    = {2209.09407},
  archivePrefix = {arXiv},
  primaryClass  = {cs.CV},
}

@article{ren2024groundingdino15,
  title   = {Grounding {DINO} 1.5: Advance the ``Edge'' of Open-Set Object Detection},
  author  = {Ren, Tianhe and Liu, Shilong and Zeng, Zhaoyang and Lin, Hao and Li, Feng and Tang, Hao and Jiang, Qing and Li, Chunyuan and Yang, Jianwei and Su, Hang and Zhu, Jun and Zhang, Lei},
  journal = {arXiv preprint arXiv:2405.10300},
  year    = {2024},
  eprint  = {2405.10300},
  archivePrefix = {arXiv},
  primaryClass  = {cs.CV},
}

@article{jiang2024trex2,
  title         = {{T-Rex2}: Towards Generic Object Detection via Text-Visual Prompt Synergy},
  author        = {Jiang, Qing and Li, Feng and Zeng, Zhaoyang and Ren, Tianhe and Liu, Shilong and Zhang, Lei},
  journal       = {arXiv preprint arXiv:2403.14610},
  year          = {2024},
  eprint        = {2403.14610},
  archivePrefix = {arXiv},
  primaryClass  = {cs.CV},
}

@inproceedings{cheng2024yoloworld,
  title     = {{YOLO-World}: Real-Time Open-Vocabulary Object Detection},
  author    = {Cheng, Tianheng and Song, Lin and Ge, Yixiao and Liu, Wenyu and Wang, Xinggang and Shan, Ying},
  booktitle = {Proceedings of the IEEE/CVF Conference on Computer Vision and Pattern Recognition (CVPR)},
  year      = {2024},
  eprint    = {2401.17270},
  archivePrefix = {arXiv},
  primaryClass  = {cs.CV},
}

@inproceedings{lin2024generateu,
  title     = {Generative Region-Language Pretraining for Open-Ended Object Detection},
  author    = {Lin, Chuang and Jiang, Yi and Qu, Lizhen and Yuan, Zehuan and Cai, Jianfei},
  booktitle = {Proceedings of the IEEE/CVF Conference on Computer Vision and Pattern Recognition (CVPR)},
  year      = {2024},
  eprint    = {2403.10191},
  archivePrefix = {arXiv},
  primaryClass  = {cs.CV},
}

@article{yu2023mkiou,
  title={MKIoU loss: Toward accurate oriented object detection in aerial images},
  author={Yu, Xinyi and Lu, Jiangping and Lin, Mi and Zhou, Libo and Ou, Linlin},
  journal={Journal of Electronic Imaging},
  volume={32},
  number={3},
  pages={033030--033030},
  year={2023},
  publisher={Society of Photo-Optical Instrumentation Engineers}
}

@inproceedings{yu2023phase,
  title={Phase-shifting coder: Predicting accurate orientation in oriented object detection},
  author={Yu, Yi and Da, Feipeng},
  booktitle={Proceedings of the IEEE/CVF conference on computer vision and pattern recognition},
  pages={13354--13363},
  year={2023}
}

@inproceedings{huang2025open,
  title={Open-set image tagging with multi-grained text supervision},
  author={Huang, Xinyu and Huang, Yi-Jie and Zhang, Youcai and Tian, Weiwei and Feng, Rui and Zhang, Yuejie and Xie, Yanchun and Li, Yaqian and Zhang, Lei},
  booktitle={Proceedings of the 33rd ACM International Conference on Multimedia},
  pages={4117--4126},
  year={2025}
}

@article{wang2024yolouniow,
  title={YOLO-UniOW: Efficient Universal Open-World Object Detection},
  author={Liu, Lihao and Zeng, Juexiao and Gao, Xu and Yan, Baize and Luo, Yi and Wang, Guang and Zhuge, Yunzhe},
  journal={arXiv preprint arXiv:2412.20645},
  year={2024}
}

@inproceedings{minderer2022owlvit,
  title={Simple Open-Vocabulary Object Detection with Vision Transformers},
  author={Minderer, Matthias and Gritsenko, Alexey and Stone, Austin and Neumann, Maxim and Weissenborn, Dirk and Dosovitskiy, Alexey and Mahendran, Aravindh and Arnab, Anurag and Dehghani, Mostafa and Shen, Zhuoran and Wang, Xiao and Zhai, Xiaohua and Kipf, Thomas and Houlsby, Neil},
  booktitle={ECCV},
  year={2022}
}

@inproceedings{zang2022ovdetr,
  title={Open-Vocabulary DETR with Conditional Matching},
  author={Zang, Yuhang and Li, Wei and Zhou, Kaiyang and Huang, Chen and Loy, Chen Change},
  booktitle={ECCV},
  pages={106--122},
  year={2022}
}

@inproceedings{li2024dinov,
  title={Visual In-Context Prompting},
  author={Li, Feng and Jiang, Qing and Zhang, Hao and Ren, Tianhe and Liu, Shilong and Zou, Xueyan and Xu, Huaizhe and Li, Hongyang and Yang, Jianwei and Li, Chunyuan and Zhang, Lei and Gao, Jianfeng},
  booktitle={CVPR},
  pages={12861--12871},
  year={2024}
}

@article{zou2024seem,
  title={Segment Everything Everywhere All at Once},
  author={Zou, Xueyan and Yang, Jianwei and Zhang, Hao and Li, Feng and Li, Linjie and Wang, Jianfeng and Wang, Lijuan and Gao, Jianfeng and Lee, Yong Jae},
  journal={NeurIPS},
  year={2024}
}

@article{li2024semanticsam,
  title={Semantic-SAM: Segment and Recognize Anything at Any Granularity},
  author={Li, Feng and Zhang, Hao and Sun, Peize and Zou, Xueyan and Liu, Shilong and Yang, Jianwei and Li, Chunyuan and Zhang, Lei and Gao, Jianfeng},
  journal={arXiv preprint arXiv:2307.04767},
  year={2023}
}

@article{wu2022grit,
  title={GRiT: A Generative Region-to-Text Transformer for Object Understanding},
  author={Wu, Jialian and Wang, Jianfeng and Yang, Zhengyuan and Gan, Zhe and Liu, Zicheng and Yuan, Junsong and Wang, Lijuan},
  journal={arXiv preprint arXiv:2212.00280},
  year={2022}
}

@article{yao2024detclipv3,
  title={DetCLIPv3: Towards Versatile Generative Open-vocabulary Object Detection},
  author={Yao, Lewei and Han, Jianhua and Liang, Xiaodan and Xu, Dan and Zhang, Wei and Li, Zhenguo and Xu, Hang},
  journal={arXiv preprint arXiv:2404.09216},
  year={2024}
}

@inproceedings{xia2018dota,
  title={DOTA: A Large-Scale Dataset for Object Detection in Aerial Images},
  author={Xia, Gui-Song and Bai, Xiang and Ding, Jian and Zhu, Zhen and Belongie, Serge and Luo, Jiebo and Datcu, Mihai and Pelillo, Marcello and Zhang, Liangpei},
  booktitle={CVPR},
  pages={3974--3983},
  year={2018}
}
}

\clearpage
\beginsupplement
\section{Supplementary Materials}
\subsection{MuSGD Transfer to Classification}
\label{sec:suppl_musgd_cls}

To test whether the MuSGD advantage transfers beyond detection, we also perform a controlled ImageNet classification comparison, summarized in Table~\ref{tab:suppl_musgd_cls}. For each scale, the paired models share the same backbone architecture and training recipe; only the optimizer differs.

\begin{table}[t]
\centering
\small
\setlength{\tabcolsep}{4pt}
\resizebox{\columnwidth}{!}{%
\begin{tabular}{llccc}
\toprule
\textbf{Model} & \textbf{Optimizer} & \textbf{Top-1} $\uparrow$ & \textbf{Params (M)} & \textbf{FLOPs (B)} \\
\midrule
YOLO11n-cls & SGD   & 70.0 & 1.6 & 3.3 \\
YOLO26n-cls & MuSGD & \textbf{71.4} & 1.6 & 3.3 \\
\midrule
YOLO11s-cls & SGD   & 75.4 & 5.5 & 12.1 \\
YOLO26s-cls & MuSGD & \textbf{76.0} & 5.5 & 12.1 \\
\midrule
YOLO11m-cls & SGD   & 77.3 & 10.4 & 39.3 \\
YOLO26m-cls & MuSGD & \textbf{78.1} & 10.4 & 39.3 \\
\midrule
YOLO11l-cls & SGD   & 78.3 & 12.9 & 49.4 \\
YOLO26l-cls & MuSGD & \textbf{79.0} & 12.9 & 49.4 \\
\midrule
YOLO11x-cls & SGD   & 79.5 & 28.4 & 110.4 \\
YOLO26x-cls & MuSGD & \textbf{79.9} & 28.4 & 110.4 \\
\bottomrule
\end{tabular}
}
\vspace{2pt}
\caption{Controlled ImageNet classification comparison for MuSGD. Within each model pair, the backbone architecture and training recipe are matched, so the reported differences isolate the effect of the optimizer. MuSGD consistently improves top-1 accuracy across all scales.}
\label{tab:suppl_musgd_cls}
\end{table}

\subsection{Architecture Visualizations}
\label{sec:suppl_architecture}

\begin{figure*}[t]
\centering
\includegraphics[width=\textwidth]{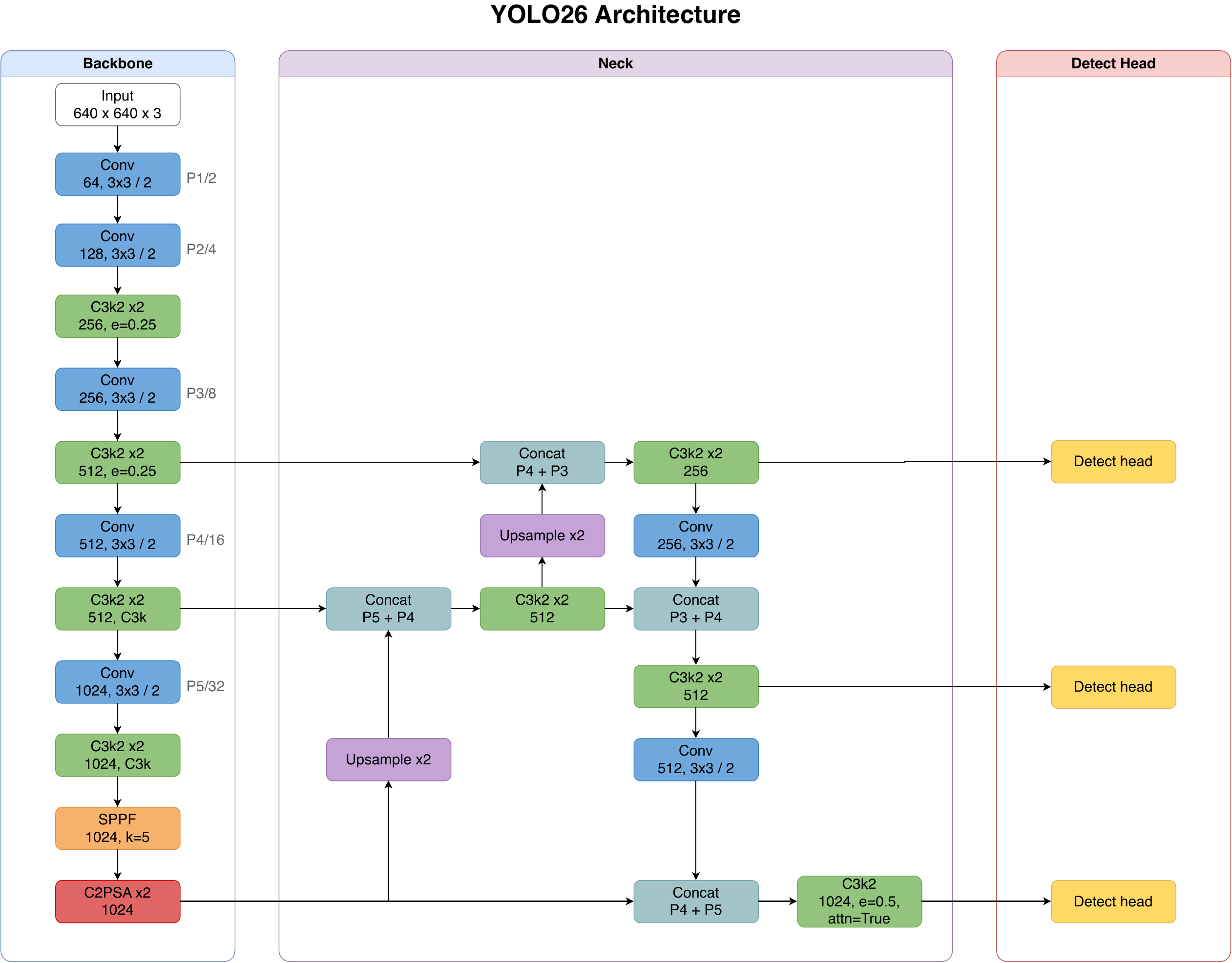}
\caption{Ultralytics YOLO26 architecture diagram. The figure summarizes the end-to-end model structure, including the shared backbone and neck, the task heads, and the high-level information flow across the network.}
\label{fig:suppl_yolo26_architecture}
\end{figure*}

\begin{figure*}[t]
\centering
\includegraphics[width=\textwidth]{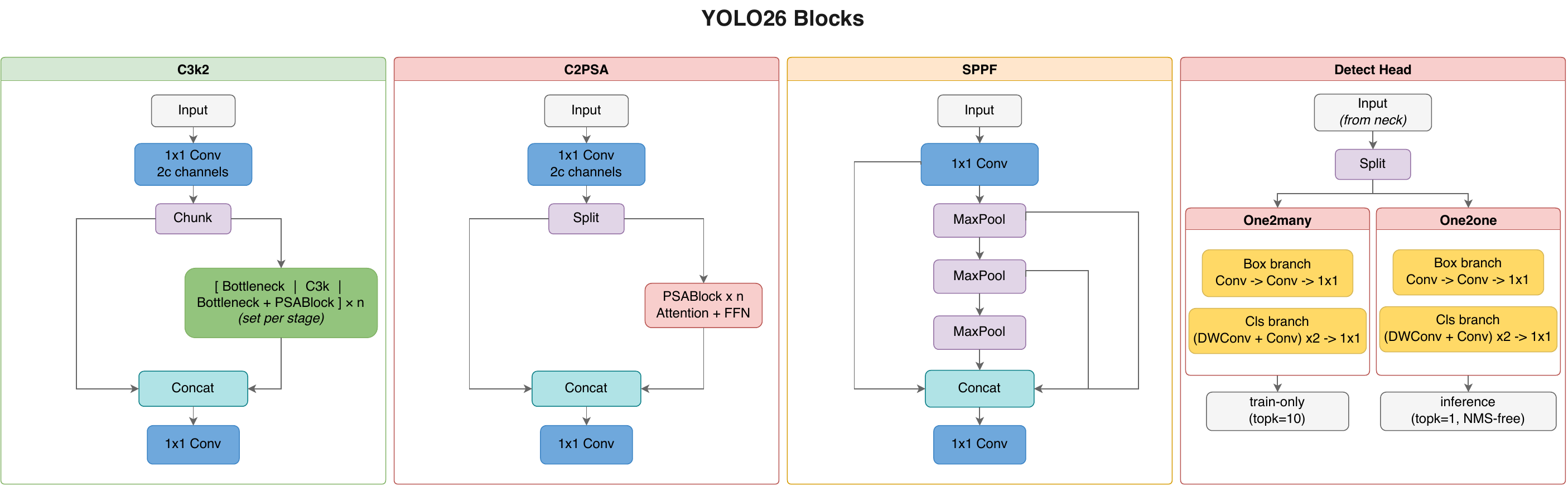}
\caption{Building blocks used to assemble the Ultralytics YOLO26 architecture. The figure details the constituent modules that compose the full network shown in Fig.~\ref{fig:suppl_yolo26_architecture}.}
\label{fig:suppl_yolo26_blocks}
\end{figure*}

For completeness, we provide two supplementary architecture figures that complement the main methodology section. Figure~\ref{fig:suppl_yolo26_architecture} presents the overall YOLO26 model architecture, while Figure~\ref{fig:suppl_yolo26_blocks} details the building blocks used to assemble the network. Together, these visualizations make the macro-architecture and the underlying module composition explicit.

\subsection{Training Recipes}
\label{sec:suppl_training_recipes}

\paragraph{Note on loss configuration fields.}
In the released Ultralytics training configuration, the \texttt{dfl} field controls the bounding-box regression loss gain. For YOLO11 and prior versions this governs the Distribution Focal Loss. In YOLO26, DFL is removed and direct regression with an L1 loss is used instead; the \texttt{dfl} field is retained for backward compatibility and now scales the L1 regression loss gain. Thus, a nonzero \texttt{dfl} value in the tables below does not indicate that DFL is active.

\paragraph{Objects365-v1 Pretraining.}

We first summarize the Objects365-v1 pretraining stage. Unless otherwise stated, Tables~\ref{tab:suppl_pretrain_recipe_optimizer}--\ref{tab:suppl_pretrain_recipe_internal} are transcribed directly from the \texttt{train\_args} stored in the released Objects365-v1 pretraining checkpoints, with most floating-point values rounded to two decimals for readability. We retain the original precision for \texttt{lr0} and \texttt{weight\_decay}. The pretraining stage uses 150 epochs at $640\times640$ resolution with batch size 128 and size-specific augmentation and internal settings.

\begin{table*}[t]
\centering
\small
\setlength{\tabcolsep}{5pt}
\begin{tabular}{lccccc}
\toprule
\textbf{Setting} & \textbf{YOLO26n} & \textbf{YOLO26s} & \textbf{YOLO26m} & \textbf{YOLO26l} & \textbf{YOLO26x} \\
\midrule
\multicolumn{6}{c}{\textbf{Optimizer and Schedule}} \\
\midrule
\texttt{optimizer} & MuSGD & MuSGD & MuSGD & MuSGD & MuSGD \\
\texttt{end2end} & -- & -- & -- & -- & -- \\
\texttt{imgsz} & 640 & 640 & 640 & 640 & 640 \\
\texttt{batch} & 128 & 128 & 128 & 128 & 128 \\
\texttt{epochs} & 150 & 150 & 150 & 150 & 150 \\
\texttt{lr0} & 0.01 & 0.01 & 0.01 & 0.01 & 0.01 \\
\texttt{lrf} & 0.01 & 0.01 & 0.01 & 0.01 & 0.01 \\
\texttt{momentum} & 0.94 & 0.94 & 0.94 & 0.94 & 0.94 \\
\texttt{weight\_decay} & 0.0005 & 0.0005 & 0.0005 & 0.0005 & 0.0005 \\
\texttt{warmup\_epochs} & 1 & 1 & 1 & 1 & 2 \\
\texttt{close\_mosaic} & 8 & 8 & 8 & 8 & 8 \\
\midrule
\multicolumn{6}{c}{\textbf{Loss Weights}} \\
\midrule
\texttt{box} & 7.50 & 7.50 & 7.50 & 7.50 & 7.50 \\
\texttt{cls} & 0.50 & 0.50 & 0.50 & 0.50 & 0.50 \\
\texttt{dfl} & 6.00 & 6.00 & 6.00 & 6.00 & 6.00 \\
\bottomrule
\end{tabular}
\vspace{2pt}
\caption{Official optimizer, schedule, and loss settings for the Objects365-v1 pretraining checkpoints.}
\label{tab:suppl_pretrain_recipe_optimizer}
\end{table*}

\begin{table*}[t]
\centering
\small
\setlength{\tabcolsep}{5pt}
\begin{tabular}{lccccc}
\toprule
\textbf{Setting} & \textbf{YOLO26n} & \textbf{YOLO26s} & \textbf{YOLO26m} & \textbf{YOLO26l} & \textbf{YOLO26x} \\
\midrule
\texttt{mosaic} & 1.00 & 1.00 & 1.00 & 1.00 & 1.00 \\
\texttt{mixup} & 0.00 & 0.05 & 0.15 & 0.15 & 0.20 \\
\texttt{copy\_paste} & 0.10 & 0.15 & 0.40 & 0.50 & 0.60 \\
\texttt{scale} & 0.50 & 0.90 & 0.90 & 0.90 & 0.90 \\
\texttt{fliplr} & 0.50 & 0.50 & 0.50 & 0.50 & 0.50 \\
\texttt{degrees} & 0.00 & 0.00 & 0.00 & 0.00 & 0.00 \\
\texttt{shear} & 0.00 & 0.00 & 0.00 & 0.00 & 0.00 \\
\texttt{translate} & 0.10 & 0.10 & 0.10 & 0.10 & 0.10 \\
\texttt{hsv\_h} & 0.02 & 0.02 & 0.02 & 0.02 & 0.02 \\
\texttt{hsv\_s} & 0.70 & 0.70 & 0.70 & 0.70 & 0.70 \\
\texttt{hsv\_v} & 0.40 & 0.40 & 0.40 & 0.40 & 0.40 \\
\texttt{bgr} & 0.00 & 0.00 & 0.00 & 0.00 & 0.00 \\
\bottomrule
\end{tabular}
\vspace{2pt}
\caption{Official size-specific augmentation settings for the Objects365-v1 pretraining checkpoints.}
\label{tab:suppl_pretrain_recipe_aug}
\end{table*}

\begin{table*}[t]
\centering
\small
\setlength{\tabcolsep}{5pt}
\begin{tabular}{lccccc}
\toprule
\textbf{Setting} & \textbf{YOLO26n} & \textbf{YOLO26s} & \textbf{YOLO26m} & \textbf{YOLO26l} & \textbf{YOLO26x} \\
\midrule
\texttt{muon\_w} & 0.45 & 0.50 & 0.45 & 0.45 & 0.50 \\
\texttt{sgd\_w} & 0.55 & 0.50 & 0.55 & 0.55 & 0.60 \\
\texttt{cls\_w} & 1.00 & 1.00 & 1.00 & 1.00 & 1.00 \\
\texttt{o2m} & 0.10 & 0.10 & 0.10 & 1.00 & 1.00 \\
\bottomrule
\end{tabular}
\vspace{2pt}
\caption{Checkpoint-recorded internal training parameters associated with the Objects365-v1 pretraining checkpoints.}
\label{tab:suppl_pretrain_recipe_internal}
\end{table*}

\paragraph{COCO Fine-Tuning.}

We next summarize the COCO fine-tuning stage. Unless otherwise stated, Tables~\ref{tab:suppl_training_recipe_optimizer}--\ref{tab:suppl_training_recipe_internal} are transcribed directly from the \texttt{train\_args} stored in the released COCO checkpoints, with most floating-point values rounded to two decimals for readability. We retain the original precision for \texttt{lr0} and \texttt{weight\_decay}, and use $\sim 0$ when a positive value rounds to zero at two-decimal precision. The released detection checkpoints use end-to-end training at $640\times640$ resolution with batch size 128, are initialized from the Objects365-v1 pretraining checkpoints above, and use size-specific hyperparameters obtained by evolutionary search.

\begin{table*}[t]
\centering
\small
\setlength{\tabcolsep}{5pt}
\begin{tabular}{lccccc}
\toprule
\textbf{Setting} & \textbf{YOLO26n} & \textbf{YOLO26s} & \textbf{YOLO26m} & \textbf{YOLO26l} & \textbf{YOLO26x} \\
\midrule
\multicolumn{6}{c}{\textbf{Optimizer and Schedule}} \\
\midrule
\texttt{optimizer} & MuSGD & MuSGD & MuSGD & MuSGD & MuSGD \\
\texttt{end2end} & True & True & True & True & True \\
\texttt{imgsz} & 640 & 640 & 640 & 640 & 640 \\
\texttt{batch} & 128 & 128 & 128 & 128 & 128 \\
\texttt{epochs} & 245 & 70 & 80 & 60 & 40 \\
\texttt{lr0} & 0.0054 & 0.00038 & 0.00038 & 0.00038 & 0.00038 \\
\texttt{lrf} & 0.05 & 0.88 & 0.88 & 0.88 & 0.88 \\
\texttt{momentum} & 0.95 & 0.95 & 0.95 & 0.95 & 0.95 \\
\texttt{weight\_decay} & 0.00064 & 0.00027 & 0.00027 & 0.00027 & 0.00027 \\
\texttt{warmup\_epochs} & 0.98 & 0.99 & 0.99 & 0.99 & 0.99 \\
\texttt{close\_mosaic} & 10 & 10 & 10 & 10 & 10 \\
\midrule
\multicolumn{6}{c}{\textbf{Loss Weights}} \\
\midrule
\texttt{box} & 5.63 & 9.83 & 9.83 & 9.83 & 9.83 \\
\texttt{cls} & 0.56 & 0.65 & 0.65 & 0.65 & 0.65 \\
\texttt{dfl} & 9.04 & 0.96 & 0.96 & 0.96 & 0.96 \\
\bottomrule
\end{tabular}
\vspace{2pt}
\caption{Official optimizer, schedule, and loss settings for the released YOLO26 COCO checkpoints.}
\label{tab:suppl_training_recipe_optimizer}
\end{table*}

\begin{table*}[t]
\centering
\small
\setlength{\tabcolsep}{5pt}
\begin{tabular}{lccccc}
\toprule
\textbf{Setting} & \textbf{YOLO26n} & \textbf{YOLO26s} & \textbf{YOLO26m} & \textbf{YOLO26l} & \textbf{YOLO26x} \\
\midrule
\texttt{mosaic} & 0.91 & 0.99 & 0.99 & 0.99 & 0.99 \\
\texttt{mixup} & 0.01 & 0.05 & 0.43 & 0.43 & 0.43 \\
\texttt{copy\_paste} & 0.08 & 0.40 & 0.30 & 0.40 & 0.40 \\
\texttt{scale} & 0.56 & 0.90 & 0.95 & 0.95 & 0.95 \\
\texttt{fliplr} & 0.61 & 0.30 & 0.30 & 0.30 & 0.30 \\
\texttt{degrees} & 1.11 & $\sim 0$ & $\sim 0$ & $\sim 0$ & $\sim 0$ \\
\texttt{shear} & 1.46 & $\sim 0$ & $\sim 0$ & $\sim 0$ & $\sim 0$ \\
\texttt{translate} & 0.07 & 0.27 & 0.27 & 0.27 & 0.27 \\
\texttt{hsv\_h} & 0.01 & 0.01 & 0.01 & 0.01 & 0.01 \\
\texttt{hsv\_s} & 0.64 & 0.35 & 0.35 & 0.35 & 0.35 \\
\texttt{hsv\_v} & 0.57 & 0.19 & 0.19 & 0.19 & 0.19 \\
\texttt{bgr} & 0.11 & 0.00 & 0.00 & 0.00 & 0.00 \\
\bottomrule
\end{tabular}
\vspace{2pt}
\caption{Official size-specific augmentation settings for the released YOLO26 COCO checkpoints.}
\label{tab:suppl_training_recipe_aug}
\end{table*}

\begin{table*}[t]
\centering
\small
\setlength{\tabcolsep}{5pt}
\begin{tabular}{lccccc}
\toprule
\textbf{Setting} & \textbf{YOLO26n} & \textbf{YOLO26s} & \textbf{YOLO26m} & \textbf{YOLO26l} & \textbf{YOLO26x} \\
\midrule
\texttt{muon\_w} & 0.53 & 0.44 & 0.44 & 0.44 & 0.44 \\
\texttt{sgd\_w} & 0.67 & 0.48 & 0.48 & 0.48 & 0.48 \\
\texttt{cls\_w} & 2.74 & 3.48 & 3.48 & 3.48 & 3.48 \\
\texttt{o2m} & 1.00 & 0.71 & 0.71 & 0.71 & 0.71 \\
\texttt{topk} & 8 & 5 & 5 & 5 & 5 \\
\bottomrule
\end{tabular}
\vspace{2pt}
\caption{Checkpoint-recorded internal training parameters associated with the released YOLO26 COCO checkpoints.}
\label{tab:suppl_training_recipe_internal}
\end{table*}

\subsection{Comparison with Recent Real-Time Detectors}
\label{sec:suppl_literature_comparison}

Table~\ref{tab:suppl_selected_literature_comparison} provides the full grouped s/m/l/x comparison with recent real-time detectors. It complements Sec.~\ref{sec:main_coco_detection} by giving the complete per-scale breakdown used to position YOLO26 in the main paper.

\begin{table*}[t]
\centering
\small
\setlength{\tabcolsep}{4pt}
\begin{tabular}{lccccccccc}
\toprule
\textbf{Model} & \textbf{Params} & \textbf{GFLOPs} & \textbf{Latency} & \textbf{AP$^{val}$} & \textbf{AP$^{val}_{50}$} & \textbf{AP$^{val}_{75}$} & \textbf{AP$^{val}_S$} & \textbf{AP$^{val}_M$} & \textbf{AP$^{val}_L$} \\
 & \textbf{(M)} &  & \textbf{(ms)} &  &  &  &  &  &  \\
\midrule
\multicolumn{10}{c}{\textbf{S-scale}} \\
\midrule
YOLOv9-S~\cite{wang2024yolov9} & \textbf{7} & 27 & 3.5 & 46.8 & 61.8 & 48.6 & 25.7 & 49.9 & 61.0 \\
YOLOv10-S~\cite{wang2024yolov10} & \textbf{7} & 22 & \textbf{2.5} & 46.3 & 63.0 & 50.4 & 26.8 & 51.0 & 63.8 \\
YOLOv12-S~\cite{tian2025yolov12} & 9 & 21 & 2.6 & 48.0 & 65.0 & 51.8 & 29.8 & 53.2 & 65.6 \\
YOLO11-S~\cite{ultralytics2024yolo11_docs} & \underline{9} & 22 & \textbf{2.5} & 47.0 & 63.9 & 50.7 & 29.0 & 51.7 & 64.4 \\
D-FINE-S~\cite{dfine} & 10 & 25 & \underline{3.5} & 48.5 & 65.6 & 52.6 & 29.1 & 52.2 & 65.4 \\
DEIM-S~\cite{deim} & 10 & 25 & \underline{3.5} & \textbf{49.0} & \textbf{65.9} & \textbf{53.1} & \textbf{30.4} & \underline{52.6} & \underline{65.7} \\
YOLO26s & 10 & \underline{21} & \textbf{2.5} & \underline{48.6} & \underline{65.8} & \underline{52.8} & \underline{29.5} & \textbf{53.2} & \textbf{65.8} \\
\midrule
\multicolumn{10}{c}{\textbf{M-scale}} \\
\midrule
YOLOv9-M~\cite{wang2024yolov9} & 20 & 77 & 6.4 & 51.4 & 67.2 & 54.6 & 32.0 & 55.7 & 66.4 \\
YOLOv10-M~\cite{wang2024yolov10} & \textbf{15} & \underline{59} & \underline{4.7} & 51.1 & 68.1 & 55.8 & 33.8 & 56.5 & 67.0 \\
RT-DETRv2-S~\cite{lv2024rt} & 20 & 60 & \textbf{4.6} & 48.1 & 65.1 & \underline{57.4} & \underline{36.1} & \textbf{57.9} & \textbf{70.8} \\
YOLOv12-M~\cite{tian2025yolov12} & 20 & 68 & 4.9 & 52.5 & 69.6 & 57.1 & 35.7 & 58.2 & 68.8 \\
YOLO11-M~\cite{ultralytics2024yolo11_docs} & 20 & 68 & \underline{4.7} & 51.5 & 68.5 & 55.7 & 33.4 & 57.1 & 67.9 \\
D-FINE-M~\cite{dfine} & \underline{19} & \textbf{57} & 5.6 & 52.3 & 69.8 & 56.4 & 33.2 & 56.5 & \underline{70.2} \\
DEIM-M~\cite{deim} & \underline{19} & \textbf{57} & 5.6 & \underline{52.7} & \underline{70.0} & 57.3 & 35.3 & 56.7 & 69.5 \\
YOLO26m & 20 & 68 & \underline{4.7} & \textbf{53.1} & \textbf{70.7} & \textbf{57.7} & \textbf{36.7} & \underline{57.8} & 68.9 \\
\midrule
\multicolumn{10}{c}{\textbf{L-scale}} \\
\midrule
YOLOv9-C~\cite{wang2024yolov9} & 26 & 103 & 7.2 & 53.0 & 70.2 & 57.8 & 36.2 & 58.5 & 69.3 \\
YOLOv10-L~\cite{wang2024yolov10} & \textbf{24} & 120 & 7.3 & 53.2 & 70.1 & 58.1 & 35.8 & 58.5 & 69.4 \\
YOLOv12-L~\cite{tian2025yolov12} & 26 & 89 & 6.8 & 53.7 & 70.7 & 58.5 & 36.9 & 59.5 & 69.9 \\
YOLO11-L~\cite{ultralytics2024yolo11_docs} & \underline{25} & 87 & \textbf{6.2} & 53.4 & 70.1 & 58.2 & 35.6 & 59.1 & 69.2 \\
D-FINE-L~\cite{dfine} & 31 & 91 & 8.1 & 54.0 & 71.6 & 58.4 & 36.5 & 58.0 & \underline{71.9} \\
DEIM-L~\cite{deim} & 31 & 91 & 8.1 & \underline{54.7} & \underline{72.4} & \underline{59.4} & \underline{36.9} & \textbf{59.6} & 71.8 \\
RT-DETRv2-M~\cite{lv2024rt} & 31 & 92 & \underline{6.9} & 49.9 & 67.5 & 58.6 & 35.8 & 58.6 & \textbf{72.1} \\
YOLO26l & \underline{25} & \underline{86} & \textbf{6.2} & \textbf{55.0} & \textbf{72.5} & \textbf{60.0} & \textbf{38.4} & \underline{59.5} & 71.1 \\
\midrule
\multicolumn{10}{c}{\textbf{X-scale}} \\
\midrule
YOLOv9-E~\cite{wang2024yolov9} & 58 & 193 & 16.8 & 55.6 & 72.8 & 60.6 & 40.2 & 61.0 & 71.4 \\
YOLOv10-X~\cite{wang2024yolov10} & \textbf{30} & \textbf{160} & \textbf{10.7} & 54.4 & 71.3 & 59.3 & 37.0 & 59.8 & 70.9 \\
YOLOv12-X~\cite{tian2025yolov12} & 59 & 199 & 11.8 & 55.2 & 72.0 & 60.2 & 39.6 & 60.7 & 70.9 \\
YOLO11-X~\cite{ultralytics2024yolo11_docs} & 57 & 195 & \underline{11.3} & 54.7 & 71.6 & 59.5 & 37.7 & 59.7 & 70.2 \\
D-FINE-X~\cite{dfine} & 62 & 202 & 12.9 & 55.8 & 73.7 & 60.2 & 37.3 & 60.5 & \underline{73.4} \\
DEIM-X~\cite{deim} & 62 & 202 & 12.9 & \underline{56.5} & \underline{74.0} & \underline{61.5} & \underline{38.8} & \underline{61.4} & \textbf{74.2} \\
RT-DETRv2-X~\cite{lv2024rt} & 76 & 259 & 13.9 & 54.3 & 72.8 & 58.8 & 35.8 & 58.8 & 72.1 \\
YOLO26x & \underline{56} & 194 & 11.8 & \textbf{57.5} & \textbf{75.0} & \textbf{62.7} & \textbf{41.8} & \textbf{62.1} & 73.3 \\
\bottomrule
\end{tabular}
\vspace{2pt}
\caption{\textbf{Comparison with recent real-time detectors on COCO \texttt{val2017}, shown in grouped s/m/l/x form for readability.} Literature baselines are taken from the corresponding primary papers or release benchmarks. To keep the computational-complexity columns aligned with released benchmark tables, we report only the standard YOLO26 NMS variants here. For YOLO26, Params/GFLOPs/T4 TensorRT latency are taken from Table~\ref{tab:yolo26_coco_detection}, while the AP breakdowns come from the corresponding NMS validation runs.}
\label{tab:suppl_selected_literature_comparison}
\end{table*}

\subsection{Additional Task Benchmarks}
\label{sec:suppl_task_benchmarks}

This subsection collects the full YOLO11-versus-YOLO26
benchmark tables for the additional task results
discussed in Sec.~\ref{sec:task_results}. We place
these full model-family comparisons in the
supplementary materials so that the main paper can
focus on the task-specific ablations and methodological
takeaways.

\paragraph{Instance Segmentation.}

Table~\ref{tab:yolo26_coco_ins_seg} reports the full
COCO instance segmentation benchmark comparison
between YOLO11 and YOLO26.

\begin{table*}[t]
\centering
\small
\setlength{\tabcolsep}{6pt}
\begin{tabular}{lccccccc}
\toprule
\textbf{Model} & \textbf{Size} & \textbf{mAP$^{box}$} & \textbf{mAP$^{mask}$} & \textbf{CPU ONNX} & \textbf{T4 TRT10} & \textbf{Params} & \textbf{FLOPs} \\
  & \textbf{(px)} & \textbf{val 50--95} & \textbf{val 50--95} & \textbf{(ms)} & \textbf{(ms)} & \textbf{(M)} & \textbf{(B)} \\
\midrule
YOLO11n-seg & 640 & 38.9 & 32.0 & 65.9 & 1.8 & 2.9 & 9.7 \\
YOLO26n-seg (E2E) & 640 & 39.6 & 33.9 & 53.5 & 2.1 & 2.7 & 9.1 \\
YOLO26n-seg (Non-E2E) & 640 & 40.6 & 34.4 & 52.7 & 2.0 & 2.7 & 9.1 \\
\midrule
YOLO11s-seg & 640 & 46.6 & 37.8 & 117.6 & 2.9 & 10.1 & 33.0 \\
YOLO26s-seg (E2E) & 640 & 47.3 & 40.0 & 118.4 & 3.3 & 10.4 & 34.2 \\
YOLO26s-seg (Non-E2E) & 640 & 48.2 & 40.5 & 102.4 & 3.2 & 10.4 & 34.2 \\
\midrule
YOLO11m-seg & 640 & 51.5 & 41.5 & 281.6 & 6.3 & 22.4 & 113.2 \\
YOLO26m-seg (E2E) & 640 & 52.5 & 44.1 & 328.2 & 6.7 & 23.6 & 121.5 \\
YOLO26m-seg (Non-E2E) & 640 & 53.1 & 44.4 & 337.7 & 6.9 & 23.6 & 121.5 \\
\midrule
YOLO11l-seg & 640 & 53.4 & 42.9 & 344.2 & 7.8 & 27.6 & 132.2 \\
YOLO26l-seg (E2E) & 640 & 54.4 & 45.5 & 387.0 & 8.0 & 28.0 & 139.8 \\
YOLO26l-seg (Non-E2E) & 640 & 55.2 & 46.0 & 395.8 & 8.2 & 28.0 & 139.8 \\
\midrule
YOLO11x-seg & 640 & 54.7 & 43.8 & 664.5 & 15.8 & 62.1 & 296.4 \\
YOLO26x-seg (E2E) & 640 & 56.5 & 47.0 & 787.0 & 16.4 & 62.8 & 313.5 \\
YOLO26x-seg (Non-E2E) & 640 & 57.2 & 47.5 & 795.9 & 16.8 & 62.8 & 313.5 \\
\bottomrule
\end{tabular}
\vspace{2pt}
\caption{Comparisons of instance segmentation models on COCO validation set, where `E2E` corresponds to using the one-to-one branch for inference without NMS and `Non-E2E` corresponds to using the one-to-many branch for inference with NMS. CPU ONNX and T4 TensorRT latency values exclude NMS time.}
\label{tab:yolo26_coco_ins_seg}
\end{table*}

\paragraph{Pose Estimation.}

Table~\ref{tab:yolo26_coco_pose} reports the full COCO
pose estimation benchmark comparison between YOLO11
and YOLO26.

\begin{table*}[t]
\centering
\small
\setlength{\tabcolsep}{6pt}
\begin{tabular}{lccccccc}
\toprule
\textbf{Model} & \textbf{Size} & \textbf{mAP} & \textbf{AP} & \textbf{CPU ONNX} & \textbf{T4 TRT10} & \textbf{Params} & \textbf{FLOPs} \\
  & \textbf{(px)} & \textbf{val 50--95} & \textbf{val 50} & \textbf{(ms)} & \textbf{(ms)} & \textbf{(M)} & \textbf{(B)} \\
\midrule
YOLO11n-pose & 640 & 50.0 & 81.0 & 52.4 & 1.7 & 2.9 & 7.4 \\
YOLO26n-pose (E2E) & 640 & 57.2 & 83.3 & 40.3 & 1.8 & 2.9 & 7.5 \\
YOLO26n-pose (Non-E2E) & 640 & 57.0 & 83.1 & 40.8 & 1.8 & 2.9 & 7.5 \\
\midrule
YOLO11s-pose & 640 & 58.9 & 86.3 & 90.5 & 2.6 & 9.9 & 23.1 \\
YOLO26s-pose (E2E) & 640 & 63.0 & 86.6 & 85.3 & 2.7 & 10.4 & 23.9 \\
YOLO26s-pose (Non-E2E) & 640 & 62.9 & 86.6 & 88.6 & 2.8 & 10.4 & 23.9 \\
\midrule
YOLO11m-pose & 640 & 64.9 & 89.4 & 187.3 & 4.9 & 20.9 & 71.4 \\
YOLO26m-pose (E2E) & 640 & 68.8 & 89.9 & 218.0 & 5.0 & 21.5 & 73.1 \\
YOLO26m-pose (Non-E2E) & 640 & 68.8 & 89.6 & 228.9 & 5.1 & 21.5 & 73.1 \\
\midrule
YOLO11l-pose & 640 & 66.1 & 89.9 & 247.7 & 6.4 & 26.1 & 90.3 \\
YOLO26l-pose (E2E) & 640 & 70.4 & 90.5 & 275.4 & 6.5 & 25.9 & 91.3 \\
YOLO26l-pose (Non-E2E) & 640 & 70.3 & 90.6 & 268.9 & 6.5 & 25.9 & 91.3 \\
\midrule
YOLO11x-pose & 640 & 69.5 & 91.1 & 488.0 & 12.1 & 58.8 & 202.8 \\
YOLO26x-pose (E2E) & 640 & 71.6 & 91.6 & 565.4 & 12.2 & 57.6 & 201.7 \\
YOLO26x-pose (Non-E2E)& 640 & 71.7 & 91.0 & 574.4 & 12.4 & 57.6 & 201.7 \\
\bottomrule
\end{tabular}
\vspace{2pt}
\caption{Comparisons of pose estimation models on COCO validation set, where `E2E` corresponds to using the one-to-one branch for inference without NMS and `Non-E2E` corresponds to using the one-to-many branch for inference with NMS. CPU ONNX and T4 TensorRT latency values exclude NMS time.}
\label{tab:yolo26_coco_pose}
\end{table*}

\paragraph{Oriented Bounding Box Detection.}

Table~\ref{tab:yolo26_dota_obb} reports the full
DOTA-v1.0 OBB benchmark comparison between YOLO11 and
YOLO26.

\begin{table*}[t]
\centering
\small
\setlength{\tabcolsep}{6pt}
\begin{tabular}{lcccccccc}
\toprule
\textbf{Model} & \textbf{Size} & \textbf{mAP} & \textbf{AP} & \textbf{AP} & \textbf{CPU ONNX} & \textbf{T4 TRT10} & \textbf{Params} & \textbf{FLOPs} \\
  & \textbf{(px)} & \textbf{val 50--95} & \textbf{val 50} & \textbf{val 75} & \textbf{(ms)} & \textbf{(ms)} & \textbf{(M)} & \textbf{(B)} \\
\midrule
YOLO11n-obb & 1024 & 49.7 & 78.4 & 52.1 & 117.6 & 4.4 & 2.7 & 16.8 \\
YOLO26n-obb & 1024 & 52.4 & 78.9 & 56.8 & 97.7 & 2.8 & 2.5 & 14.0 \\
\midrule
YOLO11s-obb & 1024 & 51.4 & 79.5 & 54.4 & 219.4 & 5.1 & 9.7 & 57.1 \\
YOLO26s-obb & 1024 & 54.8 & 80.9 & 60.4 & 218.0 & 4.9 & 9.8 & 55.1 \\
\midrule
YOLO11m-obb & 1024 & 52.8 & 80.9 & 56.1 & 562.8 & 10.1 & 20.9 & 182.8 \\
YOLO26m-obb & 1024 & 55.3 & 81.0 & 60.7 & 579.2 & 10.2 & 21.2 & 183.3 \\
\midrule
YOLO11l-obb & 1024 & 52.9 & 81.0 & 56.6 & 712.5 & 13.5 & 26.1 & 231.2 \\
YOLO26l-obb & 1024 & 56.2 & 81.6 & 62.2 & 735.6 & 13.0 & 25.6 & 230.0 \\
\midrule
YOLO11x-obb & 1024 & 54.1 & 81.3 & 57.8 & 1408.6 & 28.6 & 58.8 & 519.1 \\
YOLO26x-obb & 1024 & 56.7 & 81.7 & 62.6 & 1485.7 & 30.5 & 57.6 & 516.5 \\
\bottomrule
\end{tabular}
\vspace{2pt}
\caption{Comparisons of OBB models on the DOTA-v1.0 test set, where YOLO26 models are evaluated using the one-to-one branch without NMS.}
\label{tab:yolo26_dota_obb}
\end{table*}

\subsection{YOLOE-26 Implementation Details}
\label{sec:impl_details}

All YOLOE-26 models are trained with a total batch size of 256 across multiple GPUs.
The training follows a four-stage pipeline: text-prompt training (TP) is conducted first, and
its best checkpoint then serves as the initialization for three parallel downstream branches---visual
prompt (VP), prompt-free (PF), and segmentation (SEG).

\noindent\textbf{TP stage.}
Each backbone is initialized from a YOLO26 checkpoint pretrained on Objects365-v1 for 150 epochs.
We use the MuSGD optimizer with an initial learning rate of $1.25\!\times\!10^{-3}$,
a final learning rate ratio $\text{lr}_f\!=\!0.5$, momentum $0.9$, and weight decay
$5\!\times\!10^{-4}$ ($7\!\times\!10^{-4}$ for the S scale).
Mosaic augmentation is disabled for the last 2 epochs (\texttt{close\_mosaic}$\,{=}\,2$).
The number of training epochs is scale-dependent: 30 for n/s, 25 for m, 20 for l, and 15 for x.
Data augmentation strength also scales with model capacity: copy-paste probability increases from 0.1 (N) to 0.6 (X), and mixup from 0.0 (N) to 0.2 (X).
The L and X scales additionally adopt tuned hyperparameters from a separate search.

\noindent\textbf{VP and PF stages.}
Both branches fine-tune from the best TP checkpoint using AdamW with an initial learning rate of
$2\!\times\!10^{-3}$, $\text{lr}_f\!=\!0.01$, momentum $0.9$, and weight decay $0.025$.
All scales are trained for 10 epochs.
For VP training, only the SAVPE module and \texttt{cv4} heads are unfrozen;
for PF training, only the final classification layers (\texttt{cv3}) are unfrozen.

\noindent\textbf{SEG stage.}
The segmentation branch also fine-tunes from the best TP checkpoint but uses MuSGD
with the same learning rate schedule as the TP stage
($\text{lr}_0\!=\!1.25\!\times\!10^{-3}$, $\text{lr}_f\!=\!0.5$).
Only the segmentation-specific layers (\texttt{cv5}, \texttt{proto}) are unfrozen,
and training runs for 10 epochs across all scales.

\begin{figure}[t]
\centering
\includegraphics[width=0.5\textwidth]{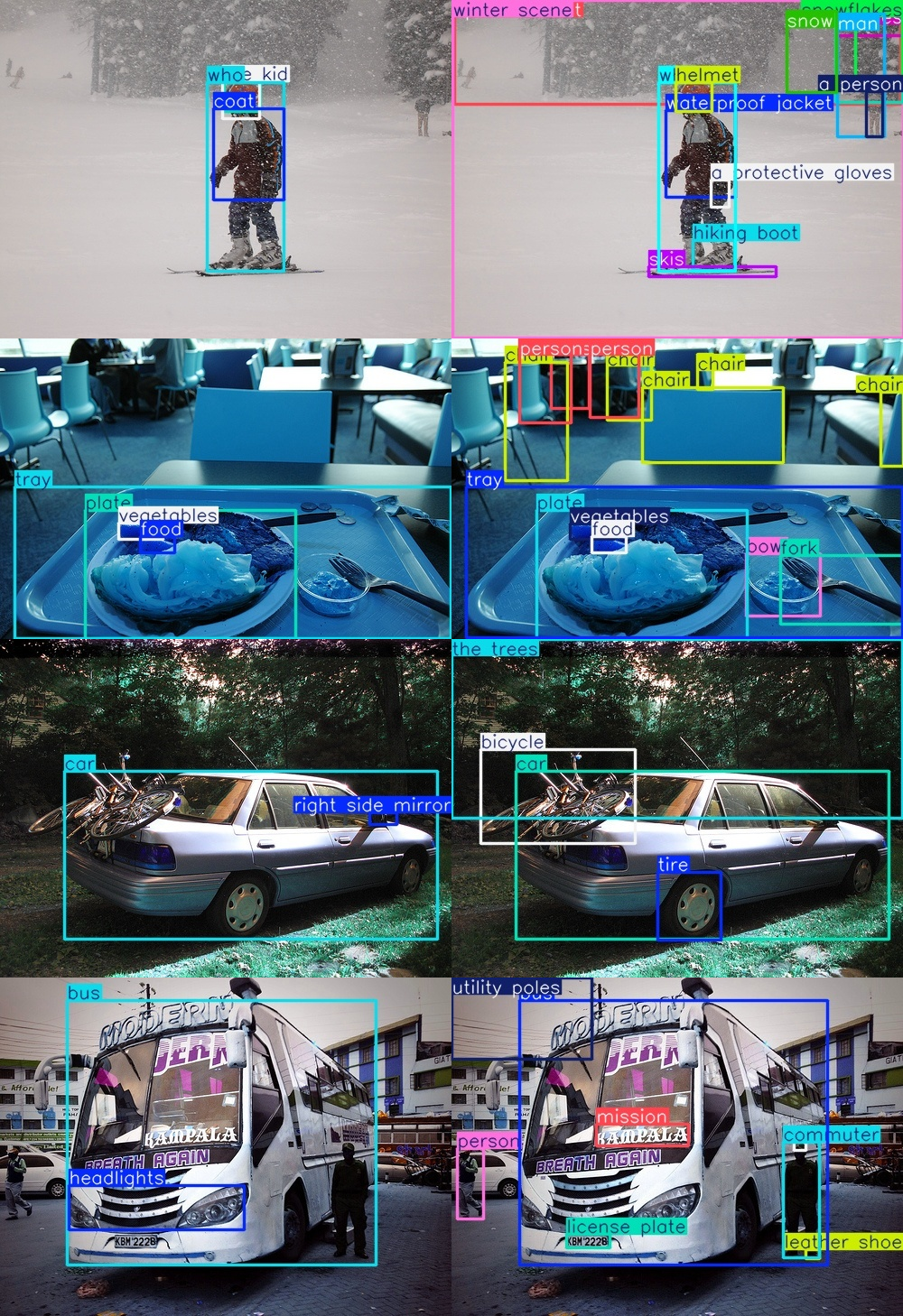}
\caption{Comparison of four samples before (left) and after (right) refinement by the YOLOE data engine.}
\label{fig:suppl_yoloe_data_engine_compare}
\end{figure}

\noindent\textbf{Training data.}
All four stages share the same training data, consisting of three grounding
datasets: Objects365-v1~\cite{shao2019objects365},
GQA~\cite{hudson2019gqa}, and Flickr30k~\cite{plummer2015flickr30k}.
Following YOLOE~\cite{yoloe}, we adopt the YOLOE data engine to
refine the annotations of all three datasets. \cref{fig:suppl_yoloe_data_engine_compare} illustrates representative
before-and-after examples, showing that the refined annotations
exhibit fewer missing instances.

\subsection{Additional YOLOE-26 Benchmarks}
\label{sec:suppl_yoloe_benchmarks}

This subsection collects the full YOLOE-26 benchmark
tables complementary to Sec.~\ref{sec:yoloe26_results}.

\paragraph{Prompt-based Detection.}

Table~\ref{tab:yoloe26_det_prompt_detail} reports the
full LVIS \texttt{minival} comparison for text-prompted
and visual-prompted open-vocabulary detection.

\begin{table*}[t]
\centering
\small
\setlength{\tabcolsep}{3pt}
\begin{tabular}{lccccccccc}
\toprule
Model & Size & Prompt &
mAP$^{\text{e2e}}_{\text{50--95}}$ &
mAP$_{\text{50--95}}$ &
mAP$_r$ &
mAP$_c$ &
mAP$_f$ &
Params~(M) & FLOPs~(B) \\
\midrule
GLIP-T~\cite{li2022glip}      & --  & T   & --- & 26.0 & 20.8 & 21.4 & 31.0 & 232       & ---          \\
GLIPv2-T~\cite{zhang2022glipv2}    & --  & T   & --- & 29.0 & ---  & ---  & ---  & 232       & ---          \\
GDINO-T~\cite{liu2024groundingdino}     & --  & T   & --- & 27.4 & 18.1 & 23.3 & 32.7 & 172       & ---          \\
DetCLIP-T~\cite{yao2022detclip}   & --  & T   & --- & 34.4 & 26.9 & 33.9 & 36.3 & 155       & ---          \\
G-1.5 Edge~\cite{ren2024groundingdino15}  & --  & T   & --- & 33.5 & 28.0 & 34.3 & 33.9 & ---       & ---          \\
T-Rex2~\cite{jiang2024trex2}       & --  & V   & --- & 37.4 & 29.9 & 33.9 & 41.8 & ---       & ---          \\
\midrule
YOLOE-26n & 640 & T/V & 23.7 / 20.9 & 24.7 / 21.9 & 20.5 / 17.6 & 24.1 / 22.3 & 26.1 / 22.4 &  \textbf{3.9} /  \textbf{3.1}  &   \textbf{6.1}  \\
\midrule
YOLOWorldv2-S~\cite{cheng2024yoloworld}   & 640 & T   & --- & 24.4 & 17.1 & 22.5 & 27.3 & 13        & ---          \\
YOLOE-v8s~\cite{yoloe} & 640 & T/V & --- & 27.9 / 26.2 & 22.3 / 21.3 & 27.8 / 27.7 & 29.0 / 25.7 & 12.3 / 12.6  & 29.8   \\
YOLOE-11s~\cite{yoloe} & 640 & T/V & --- & 27.5 / 26.3 & 21.4 / 22.5 & 26.8 / 27.1 & 29.3 / 26.4 & \textbf{10.7} / \textbf{10.9}  & 22.7  \\
YOLOE-26s & 640 & T/V & \textbf{29.9} / \textbf{27.1} & \textbf{30.8} / \textbf{28.6} & \textbf{23.9} / \textbf{25.1} & \textbf{29.6} / \textbf{27.8} & \textbf{33.0} / \textbf{29.9} & 10.7 / 11.0  &  \textbf{21.9}  \\
\midrule
YOLOWorldv2-M~\cite{cheng2024yoloworld}   & 640 & T   & --- & 32.4 & 28.4 & 29.6 & 35.5 & 29        & ---          \\
YOLOE-v8m~\cite{yoloe} & 640 & T/V & --- & 32.6 / 31.0 & 26.9 / 27.0 & 31.9 / 31.7 & 34.4 / 31.1 & 26.4 / 28.4  & 80.7   \\
YOLOE-11m~\cite{yoloe} & 640 & T/V & --- & 33.0 / 31.4 & 26.9 / 27.1 & 32.5 / 31.9 & 34.5 / 31.7 & \textbf{21.0} / \textbf{24.8}  & \textbf{70.4}  \\
YOLOE-26m & 640 & T/V & \textbf{35.4} / \textbf{31.3} & \textbf{35.4} / \textbf{33.9} & \textbf{31.1} / \textbf{33.4} & \textbf{34.7} / \textbf{34.0} & \textbf{36.9} / \textbf{33.8} & 21.3 / 25.1  &  70.6  \\
\midrule
YOLOWorldv2-L~\cite{cheng2024yoloworld}   & 640 & T   & --- & 35.5 & 25.6 & 34.6 & 38.1 & 48        & ---          \\
YOLOE-v8l~\cite{yoloe} & 640 & T/V & --- & 35.9 / 34.2 & 33.2 / 33.2 & 34.8 / 34.6 & 37.3 / 34.1 & 43.5 / 47.3  & 167.6  \\
YOLOE-11l~\cite{yoloe} & 640 & T/V & --- & 35.2 / 33.7 & 29.1 / 28.1 & 35.0 / 34.6 & 36.5 / 33.8 & \textbf{26.0} / \textbf{29.8}  & \textbf{89.5}   \\
YOLOE-26l & 640 & T/V & \textbf{36.8} / \textbf{33.7} & \textbf{37.8} / \textbf{36.3} & \textbf{35.1} / \textbf{37.6} & \textbf{37.6} / \textbf{36.2} & \textbf{38.5} / \textbf{36.1} & 25.5 / 29.3  &  89.0  \\
\midrule
YOLOE-26x & 640 & T/V & \textbf{39.5} / \textbf{36.2} & \textbf{40.6} / \textbf{38.5} & \textbf{37.4} / \textbf{35.3} & \textbf{40.9} / \textbf{38.8} & \textbf{41.0} / \textbf{38.8} & 55.2 / 65.2  & 197.7  \\
\bottomrule
\end{tabular}
\vspace{2pt}
\caption{Detection results with text and visual prompts on LVIS minival. T\,=\,Text prompt, V\,=\,Visual prompt; metric values are reported as T\,/\,V.}
\label{tab:yoloe26_det_prompt_detail}
\end{table*}

Under text prompting, YOLOE-26 consistently improves
over prior YOLOE variants at matched scales, with gains
of \textbf{+3.3/+2.4/+2.6~AP} over YOLOE-11 and
\textbf{+2.9/+2.8/+1.9~AP} over YOLOE-v8 at the s/m/l
scales. YOLOE-26x achieves the best text-prompted
result among the compared methods, reaching
\textbf{40.6~AP}. At the lightweight end, YOLOE-26n
reaches 24.7~AP with only 3.9M parameters, indicating
that the gains also extend to resource-constrained
settings. For text-prompted inference, the end-to-end
head remains close to the Non-E2E variant, trailing by
at most 1.1 AP across scales. Under visual prompting,
the gap ranges from 1.0 to 2.6 AP. YOLOE-26 again
improves consistently over earlier YOLOE families, and
YOLOE-26x attains the best visual-prompted result at
\textbf{38.5~AP}. In
addition, the s/m/l models obtain higher AP$_r$ than
their text-prompted counterparts, suggesting that
visual prompts are particularly beneficial for rare
categories at these scales.

\paragraph{Zero-shot Segmentation.}

Table~\ref{tab:seg} reports the full zero-shot
segmentation comparison on the LVIS \texttt{val} set.

\begin{table*}[t]
\centering
\small
\setlength{\tabcolsep}{6pt}
\begin{tabular}{l c cccc}
\toprule
Model & Prompt & $\text{AP}^{m}$ & $\text{AP}^{m}_{r}$ & $\text{AP}^{m}_{c}$ & $\text{AP}^{m}_{f}$ \\
\midrule
YOLOE-v8s  & T / V & 17.7 / 16.8 & 15.5 / 13.5 & 16.3 / 16.7 & 20.3 / 18.2 \\
YOLOE-11s  & T / V & 17.6 / 17.1 & 16.1 / 14.4 & 15.6 / 16.8 & 20.5 / 18.6 \\
YOLOE-26s  & T / V & \textbf{20.5} / \textbf{19.1} & \textbf{18.4} / \textbf{16.1} & \textbf{18.6} / \textbf{18.2} & \textbf{23.4} / \textbf{21.4} \\
\midrule
YOLOE-26n  & T / V & 15.1 / 13.9 & 11.9 / 11.0 & 13.9 / 13.4 & 17.9 / 15.6 \\
\midrule
YOLOWorld-M$^\dagger$   & T   & 16.7 & 12.6 & 14.6 & 20.8 \\
YOLOWorldv2-M$^\dagger$ & T   & 17.8 & 13.9 & 15.5 & 22.0 \\
YOLOE-v8m  & T / V & 20.8 / 20.3 & 17.2 / 17.0 & 19.2 / 20.1 & 24.2 / 22.0 \\
YOLOE-11m  & T / V & 21.1 / 21.0 & 17.2 / 18.3 & 19.6 / 20.6 & 24.4 / 22.6 \\
YOLOE-26m  & T / V & \textbf{23.8} / \textbf{22.9} & \textbf{24.2} / \textbf{23.3} & \textbf{25.6} / \textbf{26.6} & \textbf{29.4} / \textbf{28.4} \\
\midrule
YOLOWorld-L$^\dagger$   & T   & 19.1 & 14.2 & 17.2 & 23.5 \\
YOLOWorldv2-L$^\dagger$ & T   & 19.8 & 15.0 & 17.5 & 23.6 \\
YOLOE-v8l  & T / V & 23.5 / 22.0 & 21.9 / 16.5 & 21.6 / 22.1 & 26.4 / 24.3 \\
YOLOE-11l  & T / V & 22.6 / 22.5 & 19.3 / 20.5 & 20.9 / 21.7 & 26.0 / 24.1 \\
YOLOE-26l  & T / V & \textbf{24.8} / \textbf{24.3} & \textbf{21.9} / \textbf{21.9} & \textbf{23.3} / \textbf{23.6} & \textbf{27.8} / \textbf{26.1} \\
\midrule
YOLOE-26x  & T / V & \textbf{27.4} / \textbf{26.7} & \textbf{24.9} / \textbf{23.3} & \textbf{26.2} / \textbf{26.6} & \textbf{29.8} / \textbf{28.4} \\
\bottomrule
\end{tabular}
\vspace{2pt}
\caption{\textbf{Segmentation evaluation on LVIS.} We evaluate all models on LVIS \texttt{val} set with the standard $\text{AP}^{m}$ reported. YOLOE supports both text (T) and visual cues (V) as inputs. $\dagger$ indicates that the pretrained models are fine-tuned on \texttt{LVIS-Base} data for segmentation head. In contrast, we evaluate YOLOE in a zero-shot manner without utilizing any images from LVIS during training.}
\label{tab:seg}
\end{table*}

We extend YOLOE-26 with a lightweight segmentation
head and evaluate zero-shot mask prediction on the
LVIS \texttt{val} set. YOLOE-26 consistently
outperforms prior YOLOE variants across all model
scales. Notably, YOLOE-26s achieves 20.5 / 19.1
$\text{AP}^{m}$ (T / V), surpassing YOLOE-v8m
(20.8 / 20.3) at less than half the computational
cost, demonstrating the efficiency advantage brought
by the YOLO26 backbone. At the large scale, YOLOE-26l
reaches 24.8 / 24.3 $\text{AP}^{m}$, exceeding
YOLOE-v8l by +1.3 / +2.3 points while also
outperforming fine-tuned baselines such as
YOLOWorldv2-L$^\dagger$ (19.8) by a significant
margin. Scaling further to YOLOE-26x yields the best
overall performance of 27.4 / 26.7 $\text{AP}^{m}$,
with particularly strong gains on rare categories
($\text{AP}^{m}_{r}$~=~24.9 / 23.3).

\paragraph{Prompt-free Detection.}

Table~\ref{tab:yoloe26_pf} reports the full prompt-free
open-vocabulary detection comparison on LVIS
\texttt{minival}.

\begin{table*}[t]
\centering
\small
\setlength{\tabcolsep}{6pt}
\begin{tabular}{lcccccc}
\toprule
Model & Params & AP & AP$_r$ & AP$_c$ & AP$_f$ & FLOPs \\
\midrule
GenerateU-T~\cite{lin2024generateu} & 297  & 26.8 & 20.0 & 24.9 & 29.8 & --- \\
GenerateU-L~\cite{lin2024generateu} & 467  & 27.9 & 22.3 & 25.2 & 31.4 & --- \\
\midrule
YOLOE-26n &  \textbf{2.3} & 16.6/17.7 & 15.7/15.8 & 15.3/16.4 & 17.9/19.2 &  \textbf{5.3} \\
\midrule
YOLOE-v8s~\cite{yoloe} & 13 & 21.0 & 19.1 & 21.3 & 21.0 & --- \\
YOLOE-11s~\cite{yoloe} & \textbf{11} & 20.6 & 18.4 & 20.2 & 21.3 & --- \\
YOLOE-26s &  9.0 & \textbf{21.4}/\textbf{22.6} & \textbf{16.2}/\textbf{20.2} & \textbf{20.1}/\textbf{20.9} & \textbf{23.5}/\textbf{24.5} & 20.8 \\
\midrule
YOLOE-v8m~\cite{yoloe} & 29 & 24.7 & 22.2 & 24.5 & 25.3 & --- \\
YOLOE-11m~\cite{yoloe} & \textbf{24} & 25.5 & 21.6 & 25.5 & 26.1 & --- \\
YOLOE-26m & 19.4 & \textbf{25.7}/\textbf{26.4} & \textbf{26.7}/\textbf{24.5} & \textbf{24.0}/\textbf{25.0} & \textbf{26.9}/\textbf{27.9} & 68.4 \\
\midrule
YOLOE-v8l~\cite{yoloe} & 47 & 27.2 & 23.5 & \textbf{27.0} & \textbf{28.0} & --- \\
YOLOE-11l~\cite{yoloe} & \textbf{29} & 26.3 & 22.7 & 25.8 & 27.5 & --- \\
YOLOE-26l & 23.6 & \textbf{27.2}/\textbf{28.0} & \textbf{26.3}/\textbf{25.7} & 25.7/\textbf{26.8} & \textbf{28.7}/\textbf{29.5} & 86.8 \\
\midrule
YOLOE-26x & 53.1 & \textbf{29.9/31.1} & \textbf{27.5/28.9} & \textbf{29.1/30.7} & \textbf{31.1/31.7} & 194.4 \\
\bottomrule
\end{tabular}
\vspace{2pt}
\caption{Prompt-free detection on LVIS minival.
         AP is Fixed AP$_{50\text{--}95}$; subscripts denote
         rare/common/frequent splits.
         YOLOE-26 reports E2E\,/\,Non-E2E.
         GenerateU-T/L use Swin-T/L backbones.
         YOLOE-26 trains on Objects365-v1~\cite{shao2019objects365},
         GQA~\cite{hudson2019gqa}, Flickr30k~\cite{plummer2015flickr30k}.}
\label{tab:yoloe26_pf}
\end{table*}

YOLOE-26 achieves competitive prompt-free detection
across the model family. At the standard Non-E2E
operating point, YOLOE-26x attains the best AP in
Table~\ref{tab:yoloe26_pf} with \textbf{31.1~AP}.
YOLOE-26l reaches 28.0~AP with only 23.6M parameters,
comparable to GenerateU-L~\cite{lin2024generateu}
(27.9~AP, 467M parameters) while using nearly
20$\times$ fewer parameters. Relative to YOLOE-v8 and
YOLOE-11, YOLOE-26s/m/l (Non-E2E) improve AP by
\textbf{0.8--1.7} and \textbf{0.9--2.0},
respectively. The E2E head remains competitive,
trailing the Non-E2E variant by only
\textbf{0.7--1.2~AP} while removing post-processing.

\end{document}